\pdfoutput=1
\documentclass[12pt]{report}
\usepackage[utf8]{inputenc}
\usepackage[german, english]{babel}
\usepackage{graphicx}
\usepackage{amsmath}
\usepackage{amssymb}
\usepackage{mathtools}
\usepackage{float}
\usepackage[makeroom]{cancel}
\usepackage[numbers]{natbib}
\usepackage{booktabs}
\usepackage{subcaption}
\usepackage{siunitx}
\usepackage{awesomebox}
\usepackage[most]{tcolorbox}
\usepackage{tabularx}
\usepackage{comment}
\usepackage{bm}
\usepackage{afterpage}
\usepackage{pifont}
\usepackage{pdflscape}
\usepackage[backref=page]{hyperref}
\usepackage[a4paper, total={390pt, 548pt}]{geometry} 
\usepackage{enumitem}
\usepackage{titlesec}
\titleformat{\chapter}{\normalfont\LARGE\bfseries}{\thechapter}{1em}{}
\titlespacing*{\chapter}{0pt}{2.5ex plus 1ex minus .2ex}{5.3ex plus .2ex}

\graphicspath{ {figures/} }

\newcommand*\wildcard[2][5cm]{%
	\mbox{%
		\vbox to 1cm{%
			\vfill
			\hbox to #1{\hrulefill}\par
			\hbox to #1{\strut\hfil#2\hfil}
		}
	}
}

\newcommand{\xmark}{\ding{55}}%

\newcommand{\x}{\bm x}
\newcommand{\z}{\bm z}
\newcommand{\rep}{\bm r}

\newcommand{\expectation}[2]{\mathbb{E}_{#1} \bigg[ #2 \bigg]}

\DeclarePairedDelimiterX{\kldivx}[2]{(}{)}{%
	#1\;\delimsize\|\;#2%
}
\newcommand{\kldiv}{D_{\mathrm{KL}}\kldivx}
\DeclarePairedDelimiter{\norm}{\lVert}{\rVert}

\newcommand{\eq}[1]{Eq.\@~\ref{#1}}
\newcommand{\fig}[1]{Fig.\@~\ref{#1}}
\newcommand{\tab}[1]{Tab.\@~\ref{#1}}
\renewcommand{\sec}[1]{Sec.\@~\ref{#1}}

\newcommand{\FigDALLEsamples}
{
	\begin{figure}[H]
		\centering
		\begin{subfigure}[b]{0.48\textwidth}
			\includegraphics[width=\textwidth]{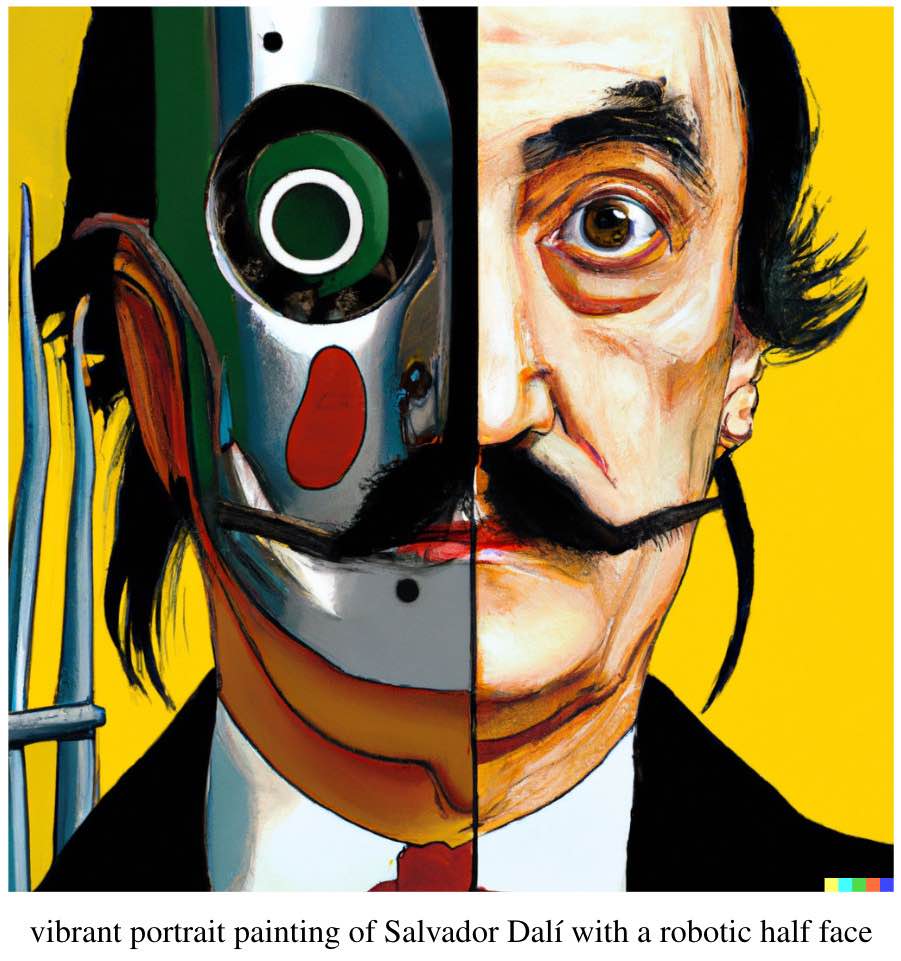}
		\end{subfigure}
		\begin{subfigure}[b]{0.48\textwidth}
			\includegraphics[width=\textwidth]{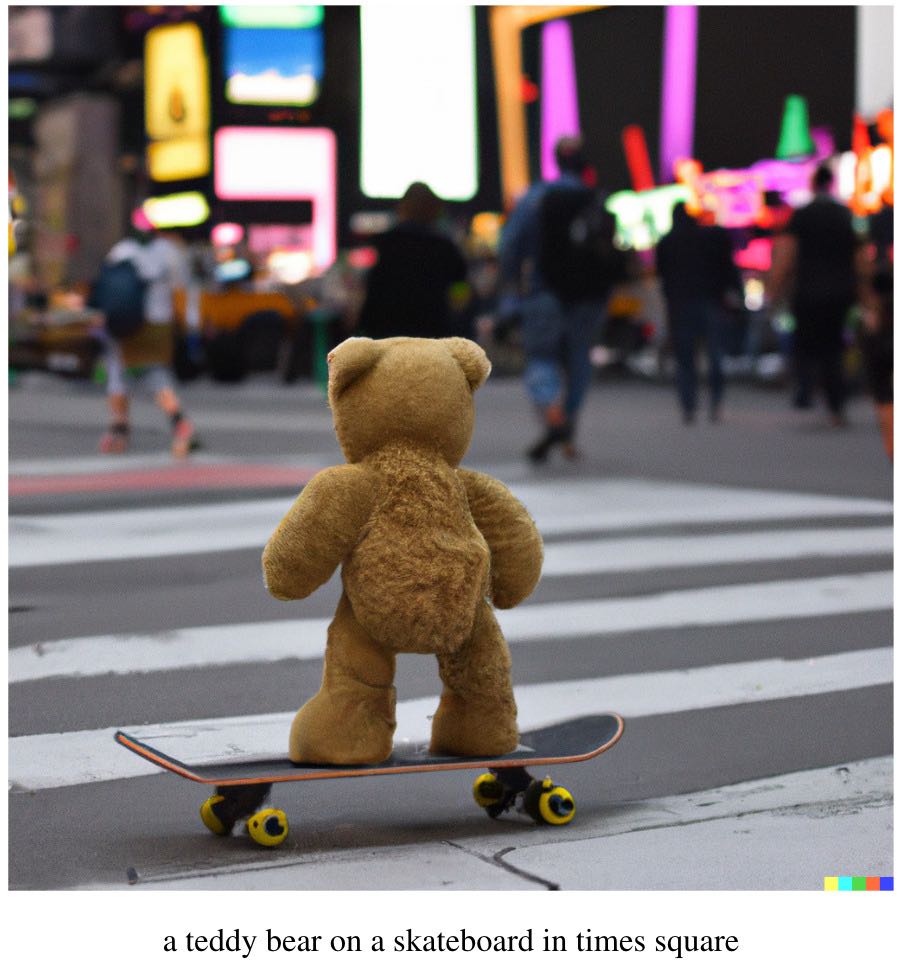}
		\end{subfigure}
		\caption[DALL-E 2 samples]{
			\begin{minipage}[t]{0.8\textwidth}
				Visual Synthesis with DALL-E 2 \cite{ramesh_hierarchical_2022}. The model generates new images based on text captions. It is based on diffusion models conditioned on CLIP \cite{radford_learning_2021} embeddings to allow for guidance through the textual captions. Taken from \cite{ramesh_hierarchical_2022}.
			\end{minipage}
		}
		\label{fig: dalle2}
	\end{figure}
}

\newcommand{\FigOverviewVAE}
{
	\begin{figure}[H]
		\centering
		\includegraphics[width=0.15\textwidth]{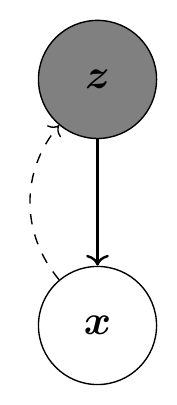}
		\caption[VAE as probabilistic graphical model]{
			\begin{minipage}[t]{0.8\textwidth}
				Graphical model representation of a VAE. Shaded circles denote latent variables, unshaded circles denote observable variables. Solid lines represent the generative path, dashed lines represent the encoder path.
			\end{minipage}
		}
		\label{fig: overview vae}
	\end{figure}
}

\newcommand{\FigDDPM}
{
	\begin{figure}[H]
		\centering
		\includegraphics[width=\textwidth]{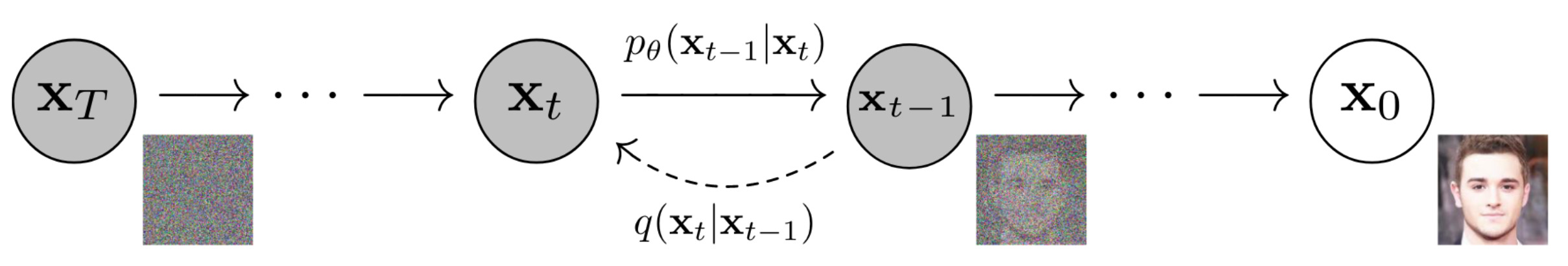}
		\caption[Diffusion model as probabilistic graphical model]{
			\begin{minipage}[t]{0.8\textwidth}
				Diffusion Model as a directed graphical model. In each step of the forward process, gaussian noise is added to the input. The model learns the reverse process by learning to recover the information that is destroyed in each step. From \cite{ho_denoising_2020}.
			\end{minipage}
		}
		\label{fig: ddpm}
	\end{figure}
}

\newcommand{\FigLossWeightings}
{
	\begin{figure}[H]
		\centering
		\begin{subfigure}[b]{0.425\textwidth}
			\includegraphics[width=\textwidth]{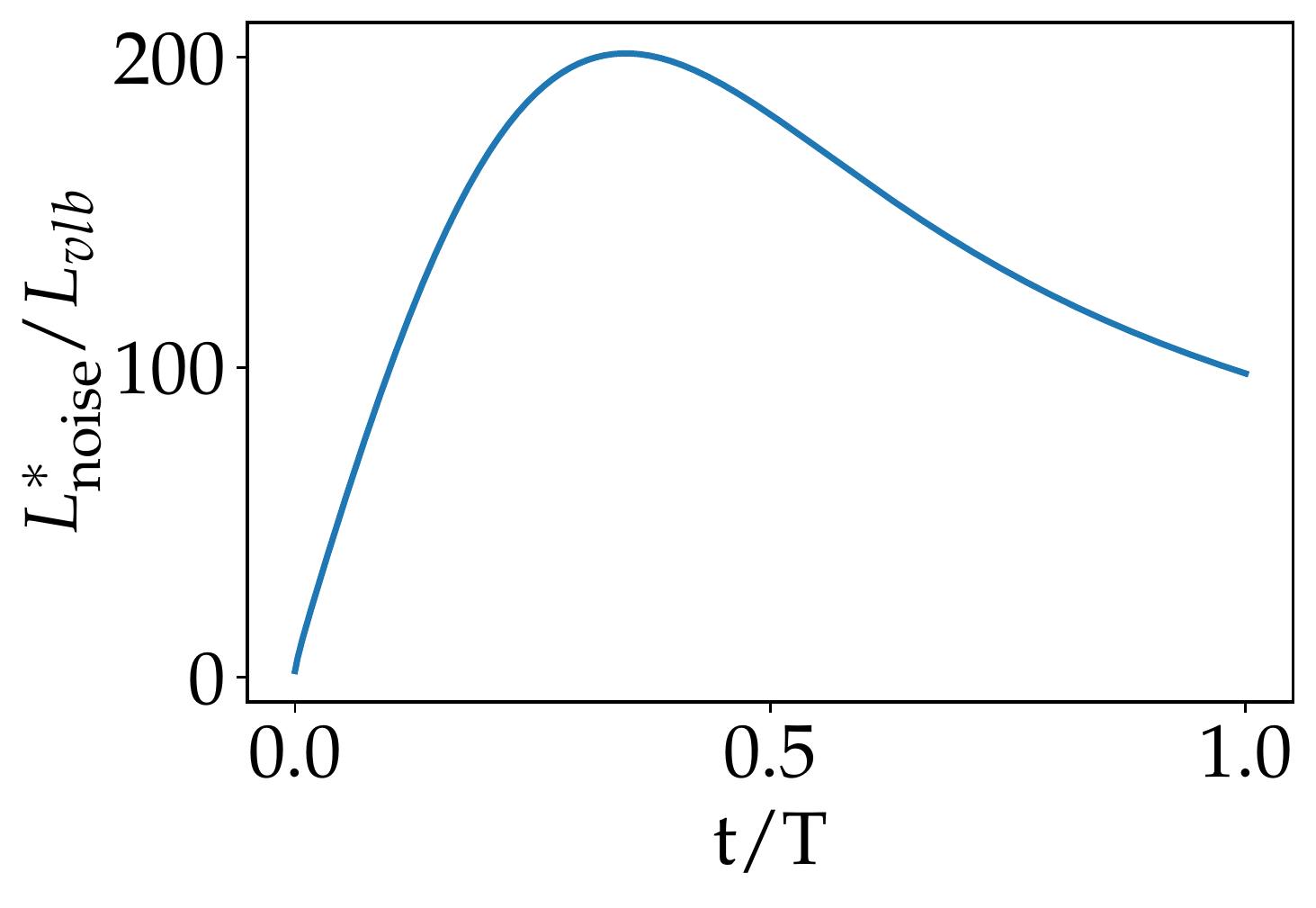}
			\subcaption{Noise-parameterization}
		\end{subfigure}
		\begin{subfigure}[b]{0.4\textwidth}
			\includegraphics[width=\textwidth]{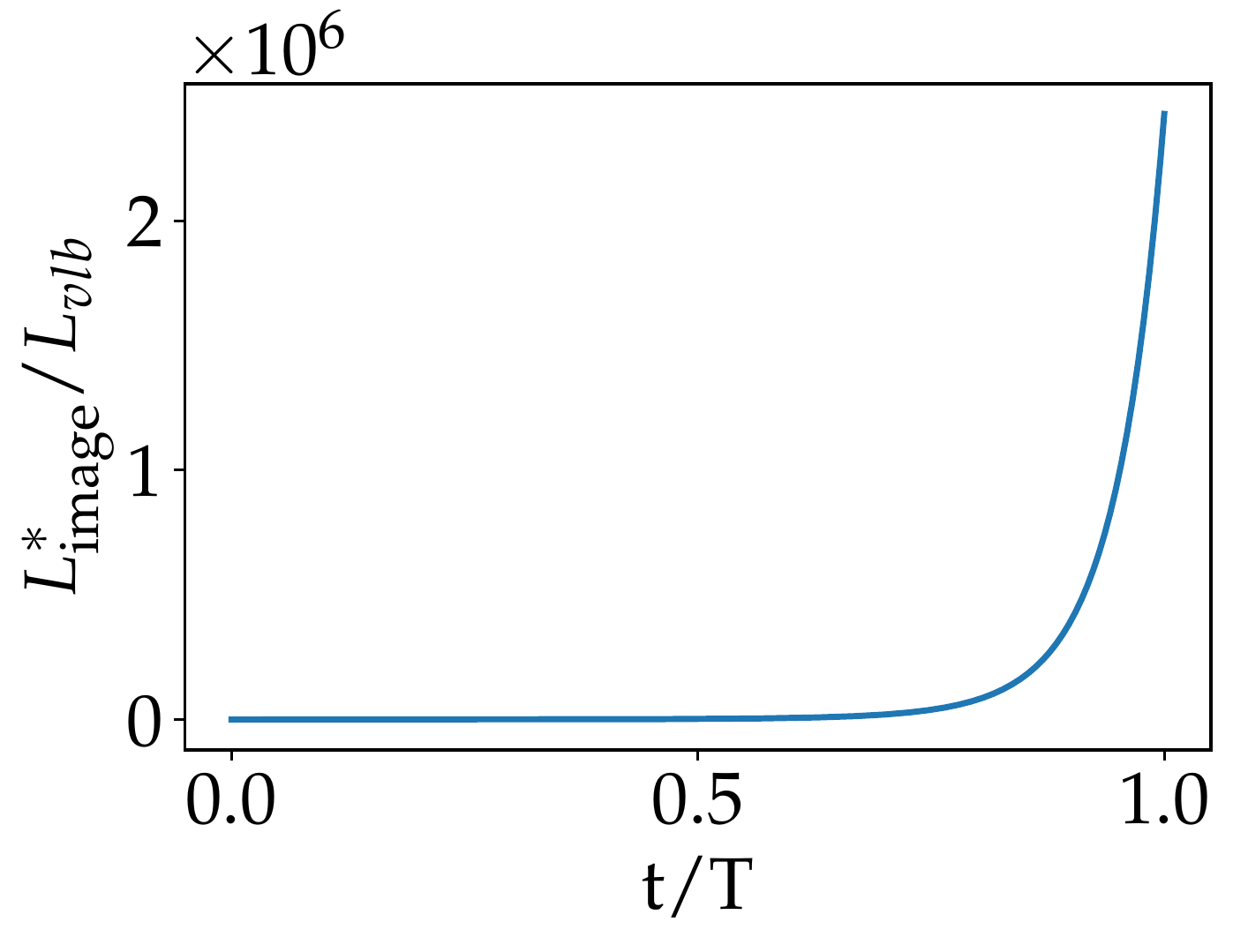}
			\subcaption{Image-parameterization}
		\end{subfigure}
		\caption[Diffusion loss reweighting]{
			\begin{minipage}[t]{0.8\textwidth}
				Weighting of the diffusion loss terms relative to the ELBO-derived loss.
			\end{minipage}
		}
		\label{fig: loss-weighting}
	\end{figure}
}

\newcommand{\FigOverviewLatentDiffusion}
{
	\begin{figure}[H]
		\centering
		\hspace{-2em}
		\includegraphics[width=\textwidth]{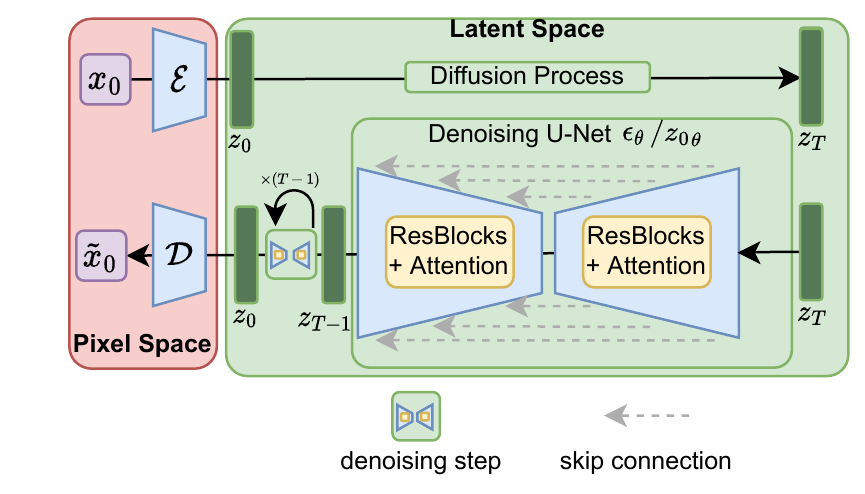}
		\caption[Overview of the LDM]{
			\begin{minipage}[t]{0.8\textwidth}
				LDM overview. A diffusion model learns the distribution of compressed continuous latents of a pretrained autoencoder. Adapted from \cite{rombach_high-resolution_2021}.
			\end{minipage}
		}
		\label{fig: overview latent diffusion}
	\end{figure}
}

\newcommand{\FigOverviewRepresentationLearning}
{
	\begin{figure}[H]
		\centering
		\includegraphics[width=0.8\textwidth]{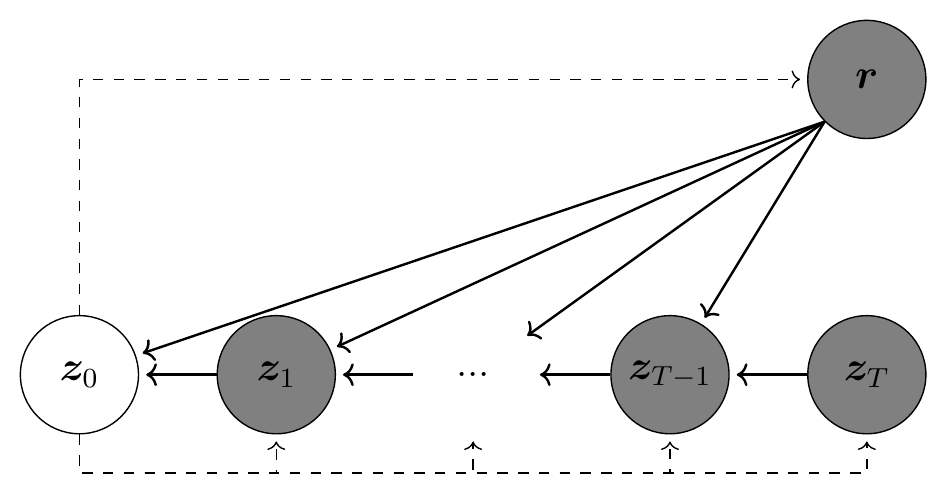}
		\caption[LRDM as probabilistic graphical model]{
			\begin{minipage}[t]{0.8\textwidth}
				Graphical model representation of a LRDM. Shaded circles denote latent variables, unshaded circles denote observable variables. Solid lines represent the generative path, dashed lines represent the encoding path.
			\end{minipage}
		}
		\label{fig: overview representation learning}
	\end{figure}
}

\newcommand{\FigOverviewTimeConditionalRepresentationLearning}
{
	\begin{figure}[H]
		\centering
		\includegraphics[width=0.8\textwidth]{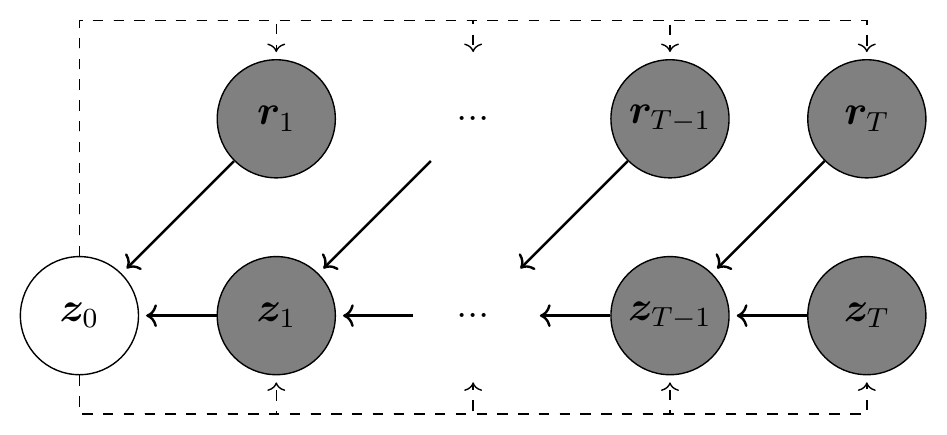}
		\caption[t-LRDM as probabilistic graphical model]{
			\begin{minipage}[t]{0.8\textwidth}
				Graphical model representation of a t-LRDM. Shaded circles denote latent variables, unshaded circles denote observable variables. Solid lines represent the generative path, dashed lines represent the encoding path.
			\end{minipage}
		}
		\label{fig: overview t-cond representation learning}
	\end{figure}
}

\newcommand{\FigRMSEoverTnoRepr}
{
	\begin{figure}[H]
		\centering
		\begin{subfigure}{0.48\textwidth}
			\includegraphics[width=\textwidth]{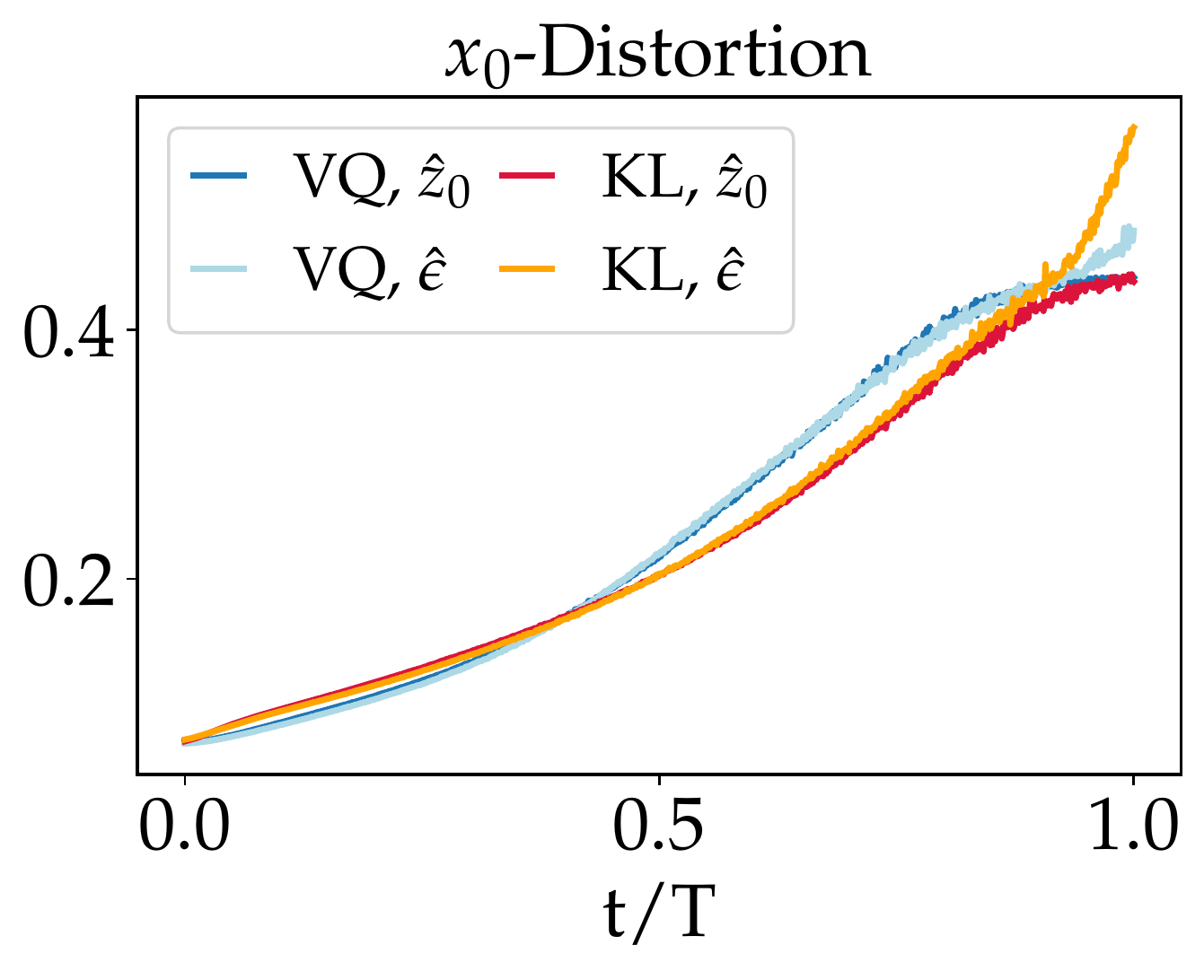}
			\subcaption{Distortion in image space for different first-stage models (VQ, KL) and parameterizations ($\hat{\z}_0$, $\hat{\bm\epsilon}$).}
			\label{fig: distortion a}
		\end{subfigure}
		\hspace{2pt}
		\begin{subfigure}{0.49\textwidth}
			\includegraphics[width=\textwidth]{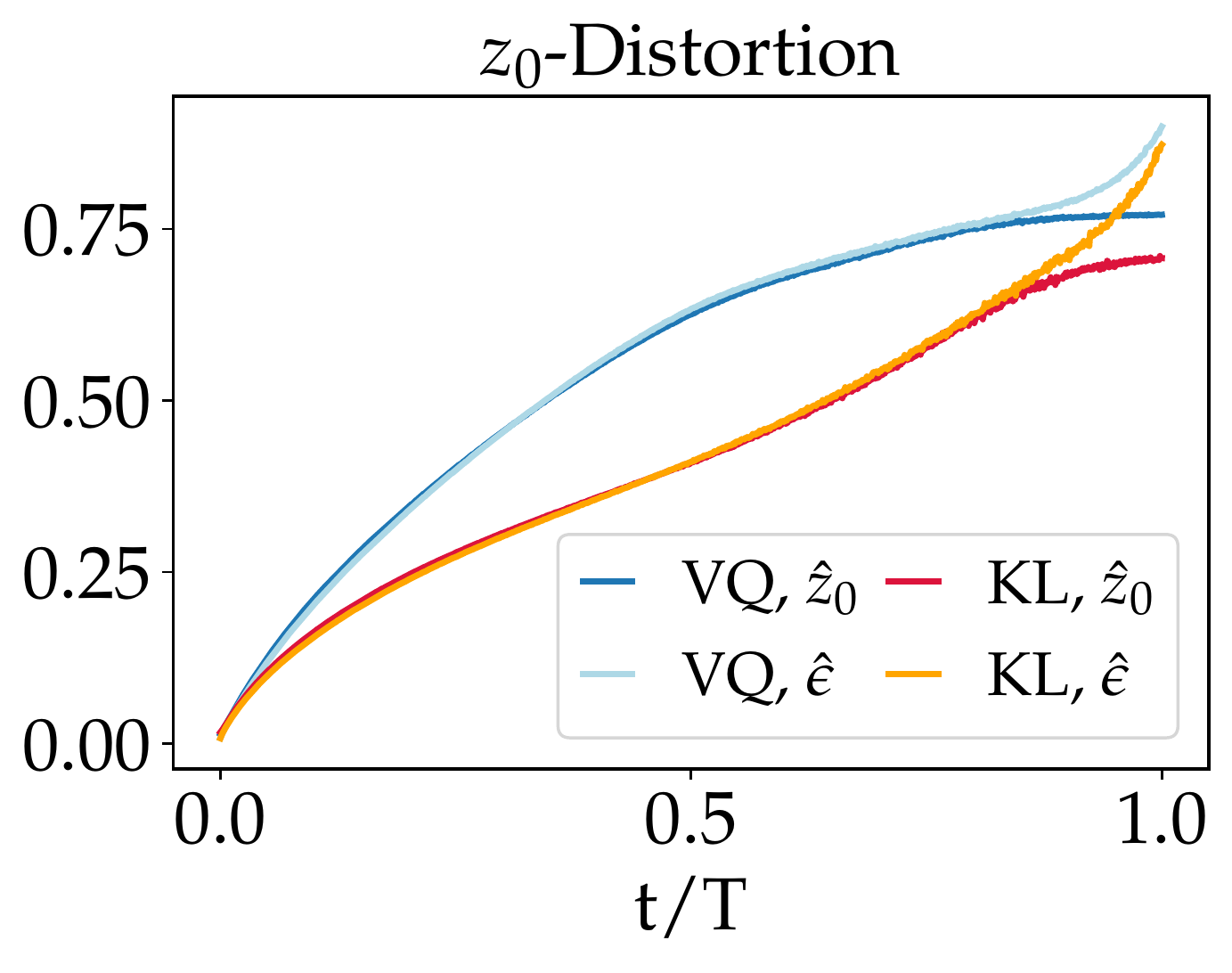}
			\subcaption{Distortion in latent space for different first-stage models (VQ, KL) and parameterizations ($\hat{\z}_0$, $\hat{\bm\epsilon}$).}
			\label{fig: distortion b}
		\end{subfigure}
		\begin{subfigure}{0.48\textwidth}
			\includegraphics[width=\textwidth]{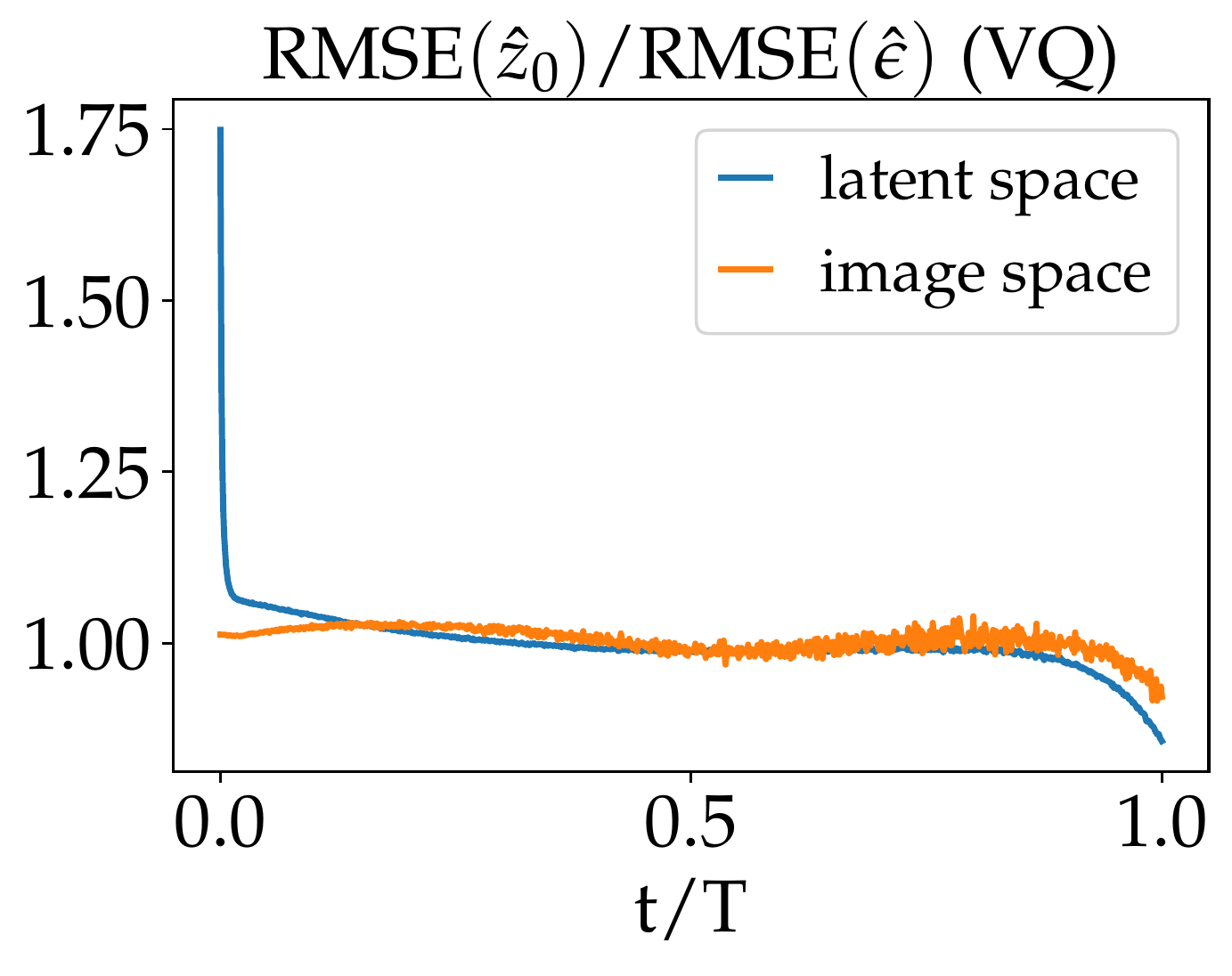}
			\subcaption{Ratio of distortions for image-parameterized and noise-parameterized models, trained on VQ-AE latents, evaluated in latent space and image space.}
			\label{fig: distortion c}
		\end{subfigure}
		\hspace{2pt}
		\begin{subfigure}{0.476\textwidth}
			\includegraphics[width=\textwidth]{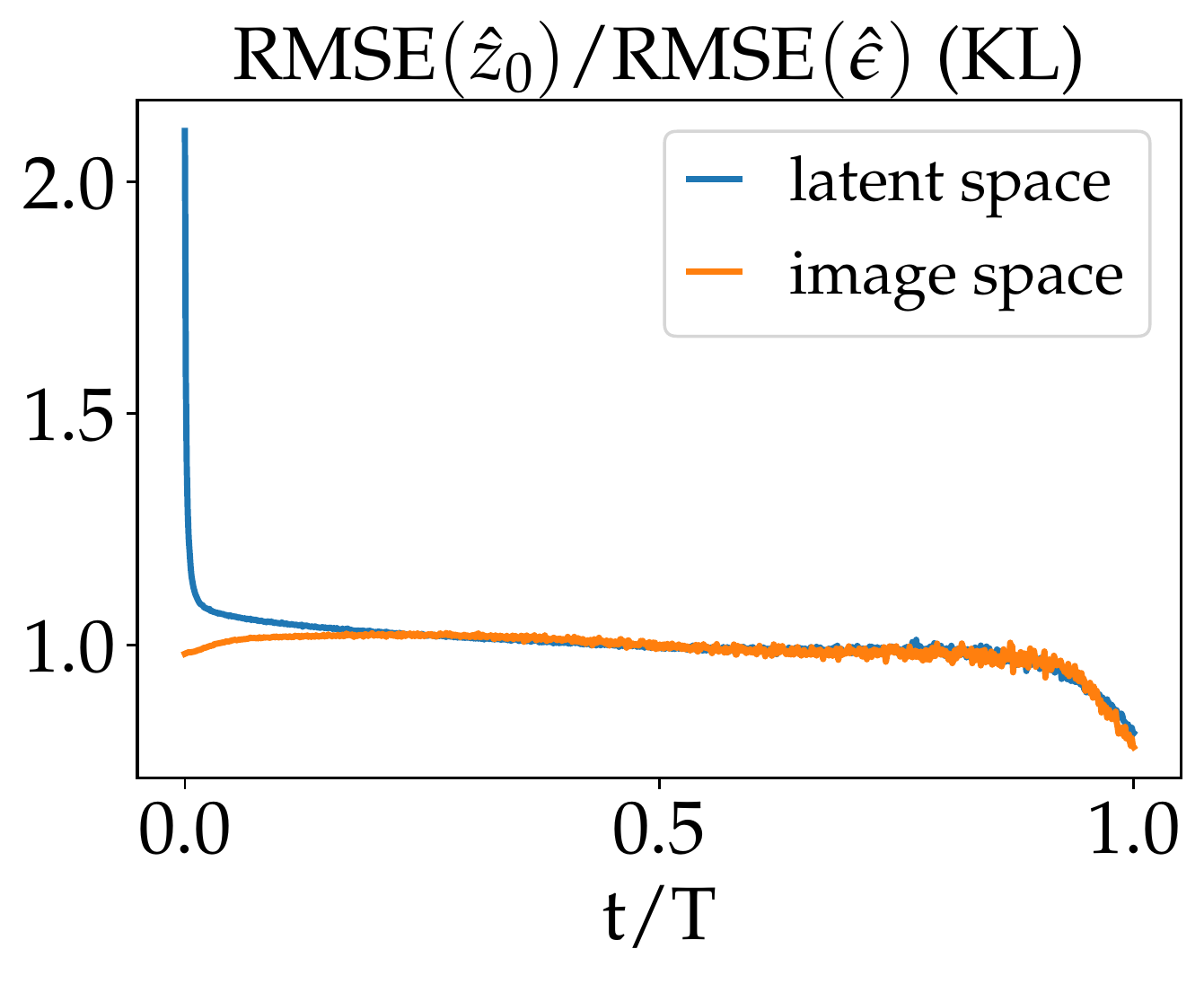}
			\subcaption{Ratio of distortions for image-parameterized and noise-parameterized models, trained on KL-AE latents, evaluated in latent space and image space.}
			\label{fig: distortion d}
		\end{subfigure}
		\caption[LDM: Image distortion]{
			\begin{minipage}[t]{0.8\textwidth}
				Analyzing the distortion (RMSE) at different timesteps in image space and in the latent space of the respective first-stage model. We observe structural differences (i) between the DMs trained on the VQ-AE latents and on those trained on KL-AE latents, and (ii) between the distortion curves in image space and latent space. See text for further discussion.
			\end{minipage}
		}
		\label{fig: eps vs x0 RMSE}
	\end{figure}
}

\newcommand{\FigSamplesLatentDiffusionEps}
{
	\begin{figure}[H]
		\centering
		
		\begin{tcbraster}[
			raster columns=10,
			raster equal height, 
			raster column skip=0pt,
			raster row skip=0pt,
			raster every box/.style={blank}
		]
			\tcbincludegraphics{FigSamplesLatentDiffusionEps/img0}
			\tcbincludegraphics{FigSamplesLatentDiffusionEps/img1}
			\tcbincludegraphics{FigSamplesLatentDiffusionEps/img2}
			\tcbincludegraphics{FigSamplesLatentDiffusionEps/img3}
			\tcbincludegraphics{FigSamplesLatentDiffusionEps/img4}
			\tcbincludegraphics{FigSamplesLatentDiffusionEps/img5}
			\tcbincludegraphics{FigSamplesLatentDiffusionEps/img6}
			\tcbincludegraphics{FigSamplesLatentDiffusionEps/img7}
			\tcbincludegraphics{FigSamplesLatentDiffusionEps/img8}
			\tcbincludegraphics{FigSamplesLatentDiffusionEps/img9}
			\tcbincludegraphics{FigSamplesLatentDiffusionEps/img10}
			\tcbincludegraphics{FigSamplesLatentDiffusionEps/img11}
			\tcbincludegraphics{FigSamplesLatentDiffusionEps/img12}
			\tcbincludegraphics{FigSamplesLatentDiffusionEps/img13}
			\tcbincludegraphics{FigSamplesLatentDiffusionEps/img14}
			\tcbincludegraphics{FigSamplesLatentDiffusionEps/img15}
			\tcbincludegraphics{FigSamplesLatentDiffusionEps/img16}
			\tcbincludegraphics{FigSamplesLatentDiffusionEps/img17}
			\tcbincludegraphics{FigSamplesLatentDiffusionEps/img18}
			\tcbincludegraphics{FigSamplesLatentDiffusionEps/img19}
			\tcbincludegraphics{FigSamplesLatentDiffusionEps/img20}
			\tcbincludegraphics{FigSamplesLatentDiffusionEps/img21}
			\tcbincludegraphics{FigSamplesLatentDiffusionEps/img22}
			\tcbincludegraphics{FigSamplesLatentDiffusionEps/img23}
			\tcbincludegraphics{FigSamplesLatentDiffusionEps/img24}
			\tcbincludegraphics{FigSamplesLatentDiffusionEps/img25}
			\tcbincludegraphics{FigSamplesLatentDiffusionEps/img26}
			\tcbincludegraphics{FigSamplesLatentDiffusionEps/img27}
			\tcbincludegraphics{FigSamplesLatentDiffusionEps/img28}
			\tcbincludegraphics{FigSamplesLatentDiffusionEps/img29}
			\tcbincludegraphics{FigSamplesLatentDiffusionEps/img30}
			\tcbincludegraphics{FigSamplesLatentDiffusionEps/img31}
			\tcbincludegraphics{FigSamplesLatentDiffusionEps/img32}
			\tcbincludegraphics{FigSamplesLatentDiffusionEps/img33}
			\tcbincludegraphics{FigSamplesLatentDiffusionEps/img34}
			\tcbincludegraphics{FigSamplesLatentDiffusionEps/img35}
			\tcbincludegraphics{FigSamplesLatentDiffusionEps/img36}
			\tcbincludegraphics{FigSamplesLatentDiffusionEps/img37}
			\tcbincludegraphics{FigSamplesLatentDiffusionEps/img38}
			\tcbincludegraphics{FigSamplesLatentDiffusionEps/img39}
			\tcbincludegraphics{FigSamplesLatentDiffusionEps/img40}
			\tcbincludegraphics{FigSamplesLatentDiffusionEps/img41}
			\tcbincludegraphics{FigSamplesLatentDiffusionEps/img42}
			\tcbincludegraphics{FigSamplesLatentDiffusionEps/img43}
			\tcbincludegraphics{FigSamplesLatentDiffusionEps/img44}
			\tcbincludegraphics{FigSamplesLatentDiffusionEps/img45}
			\tcbincludegraphics{FigSamplesLatentDiffusionEps/img46}
			\tcbincludegraphics{FigSamplesLatentDiffusionEps/img47}
			\tcbincludegraphics{FigSamplesLatentDiffusionEps/img48}
			\tcbincludegraphics{FigSamplesLatentDiffusionEps/img49}
			\tcbincludegraphics{FigSamplesLatentDiffusionEps/img50}
			\tcbincludegraphics{FigSamplesLatentDiffusionEps/img51}
			\tcbincludegraphics{FigSamplesLatentDiffusionEps/img52}
			\tcbincludegraphics{FigSamplesLatentDiffusionEps/img53}
			\tcbincludegraphics{FigSamplesLatentDiffusionEps/img54}
			\tcbincludegraphics{FigSamplesLatentDiffusionEps/img55}
			\tcbincludegraphics{FigSamplesLatentDiffusionEps/img56}
			\tcbincludegraphics{FigSamplesLatentDiffusionEps/img57}
			\tcbincludegraphics{FigSamplesLatentDiffusionEps/img58}
			\tcbincludegraphics{FigSamplesLatentDiffusionEps/img59}
			\tcbincludegraphics{FigSamplesLatentDiffusionEps/img60}
			\tcbincludegraphics{FigSamplesLatentDiffusionEps/img61}
			\tcbincludegraphics{FigSamplesLatentDiffusionEps/img62}
			\tcbincludegraphics{FigSamplesLatentDiffusionEps/img63}
			\tcbincludegraphics{FigSamplesLatentDiffusionEps/img64}
			\tcbincludegraphics{FigSamplesLatentDiffusionEps/img65}
			\tcbincludegraphics{FigSamplesLatentDiffusionEps/img66}
			\tcbincludegraphics{FigSamplesLatentDiffusionEps/img67}
			\tcbincludegraphics{FigSamplesLatentDiffusionEps/img68}
			\tcbincludegraphics{FigSamplesLatentDiffusionEps/img69}
			\tcbincludegraphics{FigSamplesLatentDiffusionEps/img70}
			\tcbincludegraphics{FigSamplesLatentDiffusionEps/img71}
			\tcbincludegraphics{FigSamplesLatentDiffusionEps/img72}
			\tcbincludegraphics{FigSamplesLatentDiffusionEps/img73}
			\tcbincludegraphics{FigSamplesLatentDiffusionEps/img74}
			\tcbincludegraphics{FigSamplesLatentDiffusionEps/img75}
			\tcbincludegraphics{FigSamplesLatentDiffusionEps/img76}
			\tcbincludegraphics{FigSamplesLatentDiffusionEps/img77}
			\tcbincludegraphics{FigSamplesLatentDiffusionEps/img78}
			\tcbincludegraphics{FigSamplesLatentDiffusionEps/img79}
			\tcbincludegraphics{FigSamplesLatentDiffusionEps/img80}
			\tcbincludegraphics{FigSamplesLatentDiffusionEps/img81}
			\tcbincludegraphics{FigSamplesLatentDiffusionEps/img82}
			\tcbincludegraphics{FigSamplesLatentDiffusionEps/img83}
			\tcbincludegraphics{FigSamplesLatentDiffusionEps/img84}
			\tcbincludegraphics{FigSamplesLatentDiffusionEps/img85}
			\tcbincludegraphics{FigSamplesLatentDiffusionEps/img86}
			\tcbincludegraphics{FigSamplesLatentDiffusionEps/img87}
			\tcbincludegraphics{FigSamplesLatentDiffusionEps/img88}
			\tcbincludegraphics{FigSamplesLatentDiffusionEps/img89}
			\tcbincludegraphics{FigSamplesLatentDiffusionEps/img90}
			\tcbincludegraphics{FigSamplesLatentDiffusionEps/img91}
			\tcbincludegraphics{FigSamplesLatentDiffusionEps/img92}
			\tcbincludegraphics{FigSamplesLatentDiffusionEps/img93}
			\tcbincludegraphics{FigSamplesLatentDiffusionEps/img94}
			\tcbincludegraphics{FigSamplesLatentDiffusionEps/img95}
			\tcbincludegraphics{FigSamplesLatentDiffusionEps/img96}
			\tcbincludegraphics{FigSamplesLatentDiffusionEps/img97}
			\tcbincludegraphics{FigSamplesLatentDiffusionEps/img98}
			\tcbincludegraphics{FigSamplesLatentDiffusionEps/img99}
		\end{tcbraster}
		
		\caption[Samples from noise-parameterized LDM]{
			\begin{minipage}[t]{0.8\textwidth}
				Unconditional samples from the noise-parameterized LDM (KL-AE) on the LSUN-Churches dataset.
			\end{minipage}
		}
		\label{fig: samples latent diffusion eps}
	\end{figure}
}

\newcommand{\FigSamplesLatentDiffusionZ}
{
	\begin{figure}[H]
		\centering
		
		\begin{tcbraster}[
			raster columns=10,
			raster equal height, 
			raster column skip=0pt,
			raster row skip=0pt,
			raster every box/.style={blank}
			]
			\tcbincludegraphics{FigSamplesLatentDiffusionZ/img0}
			\tcbincludegraphics{FigSamplesLatentDiffusionZ/img1}
			\tcbincludegraphics{FigSamplesLatentDiffusionZ/img2}
			\tcbincludegraphics{FigSamplesLatentDiffusionZ/img3}
			\tcbincludegraphics{FigSamplesLatentDiffusionZ/img4}
			\tcbincludegraphics{FigSamplesLatentDiffusionZ/img5}
			\tcbincludegraphics{FigSamplesLatentDiffusionZ/img6}
			\tcbincludegraphics{FigSamplesLatentDiffusionZ/img7}
			\tcbincludegraphics{FigSamplesLatentDiffusionZ/img8}
			\tcbincludegraphics{FigSamplesLatentDiffusionZ/img9}
			\tcbincludegraphics{FigSamplesLatentDiffusionZ/img10}
			\tcbincludegraphics{FigSamplesLatentDiffusionZ/img11}
			\tcbincludegraphics{FigSamplesLatentDiffusionZ/img12}
			\tcbincludegraphics{FigSamplesLatentDiffusionZ/img13}
			\tcbincludegraphics{FigSamplesLatentDiffusionZ/img14}
			\tcbincludegraphics{FigSamplesLatentDiffusionZ/img15}
			\tcbincludegraphics{FigSamplesLatentDiffusionZ/img16}
			\tcbincludegraphics{FigSamplesLatentDiffusionZ/img17}
			\tcbincludegraphics{FigSamplesLatentDiffusionZ/img18}
			\tcbincludegraphics{FigSamplesLatentDiffusionZ/img19}
			\tcbincludegraphics{FigSamplesLatentDiffusionZ/img20}
			\tcbincludegraphics{FigSamplesLatentDiffusionZ/img21}
			\tcbincludegraphics{FigSamplesLatentDiffusionZ/img22}
			\tcbincludegraphics{FigSamplesLatentDiffusionZ/img23}
			\tcbincludegraphics{FigSamplesLatentDiffusionZ/img24}
			\tcbincludegraphics{FigSamplesLatentDiffusionZ/img25}
			\tcbincludegraphics{FigSamplesLatentDiffusionZ/img26}
			\tcbincludegraphics{FigSamplesLatentDiffusionZ/img27}
			\tcbincludegraphics{FigSamplesLatentDiffusionZ/img28}
			\tcbincludegraphics{FigSamplesLatentDiffusionZ/img29}
			\tcbincludegraphics{FigSamplesLatentDiffusionZ/img30}
			\tcbincludegraphics{FigSamplesLatentDiffusionZ/img31}
			\tcbincludegraphics{FigSamplesLatentDiffusionZ/img32}
			\tcbincludegraphics{FigSamplesLatentDiffusionZ/img33}
			\tcbincludegraphics{FigSamplesLatentDiffusionZ/img34}
			\tcbincludegraphics{FigSamplesLatentDiffusionZ/img35}
			\tcbincludegraphics{FigSamplesLatentDiffusionZ/img36}
			\tcbincludegraphics{FigSamplesLatentDiffusionZ/img37}
			\tcbincludegraphics{FigSamplesLatentDiffusionZ/img38}
			\tcbincludegraphics{FigSamplesLatentDiffusionZ/img39}
			\tcbincludegraphics{FigSamplesLatentDiffusionZ/img40}
			\tcbincludegraphics{FigSamplesLatentDiffusionZ/img41}
			\tcbincludegraphics{FigSamplesLatentDiffusionZ/img42}
			\tcbincludegraphics{FigSamplesLatentDiffusionZ/img43}
			\tcbincludegraphics{FigSamplesLatentDiffusionZ/img44}
			\tcbincludegraphics{FigSamplesLatentDiffusionZ/img45}
			\tcbincludegraphics{FigSamplesLatentDiffusionZ/img46}
			\tcbincludegraphics{FigSamplesLatentDiffusionZ/img47}
			\tcbincludegraphics{FigSamplesLatentDiffusionZ/img48}
			\tcbincludegraphics{FigSamplesLatentDiffusionZ/img49}
			\tcbincludegraphics{FigSamplesLatentDiffusionZ/img50}
			\tcbincludegraphics{FigSamplesLatentDiffusionZ/img51}
			\tcbincludegraphics{FigSamplesLatentDiffusionZ/img52}
			\tcbincludegraphics{FigSamplesLatentDiffusionZ/img53}
			\tcbincludegraphics{FigSamplesLatentDiffusionZ/img54}
			\tcbincludegraphics{FigSamplesLatentDiffusionZ/img55}
			\tcbincludegraphics{FigSamplesLatentDiffusionZ/img56}
			\tcbincludegraphics{FigSamplesLatentDiffusionZ/img57}
			\tcbincludegraphics{FigSamplesLatentDiffusionZ/img58}
			\tcbincludegraphics{FigSamplesLatentDiffusionZ/img59}
			\tcbincludegraphics{FigSamplesLatentDiffusionZ/img60}
			\tcbincludegraphics{FigSamplesLatentDiffusionZ/img61}
			\tcbincludegraphics{FigSamplesLatentDiffusionZ/img62}
			\tcbincludegraphics{FigSamplesLatentDiffusionZ/img63}
			\tcbincludegraphics{FigSamplesLatentDiffusionZ/img64}
			\tcbincludegraphics{FigSamplesLatentDiffusionZ/img65}
			\tcbincludegraphics{FigSamplesLatentDiffusionZ/img66}
			\tcbincludegraphics{FigSamplesLatentDiffusionZ/img67}
			\tcbincludegraphics{FigSamplesLatentDiffusionZ/img68}
			\tcbincludegraphics{FigSamplesLatentDiffusionZ/img69}
			\tcbincludegraphics{FigSamplesLatentDiffusionZ/img70}
			\tcbincludegraphics{FigSamplesLatentDiffusionZ/img71}
			\tcbincludegraphics{FigSamplesLatentDiffusionZ/img72}
			\tcbincludegraphics{FigSamplesLatentDiffusionZ/img73}
			\tcbincludegraphics{FigSamplesLatentDiffusionZ/img74}
			\tcbincludegraphics{FigSamplesLatentDiffusionZ/img75}
			\tcbincludegraphics{FigSamplesLatentDiffusionZ/img76}
			\tcbincludegraphics{FigSamplesLatentDiffusionZ/img77}
			\tcbincludegraphics{FigSamplesLatentDiffusionZ/img78}
			\tcbincludegraphics{FigSamplesLatentDiffusionZ/img79}
			\tcbincludegraphics{FigSamplesLatentDiffusionZ/img80}
			\tcbincludegraphics{FigSamplesLatentDiffusionZ/img81}
			\tcbincludegraphics{FigSamplesLatentDiffusionZ/img82}
			\tcbincludegraphics{FigSamplesLatentDiffusionZ/img83}
			\tcbincludegraphics{FigSamplesLatentDiffusionZ/img84}
			\tcbincludegraphics{FigSamplesLatentDiffusionZ/img85}
			\tcbincludegraphics{FigSamplesLatentDiffusionZ/img86}
			\tcbincludegraphics{FigSamplesLatentDiffusionZ/img87}
			\tcbincludegraphics{FigSamplesLatentDiffusionZ/img88}
			\tcbincludegraphics{FigSamplesLatentDiffusionZ/img89}
			\tcbincludegraphics{FigSamplesLatentDiffusionZ/img90}
			\tcbincludegraphics{FigSamplesLatentDiffusionZ/img91}
			\tcbincludegraphics{FigSamplesLatentDiffusionZ/img92}
			\tcbincludegraphics{FigSamplesLatentDiffusionZ/img93}
			\tcbincludegraphics{FigSamplesLatentDiffusionZ/img94}
			\tcbincludegraphics{FigSamplesLatentDiffusionZ/img95}
			\tcbincludegraphics{FigSamplesLatentDiffusionZ/img96}
			\tcbincludegraphics{FigSamplesLatentDiffusionZ/img97}
			\tcbincludegraphics{FigSamplesLatentDiffusionZ/img98}
			\tcbincludegraphics{FigSamplesLatentDiffusionZ/img99}
		\end{tcbraster}
		
		\caption[Samples from image-parameterized LDM]{
			\begin{minipage}[t]{0.8\textwidth}
				Unconditional samples from the image-parameterized LDM (VQ-AE) on the LSUN-Churches dataset.
			\end{minipage}
		}
		\label{fig: samples latent diffusion z0}
	\end{figure}
}

\newcommand{\FigArchitectureRepresentationLearning}
{
	\begin{figure}[H]
		\centering
		\includegraphics[width=0.85\textwidth]{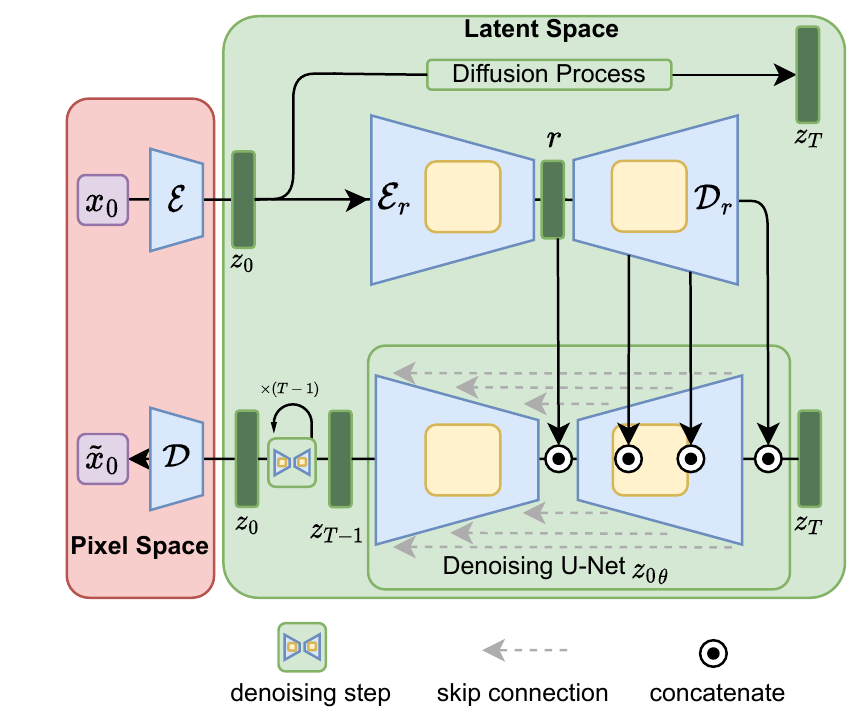}
		\caption[Overview of the LRDM]{
			\begin{minipage}[t]{0.8\textwidth}
				Overview of the LRDM architecture. The LDM is extended by a representation encoder $\mathcal{E}_r$, which extracts a spatial representation directly from the denoised latent $\z_0$. The denoising U-Net is then conditioned on the representation via concatenation on multiple spatial scales. To that end, the representation is scaled up through $\mathcal{D}_r$. Adapted from \cite{rombach_high-resolution_2021}.
			\end{minipage}
		}
		\label{fig: architecture representation learning}
	\end{figure}
}

\newcommand{\FigReconstructionsDDIMLSUNchurchDeep}
{
	\begin{figure}[H]
		\centering
		
		\begin{tabular*}{\textwidth}{@{}l @{\extracolsep{\fill}}cc}
			$\overbracket{\hspace{0.16\textwidth}}^{\mathrm{input}}$ &
			$\overbracket{\hspace{0.81\textwidth}}^{\mathrm{reconstructions}}$
		\end{tabular*}
		\begin{tcbraster}[
			raster columns=6,
			raster equal height,
			raster column skip=0pt,
			raster row skip=0pt,
			raster every box/.style={blank},
			]
			\tcbincludegraphics{FigReconstructionsLSUNchurchDeep_ddim/input1-4141}
			\tcbincludegraphics{FigReconstructionsLSUNchurchDeep_ddim/rec1-0}
			\tcbincludegraphics{FigReconstructionsLSUNchurchDeep_ddim/rec1-1}
			\tcbincludegraphics{FigReconstructionsLSUNchurchDeep_ddim/rec1-4}
			\tcbincludegraphics{FigReconstructionsLSUNchurchDeep_ddim/rec1-5}
			\tcbincludegraphics{FigReconstructionsLSUNchurchDeep_ddim/rec1-6}
			\tcbincludegraphics{FigReconstructionsLSUNchurchDeep_ddim/input2-1465}
			\tcbincludegraphics{FigReconstructionsLSUNchurchDeep_ddim/rec2-1}
			\tcbincludegraphics{FigReconstructionsLSUNchurchDeep_ddim/rec2-2}
			\tcbincludegraphics{FigReconstructionsLSUNchurchDeep_ddim/rec2-4}
			\tcbincludegraphics{FigReconstructionsLSUNchurchDeep_ddim/rec2-5}
			\tcbincludegraphics{FigReconstructionsLSUNchurchDeep_ddim/rec2-6}
			\tcbincludegraphics{FigReconstructionsLSUNchurchDeep_ddim/input7-1168}
			\tcbincludegraphics{FigReconstructionsLSUNchurchDeep_ddim/rec7-0}
			\tcbincludegraphics{FigReconstructionsLSUNchurchDeep_ddim/rec7-2}
			\tcbincludegraphics{FigReconstructionsLSUNchurchDeep_ddim/rec7-4}
			\tcbincludegraphics{FigReconstructionsLSUNchurchDeep_ddim/rec7-5}
			\tcbincludegraphics{FigReconstructionsLSUNchurchDeep_ddim/rec7-6}
			\tcbincludegraphics{FigReconstructionsLSUNchurchDeep_ddim/input8-4545}
			\tcbincludegraphics{FigReconstructionsLSUNchurchDeep_ddim/rec8-0}
			\tcbincludegraphics{FigReconstructionsLSUNchurchDeep_ddim/rec8-2}
			\tcbincludegraphics{FigReconstructionsLSUNchurchDeep_ddim/rec8-5}
			\tcbincludegraphics{FigReconstructionsLSUNchurchDeep_ddim/rec8-6}
			\tcbincludegraphics{FigReconstructionsLSUNchurchDeep_ddim/rec8-7}
			\tcbincludegraphics{FigReconstructionsLSUNchurchDeep_ddim/input9-1805}
			\tcbincludegraphics{FigReconstructionsLSUNchurchDeep_ddim/rec9-1}
			\tcbincludegraphics{FigReconstructionsLSUNchurchDeep_ddim/rec9-2}
			\tcbincludegraphics{FigReconstructionsLSUNchurchDeep_ddim/rec9-4}
			\tcbincludegraphics{FigReconstructionsLSUNchurchDeep_ddim/rec9-6}
			\tcbincludegraphics{FigReconstructionsLSUNchurchDeep_ddim/rec9-7}
			\tcbincludegraphics{FigReconstructionsLSUNchurchDeep_ddim/input10-2749}
			\tcbincludegraphics{FigReconstructionsLSUNchurchDeep_ddim/rec10-1}
			\tcbincludegraphics{FigReconstructionsLSUNchurchDeep_ddim/rec10-3}
			\tcbincludegraphics{FigReconstructionsLSUNchurchDeep_ddim/rec10-5}
			\tcbincludegraphics{FigReconstructionsLSUNchurchDeep_ddim/rec10-6}
			\tcbincludegraphics{FigReconstructionsLSUNchurchDeep_ddim/rec10-7}
		\end{tcbraster}
		
		\caption[LRDM: LSUN-Churches reconstructions]{
			\begin{minipage}[t]{0.8\textwidth}
				LRDM reconstructions from the encoded representation (mode) for randomly sampled $\z_T$, using DDIM sampling. The representation is of shape $64\times8\times8$, $\lambda=10^{-4}$, image-parameterized LRDM (VQ-AE) trained on LSUN-Churches. The representation $\rep$ captures high-level semantics and global features, changing the latent code $\z_T$ affects local details such as the shape of windows or local texture.
			\end{minipage}
		}
		\label{fig: reconstructions LSUNchurch deep DDIM}
	\end{figure}
}

\newcommand{\FigReconstructionsDDIMCelebAHQ}
{
	\begin{figure}[H]
		\centering
		\begin{tabular*}{\textwidth}{@{}l @{\extracolsep{\fill}}cc}
			$\overbracket{\hspace{0.16\textwidth}}^{\mathrm{input}}$ &
			$\overbracket{\hspace{0.81\textwidth}}^{\mathrm{reconstructions}}$
		\end{tabular*}
		\begin{tcbraster}[
			raster columns=6,
			raster equal height, 
			raster column skip=0pt,
			raster row skip=0pt,
			raster every box/.style={blank}
			]
			\tcbincludegraphics{FigReconstructionsCelebAHQ_ddim/input0-4974}
			\tcbincludegraphics{FigReconstructionsCelebAHQ_ddim/rec0-0}
			\tcbincludegraphics{FigReconstructionsCelebAHQ_ddim/rec0-1}
			\tcbincludegraphics{FigReconstructionsCelebAHQ_ddim/rec0-3}
			\tcbincludegraphics{FigReconstructionsCelebAHQ_ddim/rec0-4}
			\tcbincludegraphics{FigReconstructionsCelebAHQ_ddim/rec0-5}
			\tcbincludegraphics{FigReconstructionsCelebAHQ_ddim/input3-3348}
			\tcbincludegraphics{FigReconstructionsCelebAHQ_ddim/rec3-0}
			\tcbincludegraphics{FigReconstructionsCelebAHQ_ddim/rec3-2}
			\tcbincludegraphics{FigReconstructionsCelebAHQ_ddim/rec3-4}
			\tcbincludegraphics{FigReconstructionsCelebAHQ_ddim/rec3-5}
			\tcbincludegraphics{FigReconstructionsCelebAHQ_ddim/rec3-7}
			\tcbincludegraphics{FigReconstructionsCelebAHQ_ddim/input4-1554}
			\tcbincludegraphics{FigReconstructionsCelebAHQ_ddim/rec4-0}
			\tcbincludegraphics{FigReconstructionsCelebAHQ_ddim/rec4-1}
			\tcbincludegraphics{FigReconstructionsCelebAHQ_ddim/rec4-4}
			\tcbincludegraphics{FigReconstructionsCelebAHQ_ddim/rec4-5}
			\tcbincludegraphics{FigReconstructionsCelebAHQ_ddim/rec4-6}
			\tcbincludegraphics{FigReconstructionsCelebAHQ_ddim/input6-2041}
			\tcbincludegraphics{FigReconstructionsCelebAHQ_ddim/rec6-1}
			\tcbincludegraphics{FigReconstructionsCelebAHQ_ddim/rec6-3}
			\tcbincludegraphics{FigReconstructionsCelebAHQ_ddim/rec6-4}
			\tcbincludegraphics{FigReconstructionsCelebAHQ_ddim/rec6-5}
			\tcbincludegraphics{FigReconstructionsCelebAHQ_ddim/rec6-7}
			\tcbincludegraphics{FigReconstructionsCelebAHQ_ddim/input7-3614}
			\tcbincludegraphics{FigReconstructionsCelebAHQ_ddim/rec7-1}
			\tcbincludegraphics{FigReconstructionsCelebAHQ_ddim/rec7-2}
			\tcbincludegraphics{FigReconstructionsCelebAHQ_ddim/rec7-3}
			\tcbincludegraphics{FigReconstructionsCelebAHQ_ddim/rec7-6}
			\tcbincludegraphics{FigReconstructionsCelebAHQ_ddim/rec7-7}
		\end{tcbraster}
		
		\caption[LRDM: CelebA-HQ reconstructions]{
			\begin{minipage}[t]{0.8\textwidth}
				LRDM reconstructions from the encoded representation (mode) for randomly sampled $\z_T$, using DDIM sampling. The representation is of shape $64\times8\times8$, $\lambda=10^{-4}$, image-parameterized LRDM (VQ-AE) trained on CelebA-HQ. The representation $\rep$ captures high-level semantics and global features, changing the latent code $\z_T$ affects local details such as the eyes or mouth shape.
			\end{minipage}
		}
		\label{fig: reconstructions CelebA-HQ DDIM}
	\end{figure}
}

\newcommand{\FigProgressionRepr}
{
	\begin{figure}[H]
		\vspace{-2em}
		\centering
		\begin{tabularx}{\textwidth}{@{\extracolsep{\fill}}ccc}
			&\scriptsize \textbf{LRDM} & \\ \midrule
		\end{tabularx}
		\begin{tcbraster}[
			raster columns=10,
			raster equal height,
			raster column skip=0pt,
			raster row skip=0pt,
			raster every box/.style={blank},
		]
			\tcbincludegraphics{FigProgressionRepr/with-repr-rec/sampling0-0}
			\tcbincludegraphics{FigProgressionRepr/with-repr-rec/sampling0-1}
			\tcbincludegraphics{FigProgressionRepr/with-repr-rec/sampling0-2}
			\tcbincludegraphics{FigProgressionRepr/with-repr-rec/sampling0-3}
			\tcbincludegraphics{FigProgressionRepr/with-repr-rec/sampling0-4}
			\tcbincludegraphics{FigProgressionRepr/with-repr-rec/sampling0-5}
			\tcbincludegraphics{FigProgressionRepr/with-repr-rec/sampling0-6}
			\tcbincludegraphics{FigProgressionRepr/with-repr-rec/sampling0-7}
			\tcbincludegraphics{FigProgressionRepr/with-repr-rec/sampling0-8}
			\tcbincludegraphics{FigProgressionRepr/with-repr-rec/sampling0-9}
			\tcbincludegraphics{FigProgressionRepr/with-repr-rec/estimate0-0}
			\tcbincludegraphics{FigProgressionRepr/with-repr-rec/estimate0-1}
			\tcbincludegraphics{FigProgressionRepr/with-repr-rec/estimate0-2}
			\tcbincludegraphics{FigProgressionRepr/with-repr-rec/estimate0-3}
			\tcbincludegraphics{FigProgressionRepr/with-repr-rec/estimate0-4}
			\tcbincludegraphics{FigProgressionRepr/with-repr-rec/estimate0-5}
			\tcbincludegraphics{FigProgressionRepr/with-repr-rec/estimate0-6}
			\tcbincludegraphics{FigProgressionRepr/with-repr-rec/estimate0-7}
			\tcbincludegraphics{FigProgressionRepr/with-repr-rec/estimate0-8}
			\tcbincludegraphics{FigProgressionRepr/with-repr-rec/estimate0-9}
			\tcbincludegraphics{FigProgressionRepr/with-repr-rec2/sampling0-0}
			\tcbincludegraphics{FigProgressionRepr/with-repr-rec2/sampling0-1}
			\tcbincludegraphics{FigProgressionRepr/with-repr-rec2/sampling0-2}
			\tcbincludegraphics{FigProgressionRepr/with-repr-rec2/sampling0-3}
			\tcbincludegraphics{FigProgressionRepr/with-repr-rec2/sampling0-4}
			\tcbincludegraphics{FigProgressionRepr/with-repr-rec2/sampling0-5}
			\tcbincludegraphics{FigProgressionRepr/with-repr-rec2/sampling0-6}
			\tcbincludegraphics{FigProgressionRepr/with-repr-rec2/sampling0-7}
			\tcbincludegraphics{FigProgressionRepr/with-repr-rec2/sampling0-8}
			\tcbincludegraphics{FigProgressionRepr/with-repr-rec2/sampling0-9}
			\tcbincludegraphics{FigProgressionRepr/with-repr-rec2/estimate0-0}
			\tcbincludegraphics{FigProgressionRepr/with-repr-rec2/estimate0-1}
			\tcbincludegraphics{FigProgressionRepr/with-repr-rec2/estimate0-2}
			\tcbincludegraphics{FigProgressionRepr/with-repr-rec2/estimate0-3}
			\tcbincludegraphics{FigProgressionRepr/with-repr-rec2/estimate0-4}
			\tcbincludegraphics{FigProgressionRepr/with-repr-rec2/estimate0-5}
			\tcbincludegraphics{FigProgressionRepr/with-repr-rec2/estimate0-6}
			\tcbincludegraphics{FigProgressionRepr/with-repr-rec2/estimate0-7}
			\tcbincludegraphics{FigProgressionRepr/with-repr-rec2/estimate0-8}
			\tcbincludegraphics{FigProgressionRepr/with-repr-rec2/estimate0-9}
		\end{tcbraster}
		\begin{tcbraster}[
			raster columns=10,
			raster equal height,
			raster column skip=0pt,
			raster row skip=0pt,
			raster every box/.style={blank},
			]
			\tcbincludegraphics{FigProgressionRepr/mean}
			\tcbox{\centering\small\vspace{0.5em} dataset mean}
		\end{tcbraster}
		\begin{tabularx}{\textwidth}{@{\extracolsep{\fill}}ccc}
			&\scriptsize \textbf{LDM} & \\ \midrule
		\end{tabularx}
		\begin{tcbraster}[
			raster columns=10,
			raster equal height,
			raster column skip=0pt,
			raster row skip=0pt,
			raster every box/.style={blank},
		]
			\tcbincludegraphics{FigProgressionRepr/no-repr-sampling/sampling1-0}
			\tcbincludegraphics{FigProgressionRepr/no-repr-sampling/sampling1-1}
			\tcbincludegraphics{FigProgressionRepr/no-repr-sampling/sampling1-2}
			\tcbincludegraphics{FigProgressionRepr/no-repr-sampling/sampling1-3}
			\tcbincludegraphics{FigProgressionRepr/no-repr-sampling/sampling1-4}
			\tcbincludegraphics{FigProgressionRepr/no-repr-sampling/sampling1-5}
			\tcbincludegraphics{FigProgressionRepr/no-repr-sampling/sampling1-6}
			\tcbincludegraphics{FigProgressionRepr/no-repr-sampling/sampling1-7}
			\tcbincludegraphics{FigProgressionRepr/no-repr-sampling/sampling1-8}
			\tcbincludegraphics{FigProgressionRepr/no-repr-sampling/sampling1-9}
			\tcbincludegraphics{FigProgressionRepr/no-repr-sampling/estimate1-0}
			\tcbincludegraphics{FigProgressionRepr/no-repr-sampling/estimate1-1}
			\tcbincludegraphics{FigProgressionRepr/no-repr-sampling/estimate1-2}
			\tcbincludegraphics{FigProgressionRepr/no-repr-sampling/estimate1-3}
			\tcbincludegraphics{FigProgressionRepr/no-repr-sampling/estimate1-4}
			\tcbincludegraphics{FigProgressionRepr/no-repr-sampling/estimate1-5}
			\tcbincludegraphics{FigProgressionRepr/no-repr-sampling/estimate1-6}
			\tcbincludegraphics{FigProgressionRepr/no-repr-sampling/estimate1-7}
			\tcbincludegraphics{FigProgressionRepr/no-repr-sampling/estimate1-8}
			\tcbincludegraphics{FigProgressionRepr/no-repr-sampling/estimate1-9}
			\tcbincludegraphics{FigProgressionRepr/no-repr-sampling/sampling3-0}
			\tcbincludegraphics{FigProgressionRepr/no-repr-sampling/sampling3-1}
			\tcbincludegraphics{FigProgressionRepr/no-repr-sampling/sampling3-2}
			\tcbincludegraphics{FigProgressionRepr/no-repr-sampling/sampling3-3}
			\tcbincludegraphics{FigProgressionRepr/no-repr-sampling/sampling3-4}
			\tcbincludegraphics{FigProgressionRepr/no-repr-sampling/sampling3-5}
			\tcbincludegraphics{FigProgressionRepr/no-repr-sampling/sampling3-6}
			\tcbincludegraphics{FigProgressionRepr/no-repr-sampling/sampling3-7}
			\tcbincludegraphics{FigProgressionRepr/no-repr-sampling/sampling3-8}
			\tcbincludegraphics{FigProgressionRepr/no-repr-sampling/sampling3-9}
			\tcbincludegraphics{FigProgressionRepr/no-repr-sampling/estimate3-0}
			\tcbincludegraphics{FigProgressionRepr/no-repr-sampling/estimate3-1}
			\tcbincludegraphics{FigProgressionRepr/no-repr-sampling/estimate3-2}
			\tcbincludegraphics{FigProgressionRepr/no-repr-sampling/estimate3-3}
			\tcbincludegraphics{FigProgressionRepr/no-repr-sampling/estimate3-4}
			\tcbincludegraphics{FigProgressionRepr/no-repr-sampling/estimate3-5}
			\tcbincludegraphics{FigProgressionRepr/no-repr-sampling/estimate3-6}
			\tcbincludegraphics{FigProgressionRepr/no-repr-sampling/estimate3-7}
			\tcbincludegraphics{FigProgressionRepr/no-repr-sampling/estimate3-8}
			\tcbincludegraphics{FigProgressionRepr/no-repr-sampling/estimate3-9}
		\end{tcbraster}
		\caption[Sampling progression for LDM and LRDM]{
			\begin{minipage}[t]{0.8\textwidth}
				Sampling progressions, $\hat{\x}_t$ (above) and $\hat{\x}_0$ (below) over time from left ($t=T$) to right ($t=0$). Top rows: LRDM sampling. Bottom rows: LDM sampling. Center row: Mean of the LSUN-Churches training dataset. For the LDM, $\hat{\x}_0$ gradually changes from close to the dataset mean towards the final sample. The progression for the LRDM exhibits an almost constant $\hat{\x}_0$ over a certain initial timestep range, which already contains the overall shape and course features. This matches with the plateau observed in \fig{fig: rmse repr vs no repr}.
			\end{minipage}
		}
		\label{fig: progression plots repr vs no repr}
	\end{figure}
}

\newcommand{\FigInterpolationsDDIMReprVsVQGAN}
{
	\begin{figure}[H]
		\centering
		
		\begin{tabular*}{\textwidth}{@{\extracolsep{\fill}}ccc}
			$\overbracket{\hspace{0.09\textwidth}}^{\mathrm{input\,1}}$ &
			$\overbracket{\hspace{0.755\textwidth}}^{\mathrm{interpolations}}$ &
			$\overbracket{\hspace{0.09\textwidth}}^{\mathrm{input\,2}}$
		\end{tabular*}
		\begin{tabularx}{\textwidth}{@{\extracolsep{\fill}}ccc}
			&\scriptsize \textbf{VQ-AE} & \\ \midrule
		\end{tabularx}
		\begin{tcbraster}[
			raster columns=10,
			raster equal height,
			raster column skip=0pt,
			raster row skip=0pt,
			raster every box/.style={blank},
			]
			\tcbincludegraphics{FigInterpolationsReprVsVQGAN/s1}
			\tcbincludegraphics{FigInterpolationsReprVsVQGAN/int1-0}
			\tcbincludegraphics{FigInterpolationsReprVsVQGAN/int1-2}
			\tcbincludegraphics{FigInterpolationsReprVsVQGAN/int1-4}
			\tcbincludegraphics{FigInterpolationsReprVsVQGAN/int1-6}
			\tcbincludegraphics{FigInterpolationsReprVsVQGAN/int1-8}
			\tcbincludegraphics{FigInterpolationsReprVsVQGAN/int1-10}
			\tcbincludegraphics{FigInterpolationsReprVsVQGAN/int1-12}
			\tcbincludegraphics{FigInterpolationsReprVsVQGAN/int1-14}
			\tcbincludegraphics{FigInterpolationsReprVsVQGAN/t1}
			\tcbincludegraphics{FigInterpolationsReprVsVQGAN/s0}
			\tcbincludegraphics{FigInterpolationsReprVsVQGAN/int0-0}
			\tcbincludegraphics{FigInterpolationsReprVsVQGAN/int0-2}
			\tcbincludegraphics{FigInterpolationsReprVsVQGAN/int0-4}
			\tcbincludegraphics{FigInterpolationsReprVsVQGAN/int0-6}
			\tcbincludegraphics{FigInterpolationsReprVsVQGAN/int0-8}
			\tcbincludegraphics{FigInterpolationsReprVsVQGAN/int0-10}
			\tcbincludegraphics{FigInterpolationsReprVsVQGAN/int0-12}
			\tcbincludegraphics{FigInterpolationsReprVsVQGAN/int0-14}
			\tcbincludegraphics{FigInterpolationsReprVsVQGAN/t0}
		\end{tcbraster}
		\begin{tabularx}{\textwidth}{@{\extracolsep{\fill}}ccc}
			&\scriptsize \textbf{DDIM} & \\ \midrule
		\end{tabularx}
		\begin{tcbraster}[
			raster columns=10,
			raster equal height,
			raster column skip=0pt,
			raster row skip=0pt,
			raster every box/.style={blank},
			]
			\tcbincludegraphics{FigInterpolationsReprVsVQGAN_ddim_true/e400/s0-3724}
			\tcbincludegraphics{FigInterpolationsReprVsVQGAN_ddim_true/e400/int0-0}
			\tcbincludegraphics{FigInterpolationsReprVsVQGAN_ddim_true/e400/int0-2}
			\tcbincludegraphics{FigInterpolationsReprVsVQGAN_ddim_true/e400/int0-5}
			\tcbincludegraphics{FigInterpolationsReprVsVQGAN_ddim_true/e400/int0-6}
			\tcbincludegraphics{FigInterpolationsReprVsVQGAN_ddim_true/e400/int0-9}
			\tcbincludegraphics{FigInterpolationsReprVsVQGAN_ddim_true/e400/int0-10}
			\tcbincludegraphics{FigInterpolationsReprVsVQGAN_ddim_true/e400/int0-12}
			\tcbincludegraphics{FigInterpolationsReprVsVQGAN_ddim_true/e400/int0-14}
			\tcbincludegraphics{FigInterpolationsReprVsVQGAN_ddim_true/e400/t0-4297}
			\tcbincludegraphics{FigInterpolationsReprVsVQGAN_ddim_true/e400/s1-1891}
			\tcbincludegraphics{FigInterpolationsReprVsVQGAN_ddim_true/e400/int1-0}
			\tcbincludegraphics{FigInterpolationsReprVsVQGAN_ddim_true/e400/int1-2}
			\tcbincludegraphics{FigInterpolationsReprVsVQGAN_ddim_true/e400/int1-4}
			\tcbincludegraphics{FigInterpolationsReprVsVQGAN_ddim_true/e400/int1-6}
			\tcbincludegraphics{FigInterpolationsReprVsVQGAN_ddim_true/e400/int1-8}
			\tcbincludegraphics{FigInterpolationsReprVsVQGAN_ddim_true/e400/int1-10}
			\tcbincludegraphics{FigInterpolationsReprVsVQGAN_ddim_true/e400/int1-12}
			\tcbincludegraphics{FigInterpolationsReprVsVQGAN_ddim_true/e400/int1-14}
			\tcbincludegraphics{FigInterpolationsReprVsVQGAN_ddim_true/e400/t1-1256}
		\end{tcbraster}
		\begin{tabularx}{\textwidth}{@{\extracolsep{\fill}}ccc}
			&\scriptsize \textbf{LRDM} & \\ \midrule
		\end{tabularx}
		\begin{tcbraster}[
			raster columns=10,
			raster equal height,
			raster column skip=0pt,
			raster row skip=0pt,
			raster every box/.style={blank},
			]
			\tcbincludegraphics{FigInterpolationsReprVsVQGAN_ddim/s0-3724}
			\tcbincludegraphics{FigInterpolationsReprVsVQGAN_ddim/int0-0}
			\tcbincludegraphics{FigInterpolationsReprVsVQGAN_ddim/int0-2}
			\tcbincludegraphics{FigInterpolationsReprVsVQGAN_ddim/int0-5}
			\tcbincludegraphics{FigInterpolationsReprVsVQGAN_ddim/int0-6}
			\tcbincludegraphics{FigInterpolationsReprVsVQGAN_ddim/int0-9}
			\tcbincludegraphics{FigInterpolationsReprVsVQGAN_ddim/int0-10}
			\tcbincludegraphics{FigInterpolationsReprVsVQGAN_ddim/int0-12}
			\tcbincludegraphics{FigInterpolationsReprVsVQGAN_ddim/int0-14}
			\tcbincludegraphics{FigInterpolationsReprVsVQGAN_ddim/t0-4297}
			\tcbincludegraphics{FigInterpolationsReprVsVQGAN_ddim/s1-1891}
			\tcbincludegraphics{FigInterpolationsReprVsVQGAN_ddim/int1-0}
			\tcbincludegraphics{FigInterpolationsReprVsVQGAN_ddim/int1-2}
			\tcbincludegraphics{FigInterpolationsReprVsVQGAN_ddim/int1-4}
			\tcbincludegraphics{FigInterpolationsReprVsVQGAN_ddim/int1-6}
			\tcbincludegraphics{FigInterpolationsReprVsVQGAN_ddim/int1-8}
			\tcbincludegraphics{FigInterpolationsReprVsVQGAN_ddim/int1-10}
			\tcbincludegraphics{FigInterpolationsReprVsVQGAN_ddim/int1-12}
			\tcbincludegraphics{FigInterpolationsReprVsVQGAN_ddim/int1-14}
			\tcbincludegraphics{FigInterpolationsReprVsVQGAN_ddim/t1-1256}
		\end{tcbraster}
		\caption[LRDM: LSUN-Churches interpolations]{
			\begin{minipage}[t]{0.8\textwidth}
				Slerp interpolation results.
				Top rows: Interpolating between two input images in the (pre-quantization) latent space of the VQ-AE.
				Middle rows: Interpolating $\z_T$ using the DDIM forward and reverse process (LDM with VQ-AE, trained for 400 epochs).
				Bottom rows: Interpolating between two input images in the representation space of a LRDM (trained for 100 epochs, representation shape: $64\times8\times8$, VQ-AE, $\lambda=10^{-4}$, DDIM sampling, LSUN-Churches). The LRDM achieves smooth semantic interpolations.
			\end{minipage}
		}
		\label{fig: interpolations LSUNchurch deep repr vs VQGAN DDIM}
	\end{figure}
}

\newcommand{\FigInterpolationsDDIMCelebAHQ}
{
	\begin{figure}[H]
		\centering
		
		\begin{tabular*}{\textwidth}{@{\extracolsep{\fill}}ccc}
			$\overbracket{\hspace{0.09\textwidth}}^{\mathrm{input\,1}}$ &
			$\overbracket{\hspace{0.755\textwidth}}^{\mathrm{interpolations}}$ &
			$\overbracket{\hspace{0.09\textwidth}}^{\mathrm{input\,2}}$
		\end{tabular*}
		\begin{tcbraster}[
			raster columns=10,
			raster equal height,
			raster column skip=0pt,
			raster row skip=0pt,
			raster every box/.style={blank},
			]
			\tcbincludegraphics{FigInterpolationsCelebAHQ_ddim/s3-3246}
			\tcbincludegraphics{FigInterpolationsCelebAHQ_ddim/int3-0}
			\tcbincludegraphics{FigInterpolationsCelebAHQ_ddim/int3-2}
			\tcbincludegraphics{FigInterpolationsCelebAHQ_ddim/int3-4}
			\tcbincludegraphics{FigInterpolationsCelebAHQ_ddim/int3-7}
			\tcbincludegraphics{FigInterpolationsCelebAHQ_ddim/int3-9}
			\tcbincludegraphics{FigInterpolationsCelebAHQ_ddim/int3-10}
			\tcbincludegraphics{FigInterpolationsCelebAHQ_ddim/int3-12}
			\tcbincludegraphics{FigInterpolationsCelebAHQ_ddim/int3-14}
			\tcbincludegraphics{FigInterpolationsCelebAHQ_ddim/t3-1829}
			\tcbincludegraphics{FigInterpolationsCelebAHQ_ddim/s1-260}
			\tcbincludegraphics{FigInterpolationsCelebAHQ_ddim/int1-0}
			\tcbincludegraphics{FigInterpolationsCelebAHQ_ddim/int1-2}
			\tcbincludegraphics{FigInterpolationsCelebAHQ_ddim/int1-4}
			\tcbincludegraphics{FigInterpolationsCelebAHQ_ddim/int1-6}
			\tcbincludegraphics{FigInterpolationsCelebAHQ_ddim/int1-8}
			\tcbincludegraphics{FigInterpolationsCelebAHQ_ddim/int1-10}
			\tcbincludegraphics{FigInterpolationsCelebAHQ_ddim/int1-12}
			\tcbincludegraphics{FigInterpolationsCelebAHQ_ddim/int1-14}
			\tcbincludegraphics{FigInterpolationsCelebAHQ_ddim/t1-3632}
			\tcbincludegraphics{FigInterpolationsCelebAHQ_ddim/s0-1035}
			\tcbincludegraphics{FigInterpolationsCelebAHQ_ddim/int0-0}
			\tcbincludegraphics{FigInterpolationsCelebAHQ_ddim/int0-2}
			\tcbincludegraphics{FigInterpolationsCelebAHQ_ddim/int0-4}
			\tcbincludegraphics{FigInterpolationsCelebAHQ_ddim/int0-6}
			\tcbincludegraphics{FigInterpolationsCelebAHQ_ddim/int0-8}
			\tcbincludegraphics{FigInterpolationsCelebAHQ_ddim/int0-10}
			\tcbincludegraphics{FigInterpolationsCelebAHQ_ddim/int0-12}
			\tcbincludegraphics{FigInterpolationsCelebAHQ_ddim/int0-15}
			\tcbincludegraphics{FigInterpolationsCelebAHQ_ddim/t0-3382}
		\end{tcbraster}
		
		\caption[LRDM: CelebA-HQ interpolations]{
			\begin{minipage}[t]{0.8\textwidth}
				Slerp interpolations between two input images in the representation space of a LRDM (representation shape: $64\times8\times8$, VQ-AE, $\lambda=10^{-4}$, DDIM sampling, CelebA-HQ). The LRDM achieves smooth transformations of head pose and facial attributes.
			\end{minipage}
		}
		\label{fig: interpolations CelebA-HQ DDIM}
	\end{figure}
}

\newcommand{\FigRMSEreprVSnorepr}
{
	\begin{figure}[H]
		\centering
		\begin{subfigure}{0.5\textwidth}
			\includegraphics[width=\textwidth]{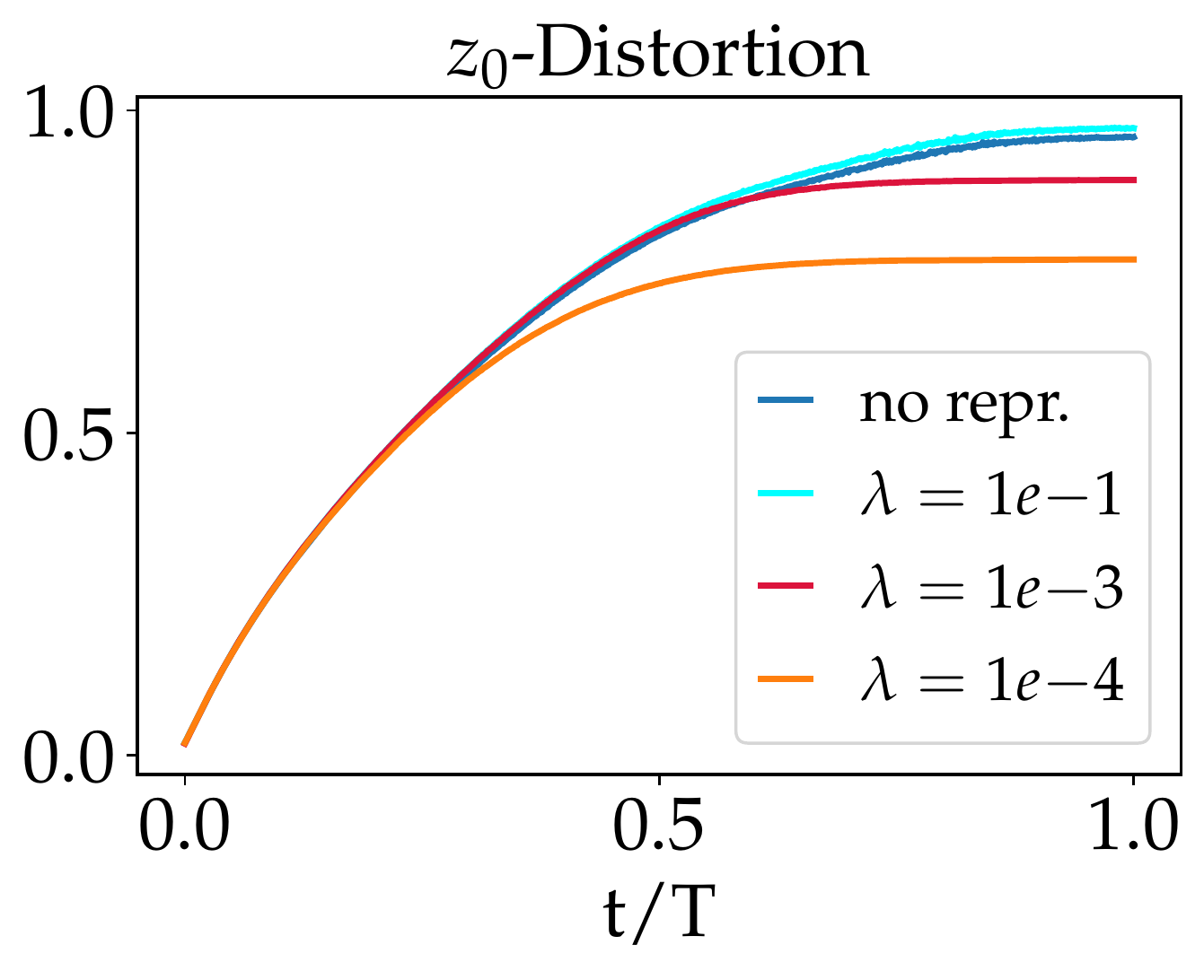}
		\end{subfigure}
		\caption[LRDM: Image distortion for different $\lambda$]{
			\begin{minipage}[t]{0.8\textwidth}
				$\z_0$-distortion (RMSE) curve for the LRDM for varying regularization strength. The distortion plateau at high $t$ stretches for weaker regularization. The plateauing distortion matches the observations made in \fig{fig: progression plots repr vs no repr}.
			\end{minipage}
		}
		\label{fig: rmse repr vs no repr}
	\end{figure}
}

\newcommand{\FigFromVAEtoDDPM}
{
	\newgeometry{top=4cm}
	\begin{figure}[H]
		\centering
		\begin{subfigure}[b]{0.49\textwidth}
			\includegraphics[width=\textwidth]{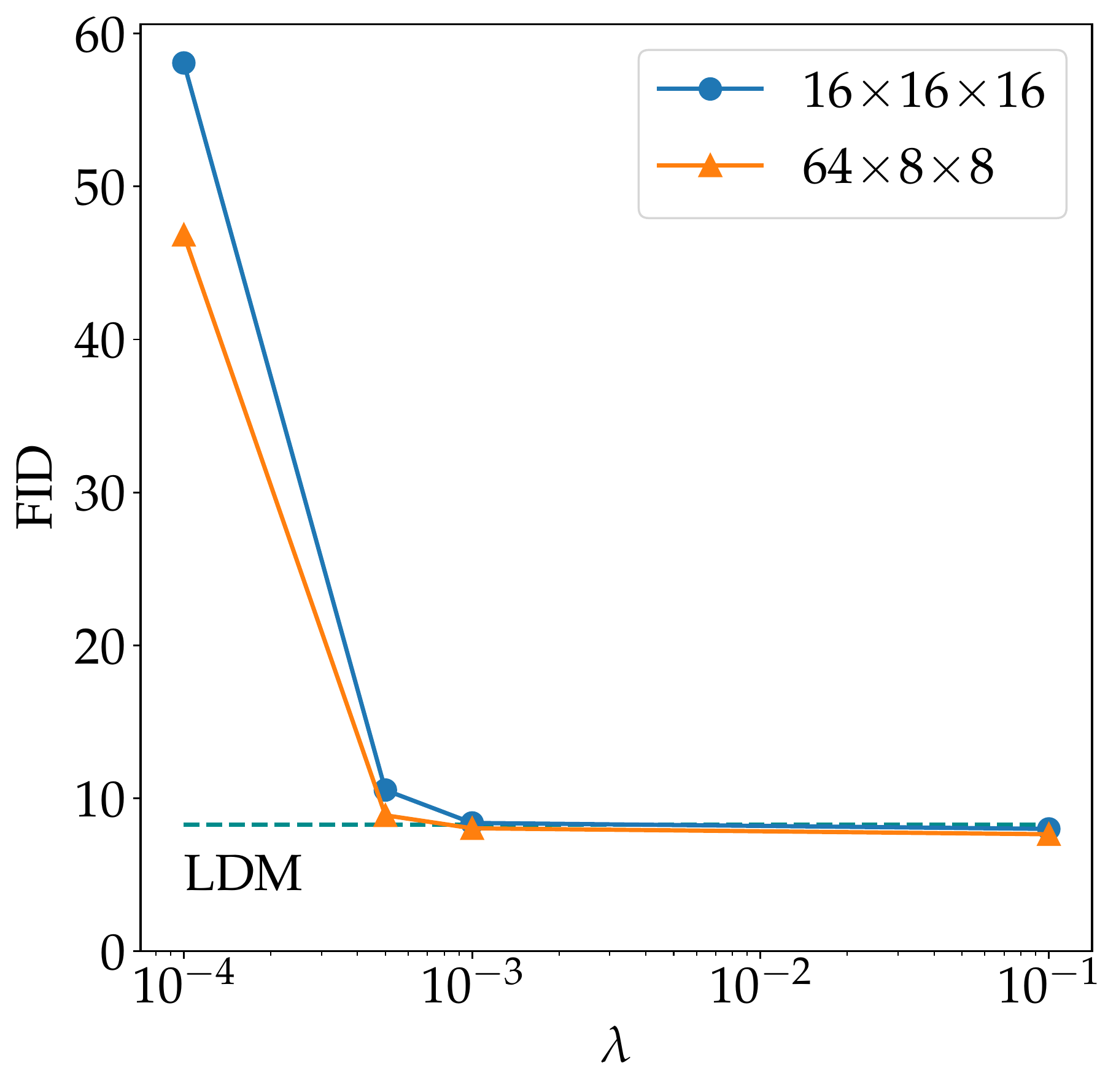}
			\subcaption{FID scores for unconditional sampling. For reference, the FID for the LDM is shown.}
		\end{subfigure}
		\begin{subfigure}[b]{0.49\textwidth}
			\includegraphics[width=\textwidth]{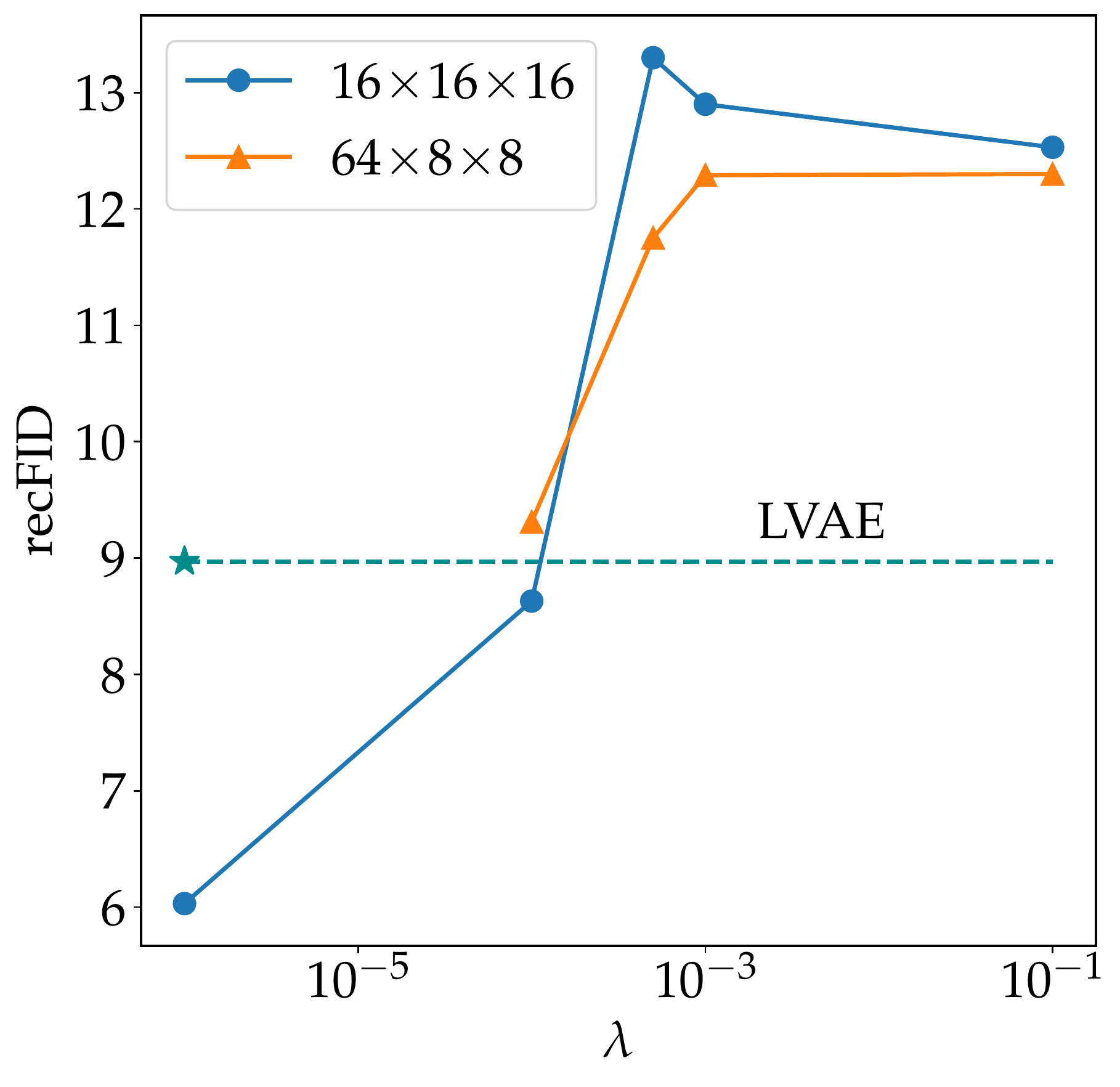}
			\subcaption{Reconstruction FID scores evaluated on the validation dataset.}
		\end{subfigure}
		\begin{subfigure}[b]{0.49\textwidth}
			\includegraphics[width=\textwidth]{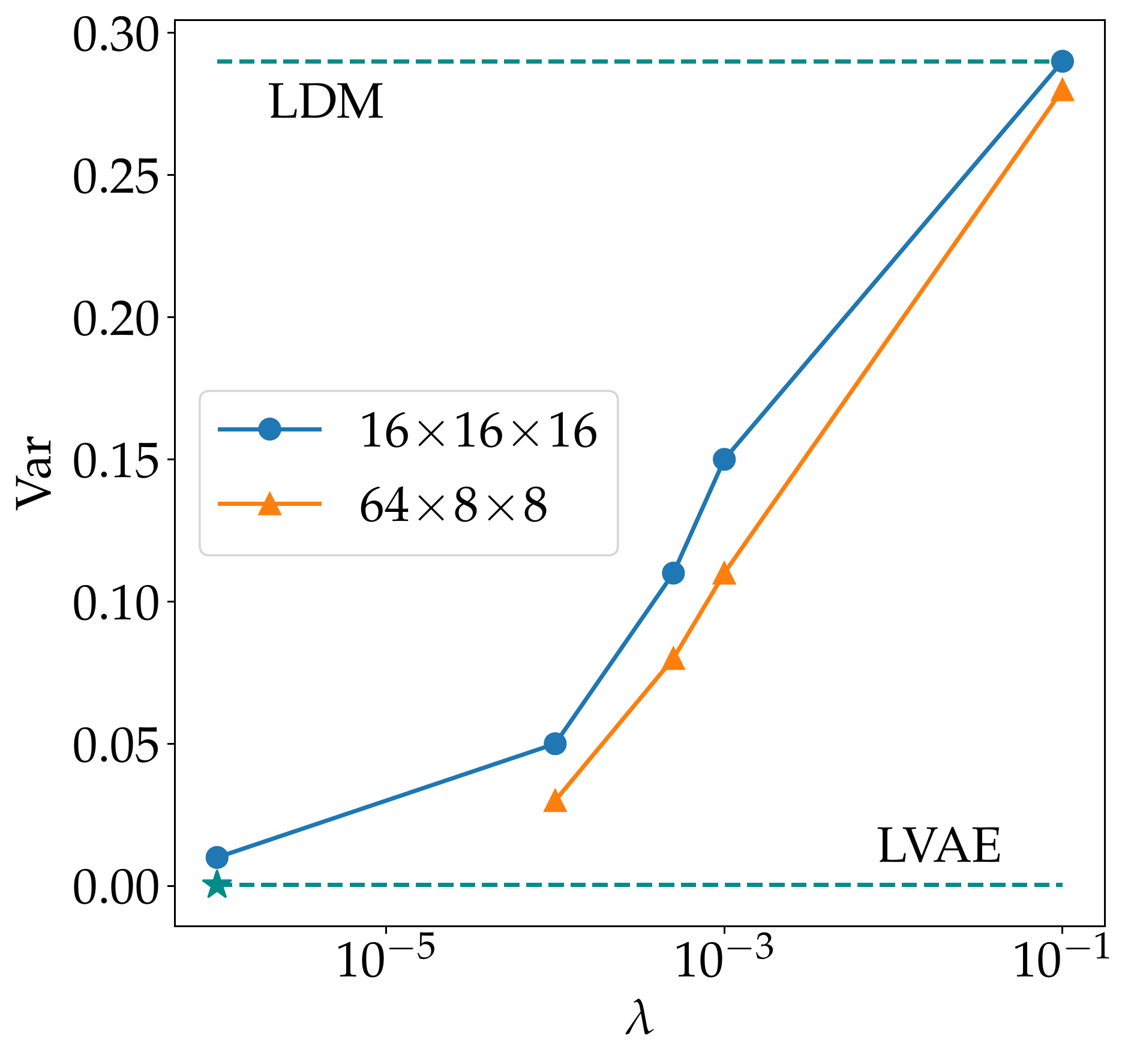}
			\subcaption{Pixel-wise variance of the image reconstructions. For reference, the reconstruction variance of the LVAE and the sampling variance of the LDM are shown.}
		\end{subfigure}
		\begin{subfigure}[b]{0.49\textwidth}
			\includegraphics[width=\textwidth]{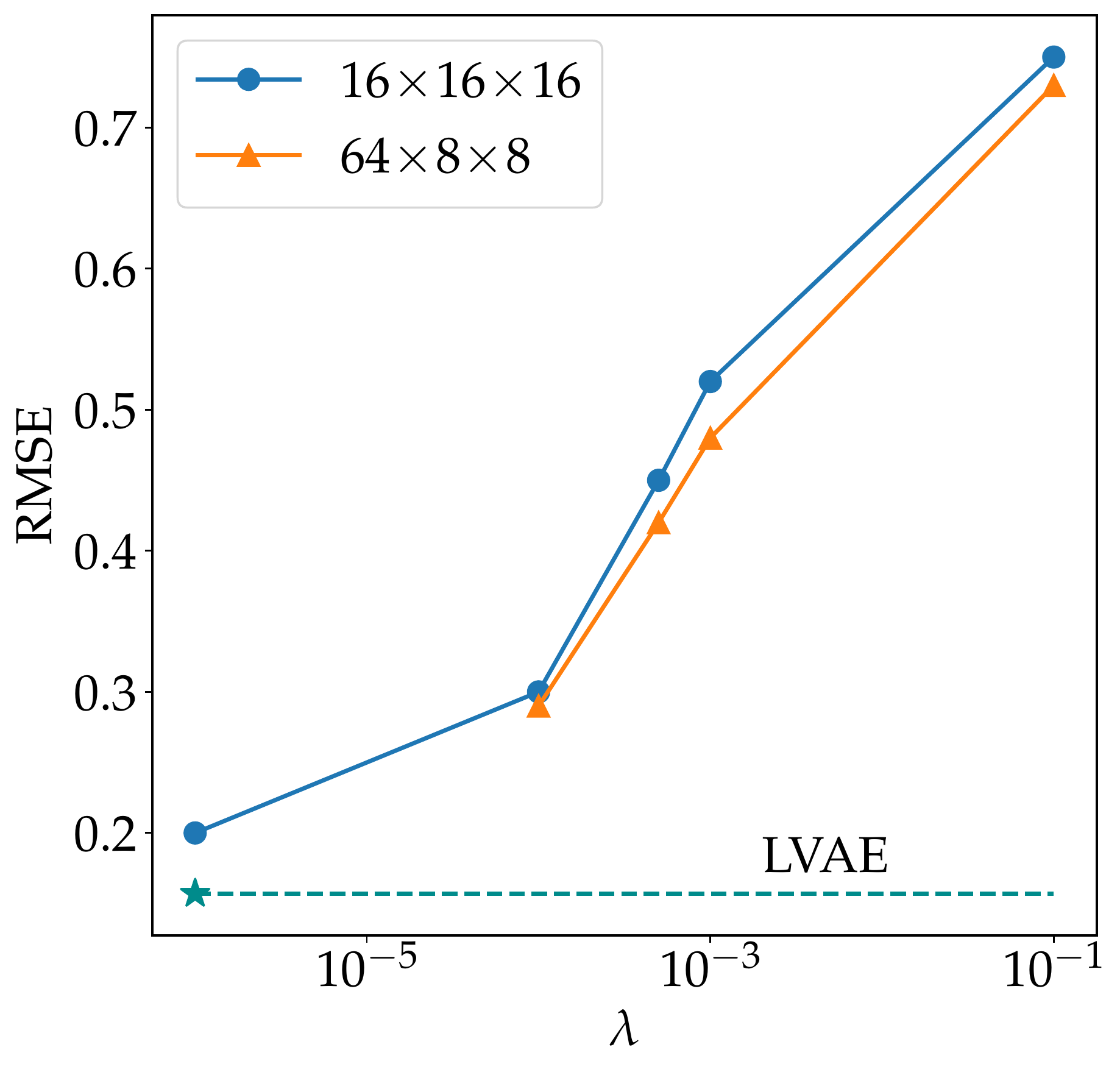}
			\subcaption{Reconstruction RMSE (pixel-wise).}
		\end{subfigure}
		\caption[]{
			\begin{minipage}[t]{0.8\textwidth}
				LRDM sampling and reconstruction quality over representation regularization $\lambda$ for different representation dimensionalities, on LSUN-Churches (values from \tab{tab: repr metrics}).
				Changing $\lambda$ interpolates, to some extent, between the LDM (when narrowing down the representation bottleneck) and a latent VAE (towards low $\lambda$).
			\end{minipage}
		}
		\label{fig: lambda sweep stats}
	\end{figure}
	\restoregeometry
}

\newcommand{\FigTcondReprRMSE}
{
	\begin{figure}[H]
		\centering
		\begin{subfigure}{0.48\textwidth}
			\includegraphics[width=\textwidth]{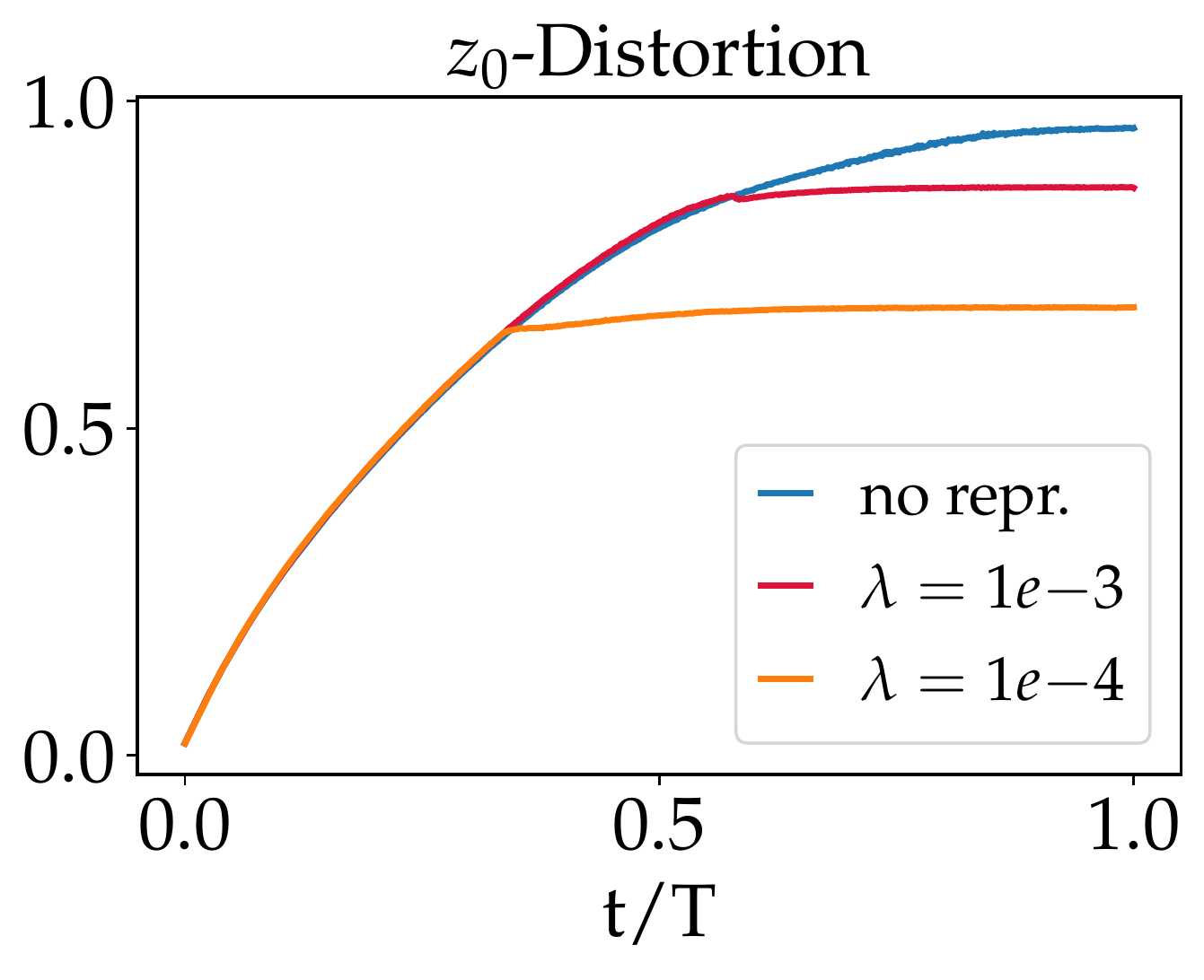}
		\end{subfigure}
		\begin{subfigure}{0.48\textwidth}
			\includegraphics[width=\textwidth]{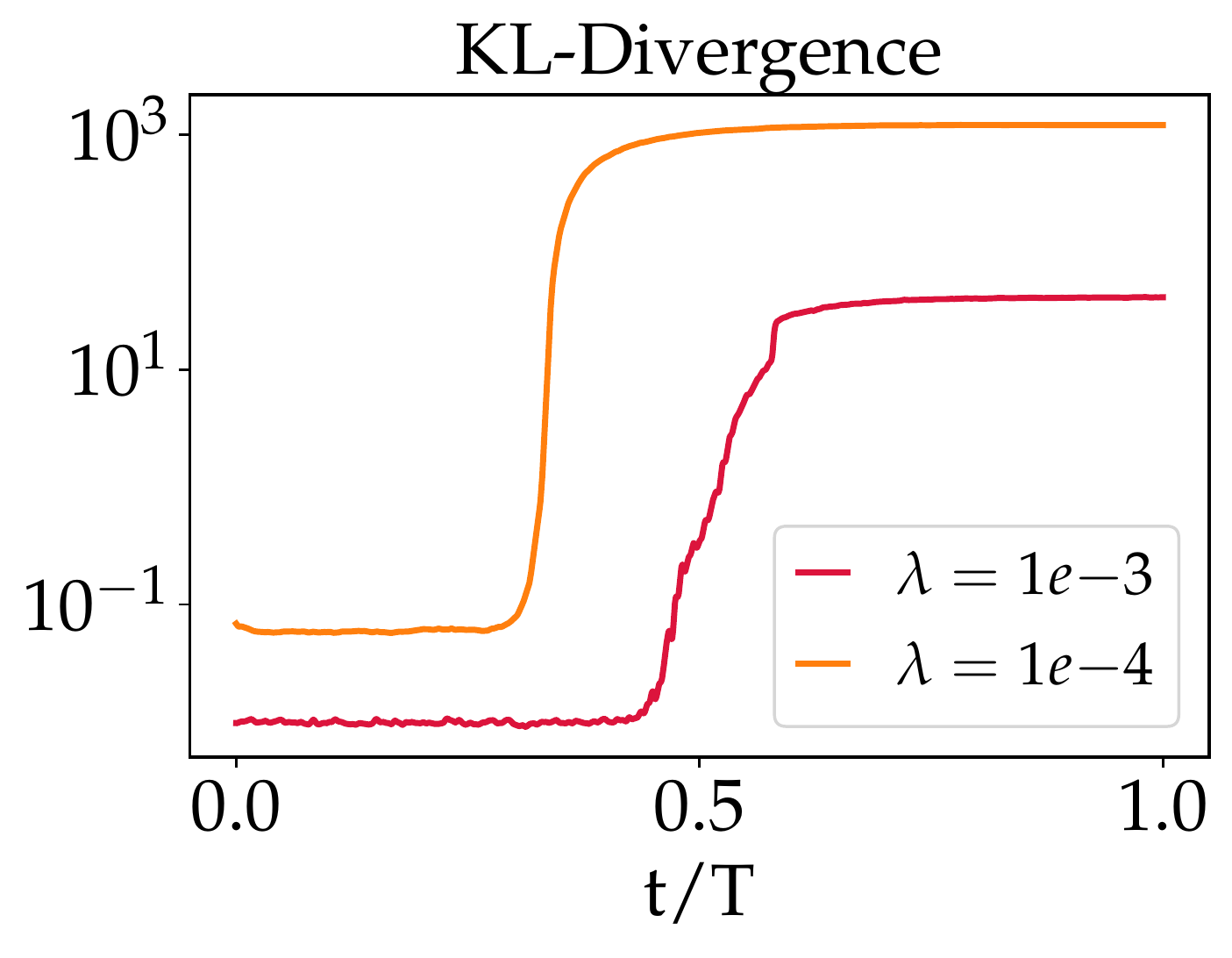}
		\end{subfigure}
		\caption[t-LRDM results]{
			\begin{minipage}[t]{0.8\textwidth}
				Results for the t-LRDM with representation shape $16\times16\times16$. Left: $\z_0$-distortion (RMSE) over $t$. Right: KL-Divergence of the encoded representation distribution of $\rep_t$ and the gaussian prior. The KL-Divergence exhibits a sharp transition depending on $\lambda$. For low $t$, the representation is almost unused, for high $t$, the distortion plateau (see \fig{fig: rmse repr vs no repr}) sharpens.
			\end{minipage}
		}
		\label{fig: rmse t-cond repr}
	\end{figure}
}

\newcommand{\FigMNIST}
{
	\begin{figure}[H]
		\centering
		\begin{subfigure}{0.48\textwidth}
			\includegraphics[width=\textwidth]{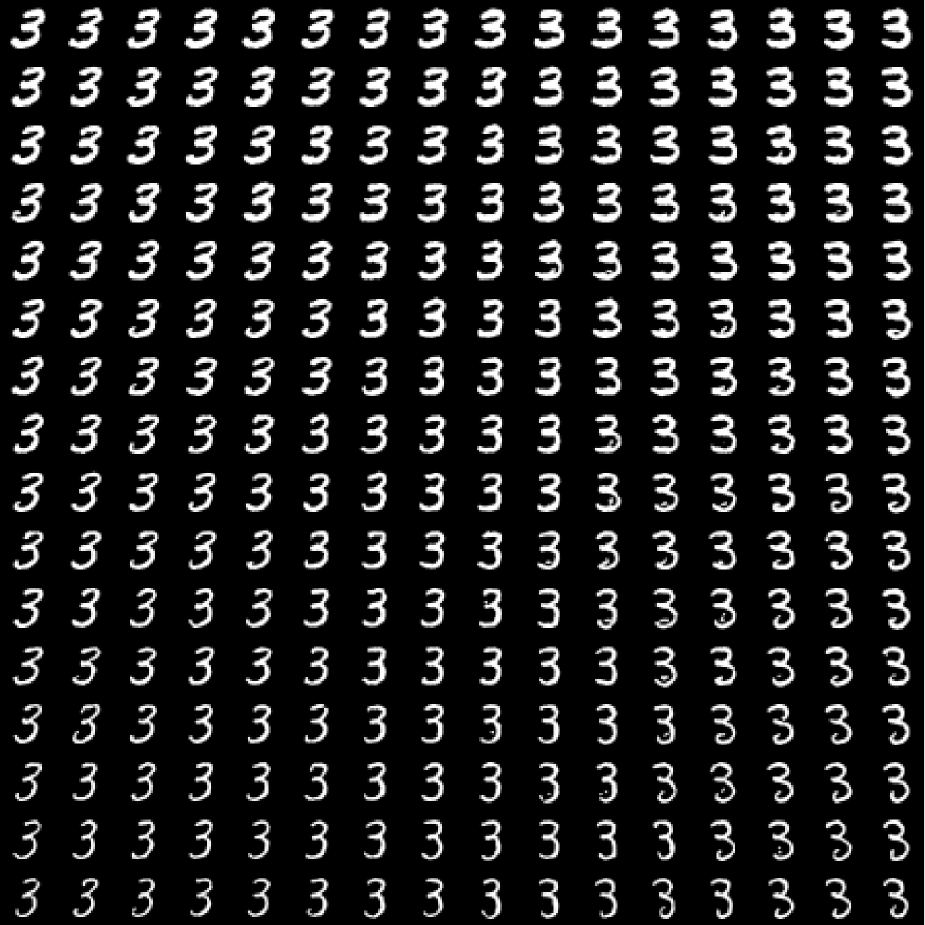}
			\subcaption{
				Class-conditional samples on a linearly spaced grid lying on a plane that is spanned by the first two principal components of the representation space ($\lambda =10^{-3}$).
			}
		\end{subfigure}
		\begin{subfigure}{0.48\textwidth}
			\begin{tabular*}{\textwidth}{@{}l @{\extracolsep{\fill}}cc}
				$\overbracket{\hspace{0.09\textwidth}}^{\mathrm{input}}$ &
				$\overbracket{\hspace{0.9\textwidth}}^{\mathrm{class-cond.\,reconstructions}}$
			\end{tabular*}
			\begin{tabularx}{\textwidth}{XXXXXXXXXXX}
				& c0 & c1 & c2 & c3 & c4 & c5 & c6 & c7 & c8 & c9
			\end{tabularx}
			\begin{tcbraster}[
				raster columns=11,
				raster equal height,
				raster column skip=0pt,
				raster row skip=0pt,
				raster every box/.style={blank},
				]
				\tcbincludegraphics{FigMNIST/rec_style_lambda1e-2_ddim/input0}
				\tcbincludegraphics{FigMNIST/rec_style_lambda1e-2_ddim/rec0-0}
				\tcbincludegraphics{FigMNIST/rec_style_lambda1e-2_ddim/rec0-1}
				\tcbincludegraphics{FigMNIST/rec_style_lambda1e-2_ddim/rec0-2}
				\tcbincludegraphics{FigMNIST/rec_style_lambda1e-2_ddim/rec0-3}
				\tcbincludegraphics{FigMNIST/rec_style_lambda1e-2_ddim/rec0-4}
				\tcbincludegraphics{FigMNIST/rec_style_lambda1e-2_ddim/rec0-5}
				\tcbincludegraphics{FigMNIST/rec_style_lambda1e-2_ddim/rec0-6}
				\tcbincludegraphics{FigMNIST/rec_style_lambda1e-2_ddim/rec0-7}
				\tcbincludegraphics{FigMNIST/rec_style_lambda1e-2_ddim/rec0-8}
				\tcbincludegraphics{FigMNIST/rec_style_lambda1e-2_ddim/rec0-9}
				\tcbincludegraphics{FigMNIST/rec_style_lambda1e-2_ddim/input2}
				\tcbincludegraphics{FigMNIST/rec_style_lambda1e-2_ddim/rec2-0}
				\tcbincludegraphics{FigMNIST/rec_style_lambda1e-2_ddim/rec2-1}
				\tcbincludegraphics{FigMNIST/rec_style_lambda1e-2_ddim/rec2-2}
				\tcbincludegraphics{FigMNIST/rec_style_lambda1e-2_ddim/rec2-3}
				\tcbincludegraphics{FigMNIST/rec_style_lambda1e-2_ddim/rec2-4}
				\tcbincludegraphics{FigMNIST/rec_style_lambda1e-2_ddim/rec2-5}
				\tcbincludegraphics{FigMNIST/rec_style_lambda1e-2_ddim/rec2-6}
				\tcbincludegraphics{FigMNIST/rec_style_lambda1e-2_ddim/rec2-7}
				\tcbincludegraphics{FigMNIST/rec_style_lambda1e-2_ddim/rec2-8}
				\tcbincludegraphics{FigMNIST/rec_style_lambda1e-2_ddim/rec2-9}
				\tcbincludegraphics{FigMNIST/rec_style_lambda1e-2_ddim/input3}
				\tcbincludegraphics{FigMNIST/rec_style_lambda1e-2_ddim/rec3-0}
				\tcbincludegraphics{FigMNIST/rec_style_lambda1e-2_ddim/rec3-1}
				\tcbincludegraphics{FigMNIST/rec_style_lambda1e-2_ddim/rec3-2}
				\tcbincludegraphics{FigMNIST/rec_style_lambda1e-2_ddim/rec3-3}
				\tcbincludegraphics{FigMNIST/rec_style_lambda1e-2_ddim/rec3-4}
				\tcbincludegraphics{FigMNIST/rec_style_lambda1e-2_ddim/rec3-5}
				\tcbincludegraphics{FigMNIST/rec_style_lambda1e-2_ddim/rec3-6}
				\tcbincludegraphics{FigMNIST/rec_style_lambda1e-2_ddim/rec3-7}
				\tcbincludegraphics{FigMNIST/rec_style_lambda1e-2_ddim/rec3-8}
				\tcbincludegraphics{FigMNIST/rec_style_lambda1e-2_ddim/rec3-9}
				\tcbincludegraphics{FigMNIST/rec_style_lambda1e-2_ddim/input4}
				\tcbincludegraphics{FigMNIST/rec_style_lambda1e-2_ddim/rec4-0}
				\tcbincludegraphics{FigMNIST/rec_style_lambda1e-2_ddim/rec4-1}
				\tcbincludegraphics{FigMNIST/rec_style_lambda1e-2_ddim/rec4-2}
				\tcbincludegraphics{FigMNIST/rec_style_lambda1e-2_ddim/rec4-3}
				\tcbincludegraphics{FigMNIST/rec_style_lambda1e-2_ddim/rec4-4}
				\tcbincludegraphics{FigMNIST/rec_style_lambda1e-2_ddim/rec4-5}
				\tcbincludegraphics{FigMNIST/rec_style_lambda1e-2_ddim/rec4-6}
				\tcbincludegraphics{FigMNIST/rec_style_lambda1e-2_ddim/rec4-7}
				\tcbincludegraphics{FigMNIST/rec_style_lambda1e-2_ddim/rec4-8}
				\tcbincludegraphics{FigMNIST/rec_style_lambda1e-2_ddim/rec4-9}
				\tcbincludegraphics{FigMNIST/rec_style_lambda1e-2_ddim/input5}
				\tcbincludegraphics{FigMNIST/rec_style_lambda1e-2_ddim/rec5-0}
				\tcbincludegraphics{FigMNIST/rec_style_lambda1e-2_ddim/rec5-1}
				\tcbincludegraphics{FigMNIST/rec_style_lambda1e-2_ddim/rec5-2}
				\tcbincludegraphics{FigMNIST/rec_style_lambda1e-2_ddim/rec5-3}
				\tcbincludegraphics{FigMNIST/rec_style_lambda1e-2_ddim/rec5-4}
				\tcbincludegraphics{FigMNIST/rec_style_lambda1e-2_ddim/rec5-5}
				\tcbincludegraphics{FigMNIST/rec_style_lambda1e-2_ddim/rec5-6}
				\tcbincludegraphics{FigMNIST/rec_style_lambda1e-2_ddim/rec5-7}
				\tcbincludegraphics{FigMNIST/rec_style_lambda1e-2_ddim/rec5-8}
				\tcbincludegraphics{FigMNIST/rec_style_lambda1e-2_ddim/rec5-9}
				\tcbincludegraphics{FigMNIST/rec_style_lambda1e-2_ddim/input6}
				\tcbincludegraphics{FigMNIST/rec_style_lambda1e-2_ddim/rec6-0}
				\tcbincludegraphics{FigMNIST/rec_style_lambda1e-2_ddim/rec6-1}
				\tcbincludegraphics{FigMNIST/rec_style_lambda1e-2_ddim/rec6-2}
				\tcbincludegraphics{FigMNIST/rec_style_lambda1e-2_ddim/rec6-3}
				\tcbincludegraphics{FigMNIST/rec_style_lambda1e-2_ddim/rec6-4}
				\tcbincludegraphics{FigMNIST/rec_style_lambda1e-2_ddim/rec6-5}
				\tcbincludegraphics{FigMNIST/rec_style_lambda1e-2_ddim/rec6-6}
				\tcbincludegraphics{FigMNIST/rec_style_lambda1e-2_ddim/rec6-7}
				\tcbincludegraphics{FigMNIST/rec_style_lambda1e-2_ddim/rec6-8}
				\tcbincludegraphics{FigMNIST/rec_style_lambda1e-2_ddim/rec6-9}
				\tcbincludegraphics{FigMNIST/rec_style_lambda1e-2_ddim/input7}
				\tcbincludegraphics{FigMNIST/rec_style_lambda1e-2_ddim/rec7-0}
				\tcbincludegraphics{FigMNIST/rec_style_lambda1e-2_ddim/rec7-1}
				\tcbincludegraphics{FigMNIST/rec_style_lambda1e-2_ddim/rec7-2}
				\tcbincludegraphics{FigMNIST/rec_style_lambda1e-2_ddim/rec7-3}
				\tcbincludegraphics{FigMNIST/rec_style_lambda1e-2_ddim/rec7-4}
				\tcbincludegraphics{FigMNIST/rec_style_lambda1e-2_ddim/rec7-5}
				\tcbincludegraphics{FigMNIST/rec_style_lambda1e-2_ddim/rec7-6}
				\tcbincludegraphics{FigMNIST/rec_style_lambda1e-2_ddim/rec7-7}
				\tcbincludegraphics{FigMNIST/rec_style_lambda1e-2_ddim/rec7-8}
				\tcbincludegraphics{FigMNIST/rec_style_lambda1e-2_ddim/rec7-9}
				\tcbincludegraphics{FigMNIST/rec_style_lambda1e-2_ddim/input10}
				\tcbincludegraphics{FigMNIST/rec_style_lambda1e-2_ddim/rec10-0}
				\tcbincludegraphics{FigMNIST/rec_style_lambda1e-2_ddim/rec10-1}
				\tcbincludegraphics{FigMNIST/rec_style_lambda1e-2_ddim/rec10-2}
				\tcbincludegraphics{FigMNIST/rec_style_lambda1e-2_ddim/rec10-3}
				\tcbincludegraphics{FigMNIST/rec_style_lambda1e-2_ddim/rec10-4}
				\tcbincludegraphics{FigMNIST/rec_style_lambda1e-2_ddim/rec10-5}
				\tcbincludegraphics{FigMNIST/rec_style_lambda1e-2_ddim/rec10-6}
				\tcbincludegraphics{FigMNIST/rec_style_lambda1e-2_ddim/rec10-7}
				\tcbincludegraphics{FigMNIST/rec_style_lambda1e-2_ddim/rec10-8}
				\tcbincludegraphics{FigMNIST/rec_style_lambda1e-2_ddim/rec10-9}
				\tcbincludegraphics{FigMNIST/rec_style_lambda1e-2_ddim/input9}
				\tcbincludegraphics{FigMNIST/rec_style_lambda1e-2_ddim/rec9-0}
				\tcbincludegraphics{FigMNIST/rec_style_lambda1e-2_ddim/rec9-1}
				\tcbincludegraphics{FigMNIST/rec_style_lambda1e-2_ddim/rec9-2}
				\tcbincludegraphics{FigMNIST/rec_style_lambda1e-2_ddim/rec9-3}
				\tcbincludegraphics{FigMNIST/rec_style_lambda1e-2_ddim/rec9-4}
				\tcbincludegraphics{FigMNIST/rec_style_lambda1e-2_ddim/rec9-5}
				\tcbincludegraphics{FigMNIST/rec_style_lambda1e-2_ddim/rec9-6}
				\tcbincludegraphics{FigMNIST/rec_style_lambda1e-2_ddim/rec9-7}
				\tcbincludegraphics{FigMNIST/rec_style_lambda1e-2_ddim/rec9-8}
				\tcbincludegraphics{FigMNIST/rec_style_lambda1e-2_ddim/rec9-9}
			\end{tcbraster}
			\subcaption{
				Class-appearance switching ($\lambda = 10^{-2}$). In rows: The representation (containing appearance information) extracted from the input image is decoded with different class-conditionings.
			}
			\label{fig: mnist class switch}
		\end{subfigure}
		\begin{subfigure}{\textwidth}
			\begin{tcbraster}[
				raster columns=18,
				raster equal height,
				raster column skip=0pt,
				raster row skip=0pt,
				raster every box/.style={blank},
				]
				\tcbincludegraphics{FigMNIST/interpolations_lambda1e-3/s2}
				\tcbincludegraphics{FigMNIST/interpolations_lambda1e-3/int2-0}
				\tcbincludegraphics{FigMNIST/interpolations_lambda1e-3/int2-1}
				\tcbincludegraphics{FigMNIST/interpolations_lambda1e-3/int2-2}
				\tcbincludegraphics{FigMNIST/interpolations_lambda1e-3/int2-3}
				\tcbincludegraphics{FigMNIST/interpolations_lambda1e-3/int2-4}
				\tcbincludegraphics{FigMNIST/interpolations_lambda1e-3/int2-5}
				\tcbincludegraphics{FigMNIST/interpolations_lambda1e-3/int2-6}
				\tcbincludegraphics{FigMNIST/interpolations_lambda1e-3/int2-7}
				\tcbincludegraphics{FigMNIST/interpolations_lambda1e-3/int2-8}
				\tcbincludegraphics{FigMNIST/interpolations_lambda1e-3/int2-9}
				\tcbincludegraphics{FigMNIST/interpolations_lambda1e-3/int2-10}
				\tcbincludegraphics{FigMNIST/interpolations_lambda1e-3/int2-11}
				\tcbincludegraphics{FigMNIST/interpolations_lambda1e-3/int2-12}
				\tcbincludegraphics{FigMNIST/interpolations_lambda1e-3/int2-13}
				\tcbincludegraphics{FigMNIST/interpolations_lambda1e-3/int2-14}
				\tcbincludegraphics{FigMNIST/interpolations_lambda1e-3/int2-15}
				\tcbincludegraphics{FigMNIST/interpolations_lambda1e-3/t2}
				\tcbincludegraphics{FigMNIST/interpolations_lambda1e-3/s5}
				\tcbincludegraphics{FigMNIST/interpolations_lambda1e-3/int5-0}
				\tcbincludegraphics{FigMNIST/interpolations_lambda1e-3/int5-1}
				\tcbincludegraphics{FigMNIST/interpolations_lambda1e-3/int5-2}
				\tcbincludegraphics{FigMNIST/interpolations_lambda1e-3/int5-3}
				\tcbincludegraphics{FigMNIST/interpolations_lambda1e-3/int5-4}
				\tcbincludegraphics{FigMNIST/interpolations_lambda1e-3/int5-5}
				\tcbincludegraphics{FigMNIST/interpolations_lambda1e-3/int5-6}
				\tcbincludegraphics{FigMNIST/interpolations_lambda1e-3/int5-7}
				\tcbincludegraphics{FigMNIST/interpolations_lambda1e-3/int5-8}
				\tcbincludegraphics{FigMNIST/interpolations_lambda1e-3/int5-9}
				\tcbincludegraphics{FigMNIST/interpolations_lambda1e-3/int5-10}
				\tcbincludegraphics{FigMNIST/interpolations_lambda1e-3/int5-11}
				\tcbincludegraphics{FigMNIST/interpolations_lambda1e-3/int5-12}
				\tcbincludegraphics{FigMNIST/interpolations_lambda1e-3/int5-13}
				\tcbincludegraphics{FigMNIST/interpolations_lambda1e-3/int5-14}
				\tcbincludegraphics{FigMNIST/interpolations_lambda1e-3/int5-15}
				\tcbincludegraphics{FigMNIST/interpolations_lambda1e-3/t5}
			\end{tcbraster}
			\subcaption{Interpolations ($\lambda = 10^{-3}$)}
		\end{subfigure}
		\caption[Class-conditional MNIST representation learning]{
			\begin{minipage}[t]{0.8\textwidth}
				Results for class-conditional representation learning with diffusion models on MNIST. Due to the explicit class-conditioning, the representation encodes the appearance, such as stroke width and tilt.
			\end{minipage}
		}
		\label{fig: mnist}
	\end{figure}
}

\newcommand{\FigStyleShapeSwap}
{
	\begin{figure}[H]
		\centering
		
		\begin{tabular*}{\textwidth}{@{}l @{\extracolsep{\fill}}cc}
			$\overbracket{\hspace{0.1\textwidth}}^{\mathrm{shape}}$ &
			$\overbracket{\hspace{0.88\textwidth}}^{\mathrm{style}}$
		\end{tabular*}
		\begin{tcbraster}[
			raster columns=9,
			raster equal height, 
			raster column skip=0pt,
			raster row skip=0pt,
			raster every box/.style={blank}
			]
			\tcbox{}
			\tcbincludegraphics{FigStyleShapeSwap/img0}
			\tcbincludegraphics{FigStyleShapeSwap/img1}
			\tcbincludegraphics{FigStyleShapeSwap/img2}
			\tcbincludegraphics{FigStyleShapeSwap/img3}
			\tcbincludegraphics{FigStyleShapeSwap/img4}
			\tcbincludegraphics{FigStyleShapeSwap/img5}
			\tcbincludegraphics{FigStyleShapeSwap/img6}
			\tcbincludegraphics{FigStyleShapeSwap/img7}
			
			\tcbincludegraphics{FigStyleShapeSwap/img0}
			\tcbincludegraphics{FigStyleShapeSwap/0}
			\tcbincludegraphics{FigStyleShapeSwap/1}
			\tcbincludegraphics{FigStyleShapeSwap/2}
			\tcbincludegraphics{FigStyleShapeSwap/3}
			\tcbincludegraphics{FigStyleShapeSwap/4}
			\tcbincludegraphics{FigStyleShapeSwap/5}
			\tcbincludegraphics{FigStyleShapeSwap/6}
			\tcbincludegraphics{FigStyleShapeSwap/7}
			
			\tcbincludegraphics{FigStyleShapeSwap/img1}
			\tcbincludegraphics{FigStyleShapeSwap/8}
			\tcbincludegraphics{FigStyleShapeSwap/9}
			\tcbincludegraphics{FigStyleShapeSwap/10}
			\tcbincludegraphics{FigStyleShapeSwap/11}
			\tcbincludegraphics{FigStyleShapeSwap/12}
			\tcbincludegraphics{FigStyleShapeSwap/13}
			\tcbincludegraphics{FigStyleShapeSwap/14}
			\tcbincludegraphics{FigStyleShapeSwap/15}
			
			\tcbincludegraphics{FigStyleShapeSwap/img2}
			\tcbincludegraphics{FigStyleShapeSwap/16}
			\tcbincludegraphics{FigStyleShapeSwap/17}
			\tcbincludegraphics{FigStyleShapeSwap/18}
			\tcbincludegraphics{FigStyleShapeSwap/19}
			\tcbincludegraphics{FigStyleShapeSwap/20}
			\tcbincludegraphics{FigStyleShapeSwap/21}
			\tcbincludegraphics{FigStyleShapeSwap/22}
			\tcbincludegraphics{FigStyleShapeSwap/23}
			
			\tcbincludegraphics{FigStyleShapeSwap/img3}
			\tcbincludegraphics{FigStyleShapeSwap/24}
			\tcbincludegraphics{FigStyleShapeSwap/25}
			\tcbincludegraphics{FigStyleShapeSwap/26}
			\tcbincludegraphics{FigStyleShapeSwap/27}
			\tcbincludegraphics{FigStyleShapeSwap/28}
			\tcbincludegraphics{FigStyleShapeSwap/29}
			\tcbincludegraphics{FigStyleShapeSwap/30}
			\tcbincludegraphics{FigStyleShapeSwap/31}
			
			\tcbincludegraphics{FigStyleShapeSwap/img4}
			\tcbincludegraphics{FigStyleShapeSwap/32}
			\tcbincludegraphics{FigStyleShapeSwap/33}
			\tcbincludegraphics{FigStyleShapeSwap/34}
			\tcbincludegraphics{FigStyleShapeSwap/35}
			\tcbincludegraphics{FigStyleShapeSwap/36}
			\tcbincludegraphics{FigStyleShapeSwap/37}
			\tcbincludegraphics{FigStyleShapeSwap/38}
			\tcbincludegraphics{FigStyleShapeSwap/39}
			
			\tcbincludegraphics{FigStyleShapeSwap/img5}
			\tcbincludegraphics{FigStyleShapeSwap/40}
			\tcbincludegraphics{FigStyleShapeSwap/41}
			\tcbincludegraphics{FigStyleShapeSwap/42}
			\tcbincludegraphics{FigStyleShapeSwap/43}
			\tcbincludegraphics{FigStyleShapeSwap/44}
			\tcbincludegraphics{FigStyleShapeSwap/45}
			\tcbincludegraphics{FigStyleShapeSwap/46}
			\tcbincludegraphics{FigStyleShapeSwap/47}
			
			\tcbincludegraphics{FigStyleShapeSwap/img6}
			\tcbincludegraphics{FigStyleShapeSwap/48}
			\tcbincludegraphics{FigStyleShapeSwap/49}
			\tcbincludegraphics{FigStyleShapeSwap/50}
			\tcbincludegraphics{FigStyleShapeSwap/51}
			\tcbincludegraphics{FigStyleShapeSwap/52}
			\tcbincludegraphics{FigStyleShapeSwap/53}
			\tcbincludegraphics{FigStyleShapeSwap/54}
			\tcbincludegraphics{FigStyleShapeSwap/55}
			
			\tcbincludegraphics{FigStyleShapeSwap/img7}
			\tcbincludegraphics{FigStyleShapeSwap/56}
			\tcbincludegraphics{FigStyleShapeSwap/57}
			\tcbincludegraphics{FigStyleShapeSwap/58}
			\tcbincludegraphics{FigStyleShapeSwap/59}
			\tcbincludegraphics{FigStyleShapeSwap/60}
			\tcbincludegraphics{FigStyleShapeSwap/61}
			\tcbincludegraphics{FigStyleShapeSwap/62}
			\tcbincludegraphics{FigStyleShapeSwap/63}
		\end{tcbraster}
		
		\caption[Style-Shape transfer on LSUN-Churches]{
			\begin{minipage}[t]{0.8\textwidth}
				Style-Shape-switching on LSUN-Churches (representation shape $4\times32\times32$, $\lambda=5\cdot10^{-6}$). The style-encoder receives warped input during training, thereby enforcing the shape information to be encoded in the KL-regularized bottleneck of the shape-encoder. For the style-shape-switches, each encoder receives a different input image. Then we sample from the diffusion model conditioned on the combined representation.
			\end{minipage}
		}
		\label{fig: style shape swap}
	\end{figure}
}

\newcommand{\FigLinearNoiseSchedule}
{
	\begin{figure}[H]
		\centering
		
		\begin{subfigure}[b]{0.32\textwidth}
			\includegraphics[width=\textwidth]{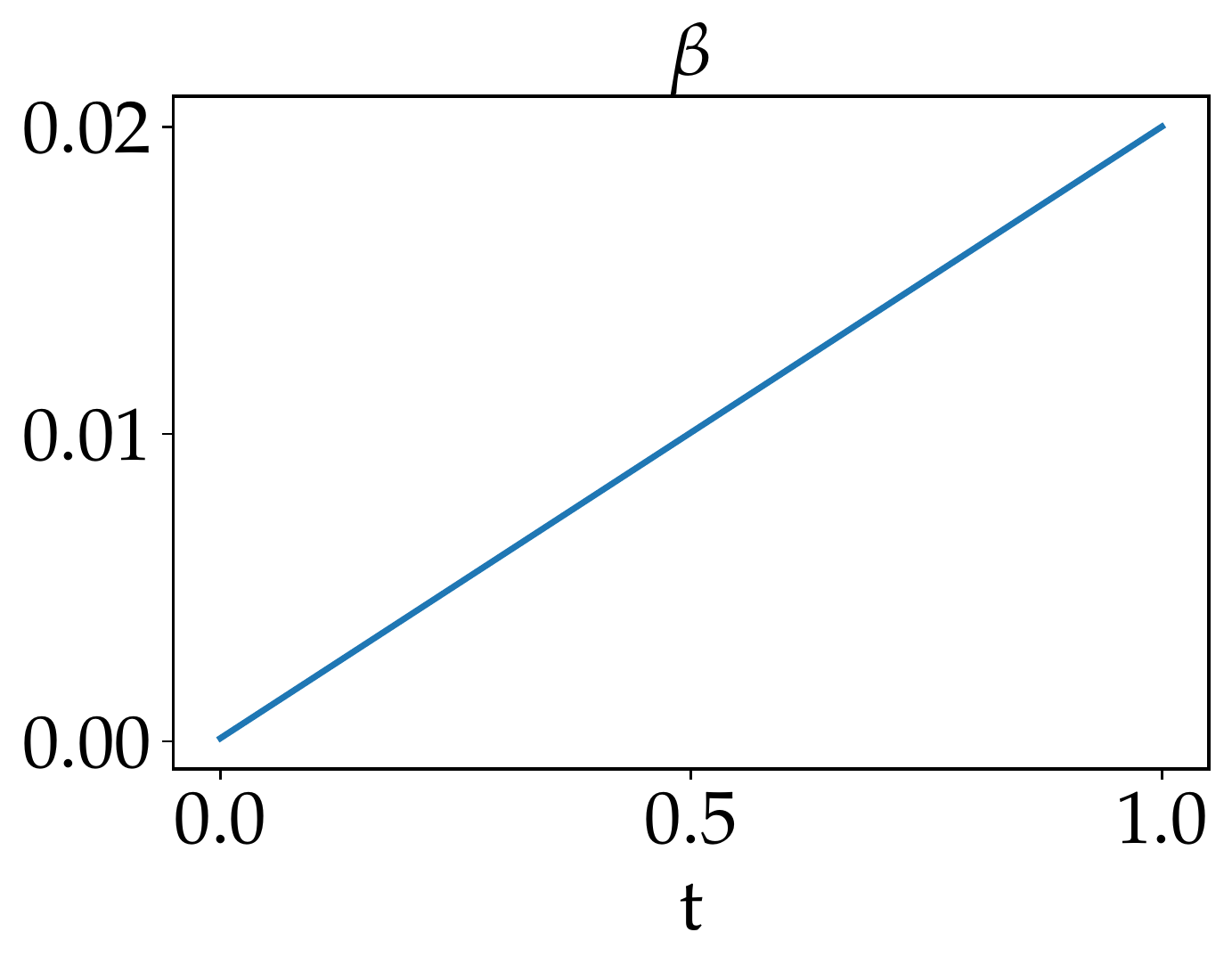}
		\end{subfigure}
		\begin{subfigure}[b]{0.32\textwidth}
			\includegraphics[width=\textwidth]{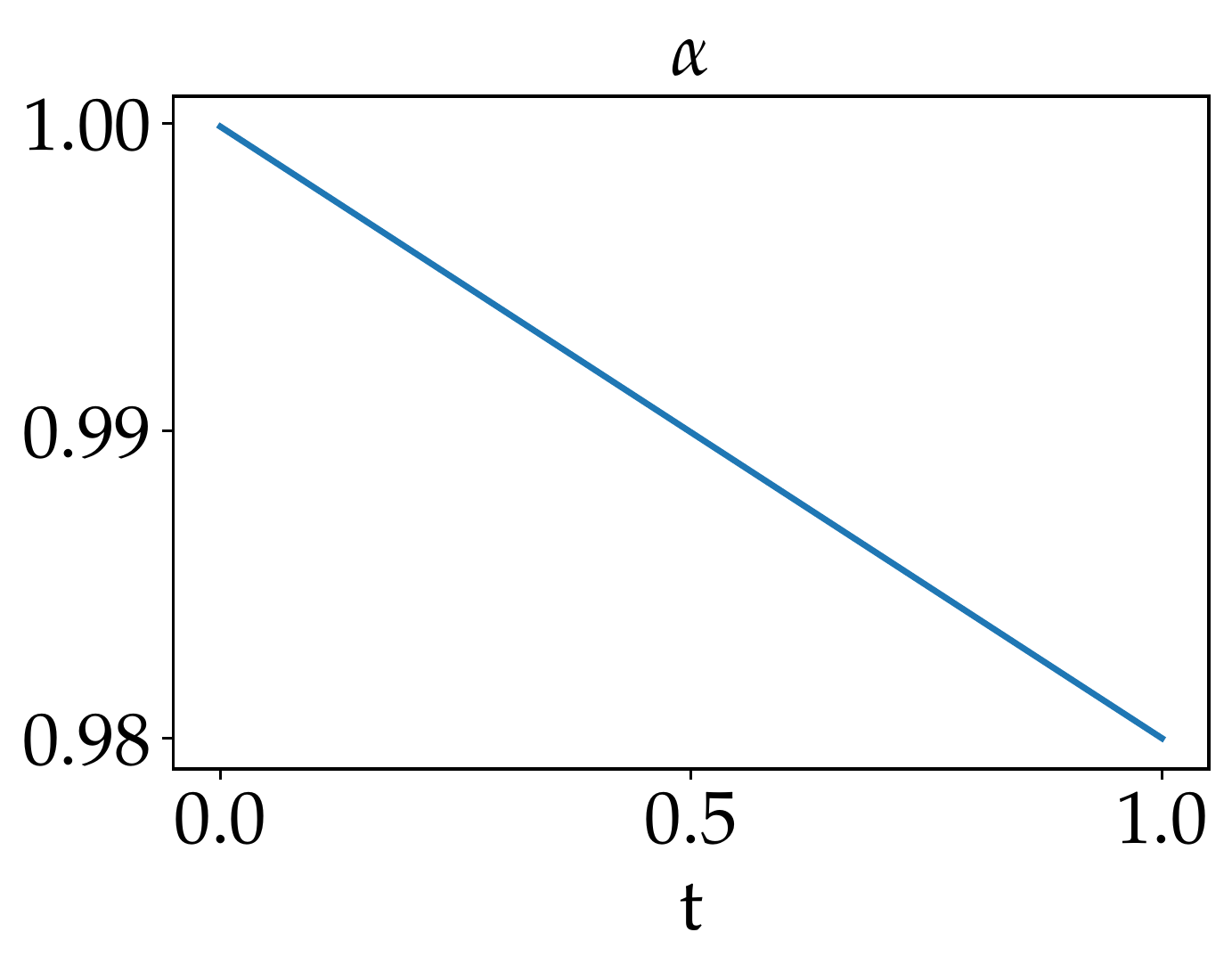}
		\end{subfigure}
		\begin{subfigure}[b]{0.32\textwidth}
			\includegraphics[width=\textwidth]{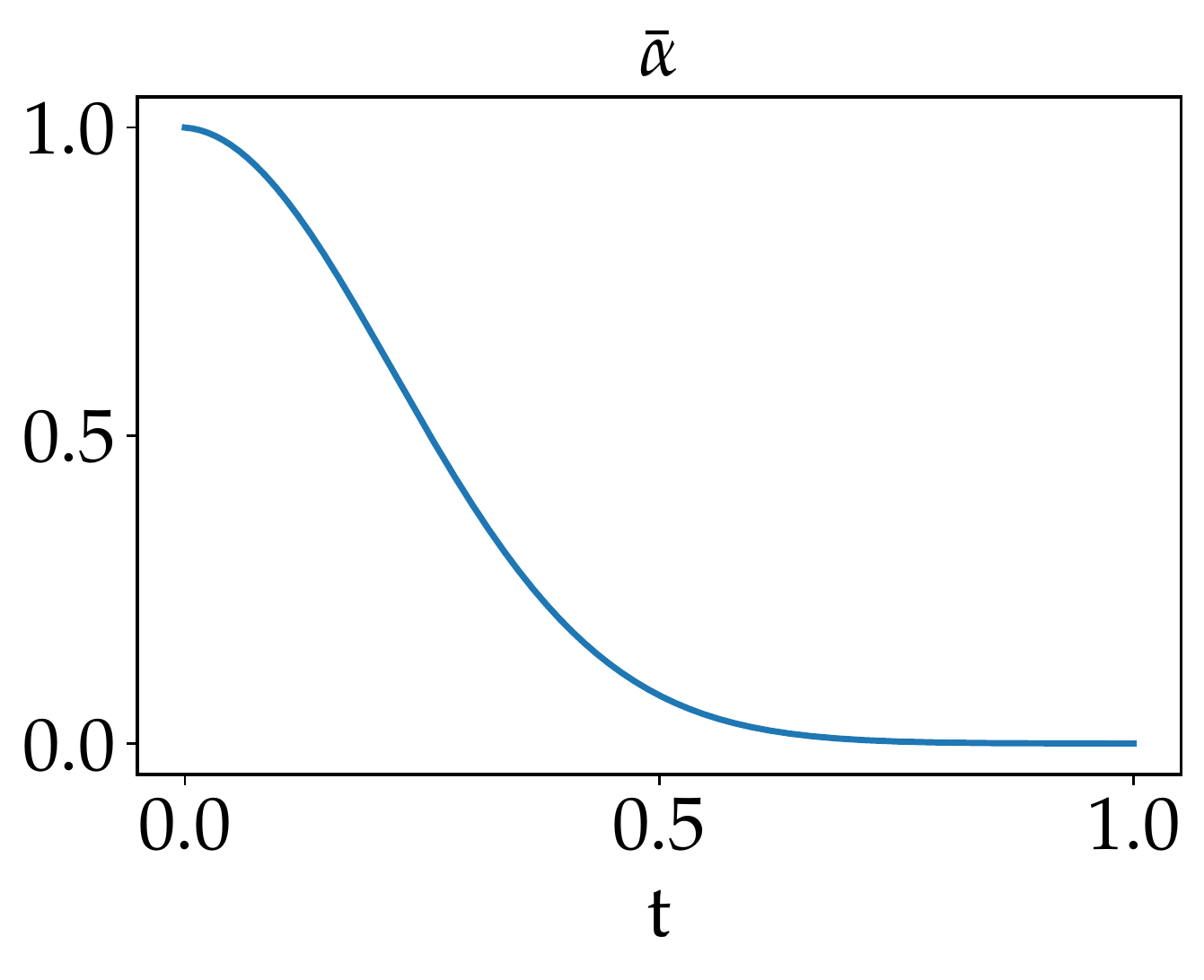}
		\end{subfigure}
		\begin{subfigure}[b]{0.32\textwidth}
			\includegraphics[width=\textwidth]{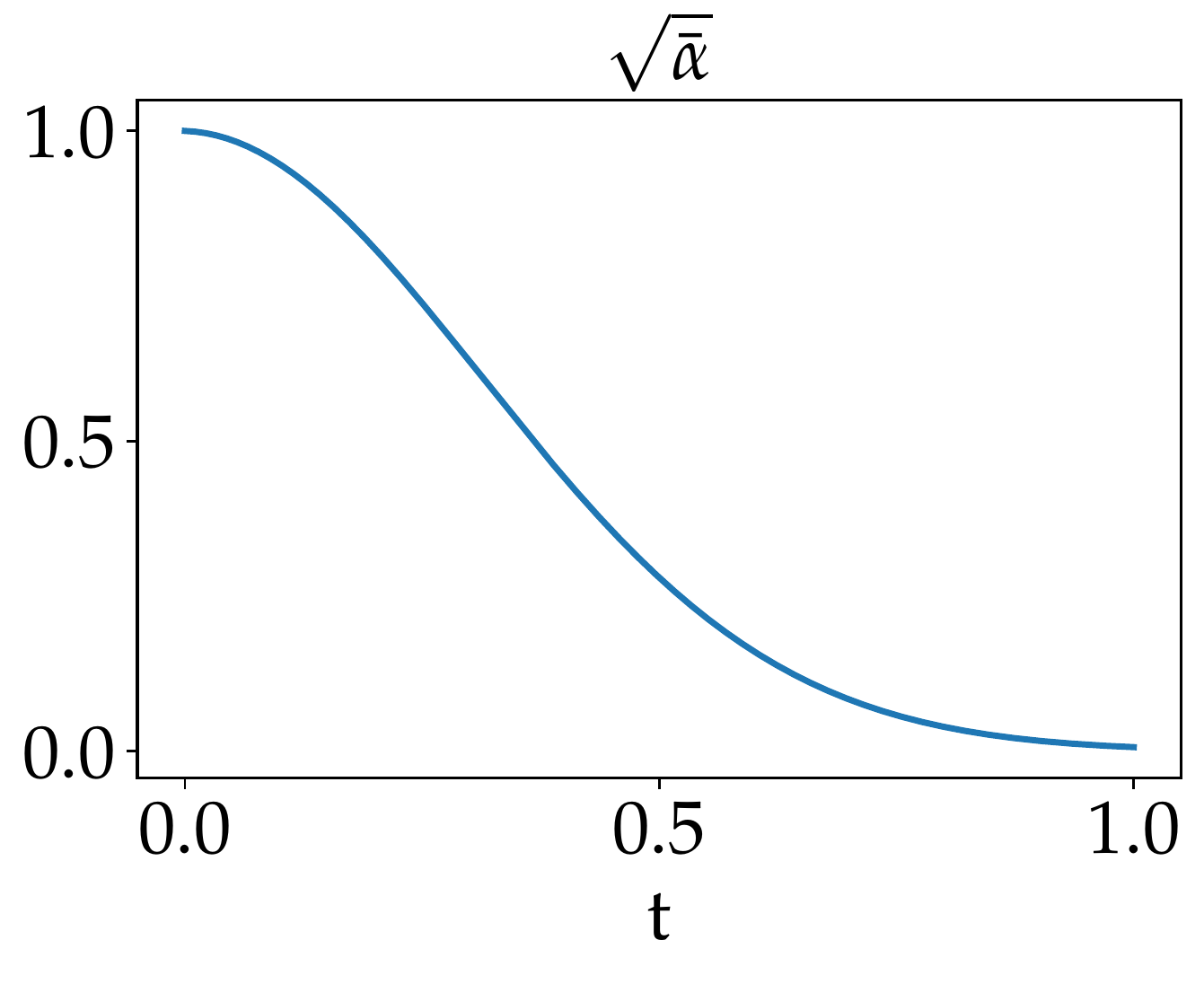}
		\end{subfigure}
		\begin{subfigure}[b]{0.32\textwidth}
			\includegraphics[width=\textwidth]{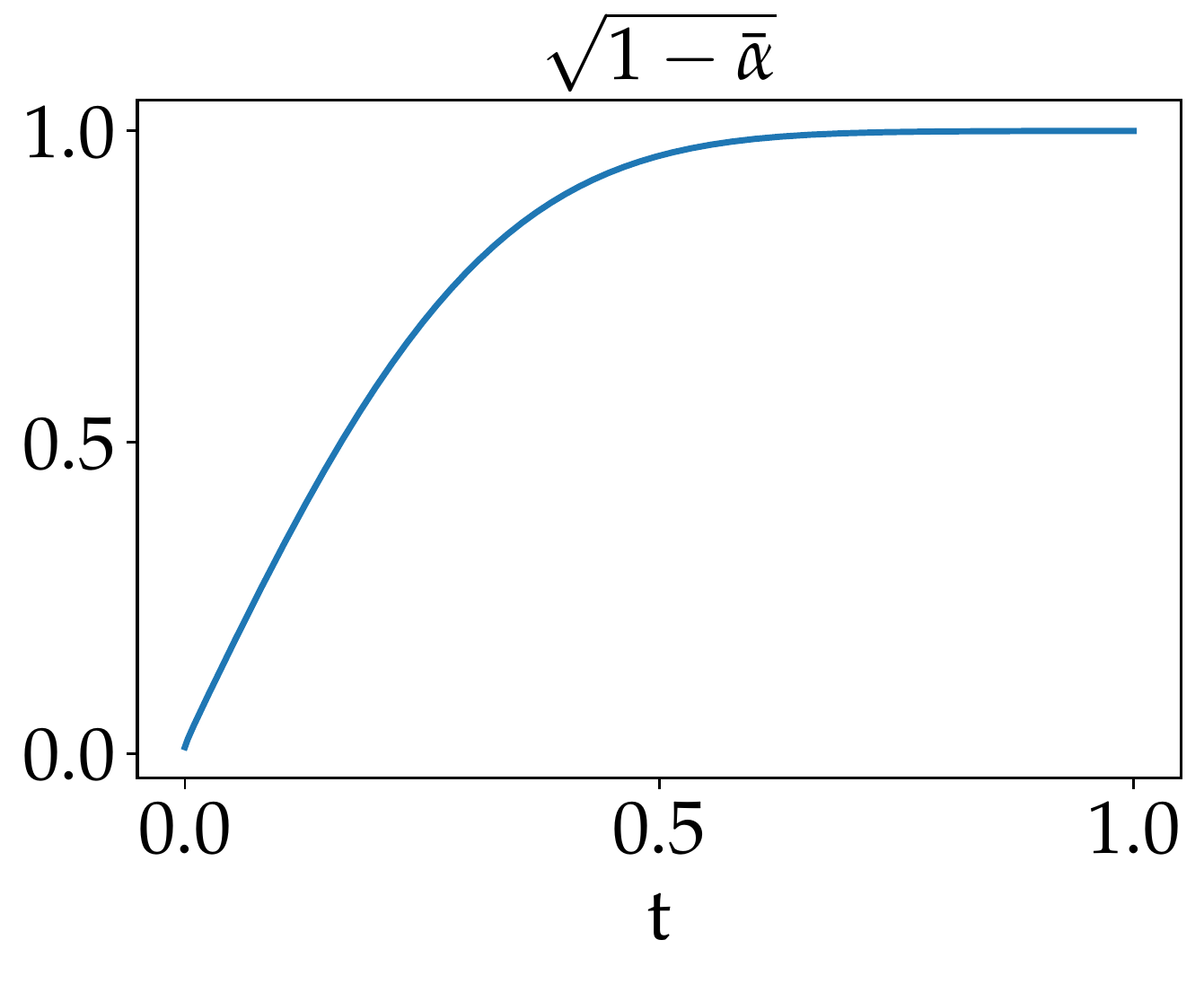}
		\end{subfigure}
		\captionsetup{justification=centering}
		\caption[Linear noise schedule]{Relevant factors for a linear $\beta$-schedule (see \sec{sec:diffusion process}).}
		\label{fig: linear noise schedule}
	\end{figure}
}

\newcommand{\FigReconstructionsLSUNchurchDeep}
{
	\newgeometry{top=2.5cm}
	\begin{figure}[H]
		\centering
		
		\begin{subfigure}[t]{0.9\textwidth}
			\begin{tabular*}{\textwidth}{@{}l @{\extracolsep{\fill}}cc}
				$\overbracket{\hspace{0.16\textwidth}}^{\mathrm{input}}$ &
				$\overbracket{\hspace{0.81\textwidth}}^{\mathrm{reconstructions}}$
			\end{tabular*}
			\begin{tcbraster}[
				raster columns=6,
				raster equal height,
				raster column skip=0pt,
				raster row skip=0pt,
				raster every box/.style={blank},
				]
				\tcbincludegraphics{FigReconstructionsLSUNchurchDeep/input14}
				\tcbincludegraphics{FigReconstructionsLSUNchurchDeep/rec14-0}
				\tcbincludegraphics{FigReconstructionsLSUNchurchDeep/rec14-1}
				\tcbincludegraphics{FigReconstructionsLSUNchurchDeep/rec14-2}
				\tcbincludegraphics{FigReconstructionsLSUNchurchDeep/rec14-3}
				\tcbincludegraphics{FigReconstructionsLSUNchurchDeep/rec14-6}
				\tcbincludegraphics{FigReconstructionsLSUNchurchDeep/input28}
				\tcbincludegraphics{FigReconstructionsLSUNchurchDeep/rec28-1}
				\tcbincludegraphics{FigReconstructionsLSUNchurchDeep/rec28-2}
				\tcbincludegraphics{FigReconstructionsLSUNchurchDeep/rec28-3}
				\tcbincludegraphics{FigReconstructionsLSUNchurchDeep/rec28-5}
				\tcbincludegraphics{FigReconstructionsLSUNchurchDeep/rec28-7}
				\tcbincludegraphics{FigReconstructionsLSUNchurchDeep/input43}
				\tcbincludegraphics{FigReconstructionsLSUNchurchDeep/rec43-1}
				\tcbincludegraphics{FigReconstructionsLSUNchurchDeep/rec43-2}
				\tcbincludegraphics{FigReconstructionsLSUNchurchDeep/rec43-4}
				\tcbincludegraphics{FigReconstructionsLSUNchurchDeep/rec43-6}
				\tcbincludegraphics{FigReconstructionsLSUNchurchDeep/rec43-7}
				\tcbincludegraphics{FigReconstructionsLSUNchurchDeep/input57}
				\tcbincludegraphics{FigReconstructionsLSUNchurchDeep/rec57-1}
				\tcbincludegraphics{FigReconstructionsLSUNchurchDeep/rec57-2}
				\tcbincludegraphics{FigReconstructionsLSUNchurchDeep/rec57-4}
				\tcbincludegraphics{FigReconstructionsLSUNchurchDeep/rec57-6}
				\tcbincludegraphics{FigReconstructionsLSUNchurchDeep/rec57-7}
			\end{tcbraster}
		\subcaption{Representation shape $64\times8\times8$}
		\end{subfigure}
		
		\begin{subfigure}[t]{0.9\textwidth}
			\begin{tcbraster}[
				raster columns=6,
				raster equal height,
				raster column skip=0pt,
				raster row skip=0pt,
				raster every box/.style={blank},
				]
				\tcbincludegraphics{FigReconstructionsLSUNchurch/input7}
				\tcbincludegraphics{FigReconstructionsLSUNchurch/rec7-1}
				\tcbincludegraphics{FigReconstructionsLSUNchurch/rec7-2}
				\tcbincludegraphics{FigReconstructionsLSUNchurch/rec7-3}
				\tcbincludegraphics{FigReconstructionsLSUNchurch/rec7-4}
				\tcbincludegraphics{FigReconstructionsLSUNchurch/rec7-6}
				\tcbincludegraphics{FigReconstructionsLSUNchurch/input0}
				\tcbincludegraphics{FigReconstructionsLSUNchurch/rec0-0}
				\tcbincludegraphics{FigReconstructionsLSUNchurch/rec0-2}
				\tcbincludegraphics{FigReconstructionsLSUNchurch/rec0-3}
				\tcbincludegraphics{FigReconstructionsLSUNchurch/rec0-4}
				\tcbincludegraphics{FigReconstructionsLSUNchurch/rec0-6}
				\tcbincludegraphics{FigReconstructionsLSUNchurch/input6}
				\tcbincludegraphics{FigReconstructionsLSUNchurch/rec6-0}
				\tcbincludegraphics{FigReconstructionsLSUNchurch/rec6-1}
				\tcbincludegraphics{FigReconstructionsLSUNchurch/rec6-3}
				\tcbincludegraphics{FigReconstructionsLSUNchurch/rec6-4}
				\tcbincludegraphics{FigReconstructionsLSUNchurch/rec6-6}
				\tcbincludegraphics{FigReconstructionsLSUNchurch/input3}
				\tcbincludegraphics{FigReconstructionsLSUNchurch/rec3-1}
				\tcbincludegraphics{FigReconstructionsLSUNchurch/rec3-2}
				\tcbincludegraphics{FigReconstructionsLSUNchurch/rec3-3}
				\tcbincludegraphics{FigReconstructionsLSUNchurch/rec3-5}
				\tcbincludegraphics{FigReconstructionsLSUNchurch/rec3-7}
			\end{tcbraster}
		\subcaption{Representation shape $16\times16\times16$}
		\end{subfigure}
		
		\caption[LRDM: LSUN-Churches reconstructions (probabilistic sampling)]{
			\begin{minipage}[t]{0.8\textwidth}
				LRDM (VQ-AE) reconstructions for different representation dimensionalities, using probabilistic sampling and re-sampling $\z_T$ for each reconstruction ($\lambda=10^{-4}$, LSUN-Churches). $\rep$ is sampled from the approximate posterior.
			\end{minipage}
		}
		\label{fig: reconstructions LSUNchurch deep}
	\end{figure}
	\restoregeometry
}

\newcommand{\FigReconstructionsCelebAHQ}
{
	\begin{figure}[H]
		\centering
		\begin{tabular*}{\textwidth}{@{}l @{\extracolsep{\fill}}cc}
			$\overbracket{\hspace{0.16\textwidth}}^{\mathrm{input}}$ &
			$\overbracket{\hspace{0.81\textwidth}}^{\mathrm{reconstructions}}$
		\end{tabular*}
		\begin{tcbraster}[
			raster columns=6,
			raster equal height, 
			raster column skip=0pt,
			raster row skip=0pt,
			raster every box/.style={blank}
			]
			\tcbincludegraphics{FigReconstructionsCelebAHQ/input15}
			\tcbincludegraphics{FigReconstructionsCelebAHQ/rec15-0}
			\tcbincludegraphics{FigReconstructionsCelebAHQ/rec15-1}
			\tcbincludegraphics{FigReconstructionsCelebAHQ/rec15-2}
			\tcbincludegraphics{FigReconstructionsCelebAHQ/rec15-3}
			\tcbincludegraphics{FigReconstructionsCelebAHQ/rec15-4}
			\tcbincludegraphics{FigReconstructionsCelebAHQ/input12}
			\tcbincludegraphics{FigReconstructionsCelebAHQ/rec12-0}
			\tcbincludegraphics{FigReconstructionsCelebAHQ/rec12-1}
			\tcbincludegraphics{FigReconstructionsCelebAHQ/rec12-2}
			\tcbincludegraphics{FigReconstructionsCelebAHQ/rec12-3}
			\tcbincludegraphics{FigReconstructionsCelebAHQ/rec12-6}
			\tcbincludegraphics{FigReconstructionsCelebAHQ/input8}
			\tcbincludegraphics{FigReconstructionsCelebAHQ/rec8-0}
			\tcbincludegraphics{FigReconstructionsCelebAHQ/rec8-1}
			\tcbincludegraphics{FigReconstructionsCelebAHQ/rec8-3}
			\tcbincludegraphics{FigReconstructionsCelebAHQ/rec8-5}
			\tcbincludegraphics{FigReconstructionsCelebAHQ/rec8-7}
			\tcbincludegraphics{FigReconstructionsCelebAHQ/input4}
			\tcbincludegraphics{FigReconstructionsCelebAHQ/rec4-0}
			\tcbincludegraphics{FigReconstructionsCelebAHQ/rec4-1}
			\tcbincludegraphics{FigReconstructionsCelebAHQ/rec4-2}
			\tcbincludegraphics{FigReconstructionsCelebAHQ/rec4-3}
			\tcbincludegraphics{FigReconstructionsCelebAHQ/rec4-6}
			\tcbincludegraphics{FigReconstructionsCelebAHQ/input5}
			\tcbincludegraphics{FigReconstructionsCelebAHQ/rec5-0}
			\tcbincludegraphics{FigReconstructionsCelebAHQ/rec5-1}
			\tcbincludegraphics{FigReconstructionsCelebAHQ/rec5-2}
			\tcbincludegraphics{FigReconstructionsCelebAHQ/rec5-3}
			\tcbincludegraphics{FigReconstructionsCelebAHQ/rec5-5}
		\end{tcbraster}
		
		\caption[LRDM: CelebA-HQ reconstructions (probabilistic sampling)]{
			\begin{minipage}[t]{0.8\textwidth}
				LRDM (VQ-AE) reconstructions with representation of shape $64\times8\times8$, using probabilistic sampling and re-sampling $\z_T$ for each reconstruction, $\lambda=10^{-4}$, CelebA-HQ. $\rep$ is sampled from the approximate posterior.
			\end{minipage}
		}
		\label{fig: reconstructions CelebA-HQ}
	\end{figure}
}

\newcommand{\FigInterpolationsLSUNchurchDeep}
{
	\begin{figure}[H]
		\centering
		
		\begin{tabular*}{\textwidth}{@{\extracolsep{\fill}}ccc}
			$\overbracket{\hspace{0.09\textwidth}}^{\mathrm{input\,1}}$ &
			$\overbracket{\hspace{0.755\textwidth}}^{\mathrm{interpolations}}$ &
			$\overbracket{\hspace{0.09\textwidth}}^{\mathrm{input\,2}}$
		\end{tabular*}
		\begin{tcbraster}[
			raster columns=10,
			raster equal height,
			raster column skip=0pt,
			raster row skip=0pt,
			raster every box/.style={blank},
			]
			\tcbincludegraphics{FigInterpolationsLSUNchurchDeep/s7-3724}
			\tcbincludegraphics{FigInterpolationsLSUNchurchDeep/int7-0}
			\tcbincludegraphics{FigInterpolationsLSUNchurchDeep/int7-2}
			\tcbincludegraphics{FigInterpolationsLSUNchurchDeep/int7-4}
			\tcbincludegraphics{FigInterpolationsLSUNchurchDeep/int7-6}
			\tcbincludegraphics{FigInterpolationsLSUNchurchDeep/int7-8}
			\tcbincludegraphics{FigInterpolationsLSUNchurchDeep/int7-10}
			\tcbincludegraphics{FigInterpolationsLSUNchurchDeep/int7-12}
			\tcbincludegraphics{FigInterpolationsLSUNchurchDeep/int7-14}
			\tcbincludegraphics{FigInterpolationsLSUNchurchDeep/s7-4297}
			\tcbincludegraphics{FigInterpolationsLSUNchurchDeep/s2-1891}
			\tcbincludegraphics{FigInterpolationsLSUNchurchDeep/int2-0}
			\tcbincludegraphics{FigInterpolationsLSUNchurchDeep/int2-2}
			\tcbincludegraphics{FigInterpolationsLSUNchurchDeep/int2-4}
			\tcbincludegraphics{FigInterpolationsLSUNchurchDeep/int2-6}
			\tcbincludegraphics{FigInterpolationsLSUNchurchDeep/int2-8}
			\tcbincludegraphics{FigInterpolationsLSUNchurchDeep/int2-10}
			\tcbincludegraphics{FigInterpolationsLSUNchurchDeep/int2-12}
			\tcbincludegraphics{FigInterpolationsLSUNchurchDeep/int2-14}
			\tcbincludegraphics{FigInterpolationsLSUNchurchDeep/s2-1256}
			\tcbincludegraphics{FigInterpolationsLSUNchurchDeep/s1-462}
			\tcbincludegraphics{FigInterpolationsLSUNchurchDeep/int1-0}
			\tcbincludegraphics{FigInterpolationsLSUNchurchDeep/int1-2}
			\tcbincludegraphics{FigInterpolationsLSUNchurchDeep/int1-4}
			\tcbincludegraphics{FigInterpolationsLSUNchurchDeep/int1-6}
			\tcbincludegraphics{FigInterpolationsLSUNchurchDeep/int1-8}
			\tcbincludegraphics{FigInterpolationsLSUNchurchDeep/int1-10}
			\tcbincludegraphics{FigInterpolationsLSUNchurchDeep/int1-12}
			\tcbincludegraphics{FigInterpolationsLSUNchurchDeep/int1-14}
			\tcbincludegraphics{FigInterpolationsLSUNchurchDeep/s1-2231}
			\tcbincludegraphics{FigInterpolationsLSUNchurchDeep/s3-703}
			\tcbincludegraphics{FigInterpolationsLSUNchurchDeep/int3-0}
			\tcbincludegraphics{FigInterpolationsLSUNchurchDeep/int3-2}
			\tcbincludegraphics{FigInterpolationsLSUNchurchDeep/int3-4}
			\tcbincludegraphics{FigInterpolationsLSUNchurchDeep/int3-6}
			\tcbincludegraphics{FigInterpolationsLSUNchurchDeep/int3-8}
			\tcbincludegraphics{FigInterpolationsLSUNchurchDeep/int3-10}
			\tcbincludegraphics{FigInterpolationsLSUNchurchDeep/int3-12}
			\tcbincludegraphics{FigInterpolationsLSUNchurchDeep/int3-14}
			\tcbincludegraphics{FigInterpolationsLSUNchurchDeep/s3-3700}
			\tcbincludegraphics{FigInterpolationsLSUNchurchDeep/s6-972}
			\tcbincludegraphics{FigInterpolationsLSUNchurchDeep/int6-0}
			\tcbincludegraphics{FigInterpolationsLSUNchurchDeep/int6-2}
			\tcbincludegraphics{FigInterpolationsLSUNchurchDeep/int6-4}
			\tcbincludegraphics{FigInterpolationsLSUNchurchDeep/int6-6}
			\tcbincludegraphics{FigInterpolationsLSUNchurchDeep/int6-8}
			\tcbincludegraphics{FigInterpolationsLSUNchurchDeep/int6-10}
			\tcbincludegraphics{FigInterpolationsLSUNchurchDeep/int6-12}
			\tcbincludegraphics{FigInterpolationsLSUNchurchDeep/int6-14}
			\tcbincludegraphics{FigInterpolationsLSUNchurchDeep/s6-3803}
			\tcbincludegraphics{FigInterpolationsLSUNchurchDeep/s8-3185}
			\tcbincludegraphics{FigInterpolationsLSUNchurchDeep/int8-0}
			\tcbincludegraphics{FigInterpolationsLSUNchurchDeep/int8-2}
			\tcbincludegraphics{FigInterpolationsLSUNchurchDeep/int8-4}
			\tcbincludegraphics{FigInterpolationsLSUNchurchDeep/int8-6}
			\tcbincludegraphics{FigInterpolationsLSUNchurchDeep/int8-8}
			\tcbincludegraphics{FigInterpolationsLSUNchurchDeep/int8-10}
			\tcbincludegraphics{FigInterpolationsLSUNchurchDeep/int8-12}
			\tcbincludegraphics{FigInterpolationsLSUNchurchDeep/int8-14}
			\tcbincludegraphics{FigInterpolationsLSUNchurchDeep/s8-2916}
		\end{tcbraster}
		\caption[LRDM: LSUN-Churches interpolations (probabilistic sampling)]{
			\begin{minipage}[t]{0.8\textwidth}
				LRDM (VQ-AE) interpolations (slerp) on LSUN-Churches for representation shape $64\times8\times8$, $\lambda=10^{-4}$, using probabilistic sampling and re-sampling $\z_T$ for each sample.
			\end{minipage}
		}
		\label{fig: interpolations LSUNchurch deep}
	\end{figure}
}

\newcommand{\FigInterpolationsCelebAHQ}
{
	\begin{figure}[H]
		\centering
		\begin{tabular*}{\textwidth}{@{\extracolsep{\fill}}ccc}
			$\overbracket{\hspace{0.09\textwidth}}^{\mathrm{input\,1}}$ &
			$\overbracket{\hspace{0.755\textwidth}}^{\mathrm{interpolations}}$ &
			$\overbracket{\hspace{0.09\textwidth}}^{\mathrm{input\,2}}$
		\end{tabular*}
		\begin{tcbraster}[
			raster columns=10,
			raster equal height,
			raster column skip=0pt,
			raster row skip=0pt,
			raster every box/.style={blank},
			]
			\tcbincludegraphics{FigInterpolationsCelebAHQ/s6-3246}
			\tcbincludegraphics{FigInterpolationsCelebAHQ/int6-0}
			\tcbincludegraphics{FigInterpolationsCelebAHQ/int6-2}
			\tcbincludegraphics{FigInterpolationsCelebAHQ/int6-4}
			\tcbincludegraphics{FigInterpolationsCelebAHQ/int6-6}
			\tcbincludegraphics{FigInterpolationsCelebAHQ/int6-8}
			\tcbincludegraphics{FigInterpolationsCelebAHQ/int6-10}
			\tcbincludegraphics{FigInterpolationsCelebAHQ/int6-12}
			\tcbincludegraphics{FigInterpolationsCelebAHQ/int6-14}
			\tcbincludegraphics{FigInterpolationsCelebAHQ/t6-1829}
			\tcbincludegraphics{FigInterpolationsCelebAHQ/s3-260}
			\tcbincludegraphics{FigInterpolationsCelebAHQ/int3-0}
			\tcbincludegraphics{FigInterpolationsCelebAHQ/int3-2}
			\tcbincludegraphics{FigInterpolationsCelebAHQ/int3-4}
			\tcbincludegraphics{FigInterpolationsCelebAHQ/int3-6}
			\tcbincludegraphics{FigInterpolationsCelebAHQ/int3-8}
			\tcbincludegraphics{FigInterpolationsCelebAHQ/int3-10}
			\tcbincludegraphics{FigInterpolationsCelebAHQ/int3-12}
			\tcbincludegraphics{FigInterpolationsCelebAHQ/int3-14}
			\tcbincludegraphics{FigInterpolationsCelebAHQ/t3-3632}
			\tcbincludegraphics{FigInterpolationsCelebAHQ/s0-1035}
			\tcbincludegraphics{FigInterpolationsCelebAHQ/int0-0}
			\tcbincludegraphics{FigInterpolationsCelebAHQ/int0-2}
			\tcbincludegraphics{FigInterpolationsCelebAHQ/int0-4}
			\tcbincludegraphics{FigInterpolationsCelebAHQ/int0-6}
			\tcbincludegraphics{FigInterpolationsCelebAHQ/int0-8}
			\tcbincludegraphics{FigInterpolationsCelebAHQ/int0-10}
			\tcbincludegraphics{FigInterpolationsCelebAHQ/int0-12}
			\tcbincludegraphics{FigInterpolationsCelebAHQ/int0-14}
			\tcbincludegraphics{FigInterpolationsCelebAHQ/t0-3382}
		\end{tcbraster}
		
		\caption[LRDM: CelebA-HQ interpolations (probabilistic sampling)]{
			\begin{minipage}[t]{0.8\textwidth}
				LRDM (VQ-AE) interpolations (slerp) on CelebA-HQ for representation shape $64\times8\times8$, $\lambda=10^{-4}$, using probabilistic sampling and re-sampling $\z_T$ for each sample. The additional stochasticity (compared to the deterministic sampling in \fig{fig: interpolations CelebA-HQ DDIM}) through the probabilistic sampling manifests in variations of small-scale features.
			\end{minipage}
		}
		\label{fig: interpolations CelebA-HQ}
	\end{figure}
}

\newcommand{\FigSamplesReprLSUNchurchDeep}
{
	\begin{figure}[H]
		\centering
		
		\begin{subfigure}[b]{\textwidth}
			\begin{tcbraster}[
				raster columns=10,
				raster equal height, 
				raster column skip=0pt,
				raster row skip=0pt,
				raster every box/.style={blank}
				]
				\tcbincludegraphics{FigSamplesReprLSUNchurchDeep/lambda1e-4/0}
				\tcbincludegraphics{FigSamplesReprLSUNchurchDeep/lambda1e-4/1}
				\tcbincludegraphics{FigSamplesReprLSUNchurchDeep/lambda1e-4/2}
				\tcbincludegraphics{FigSamplesReprLSUNchurchDeep/lambda1e-4/3}
				\tcbincludegraphics{FigSamplesReprLSUNchurchDeep/lambda1e-4/4}
				\tcbincludegraphics{FigSamplesReprLSUNchurchDeep/lambda1e-4/5}
				\tcbincludegraphics{FigSamplesReprLSUNchurchDeep/lambda1e-4/6}
				\tcbincludegraphics{FigSamplesReprLSUNchurchDeep/lambda1e-4/7}
				\tcbincludegraphics{FigSamplesReprLSUNchurchDeep/lambda1e-4/8}
				\tcbincludegraphics{FigSamplesReprLSUNchurchDeep/lambda1e-4/9}
				\tcbincludegraphics{FigSamplesReprLSUNchurchDeep/lambda1e-4/10}
				\tcbincludegraphics{FigSamplesReprLSUNchurchDeep/lambda1e-4/11}
				\tcbincludegraphics{FigSamplesReprLSUNchurchDeep/lambda1e-4/12}
				\tcbincludegraphics{FigSamplesReprLSUNchurchDeep/lambda1e-4/13}
				\tcbincludegraphics{FigSamplesReprLSUNchurchDeep/lambda1e-4/14}
				\tcbincludegraphics{FigSamplesReprLSUNchurchDeep/lambda1e-4/15}
				\tcbincludegraphics{FigSamplesReprLSUNchurchDeep/lambda1e-4/16}
				\tcbincludegraphics{FigSamplesReprLSUNchurchDeep/lambda1e-4/17}
				\tcbincludegraphics{FigSamplesReprLSUNchurchDeep/lambda1e-4/18}
				\tcbincludegraphics{FigSamplesReprLSUNchurchDeep/lambda1e-4/19}
				\tcbincludegraphics{FigSamplesReprLSUNchurchDeep/lambda1e-4/20}
				\tcbincludegraphics{FigSamplesReprLSUNchurchDeep/lambda1e-4/21}
				\tcbincludegraphics{FigSamplesReprLSUNchurchDeep/lambda1e-4/22}
				\tcbincludegraphics{FigSamplesReprLSUNchurchDeep/lambda1e-4/23}
				\tcbincludegraphics{FigSamplesReprLSUNchurchDeep/lambda1e-4/24}
				\tcbincludegraphics{FigSamplesReprLSUNchurchDeep/lambda1e-4/25}
				\tcbincludegraphics{FigSamplesReprLSUNchurchDeep/lambda1e-4/26}
				\tcbincludegraphics{FigSamplesReprLSUNchurchDeep/lambda1e-4/27}
				\tcbincludegraphics{FigSamplesReprLSUNchurchDeep/lambda1e-4/28}
				\tcbincludegraphics{FigSamplesReprLSUNchurchDeep/lambda1e-4/29}
			\end{tcbraster}
			\subcaption{$\lambda = 10^{-4}$ (FID=58.08)}
		\end{subfigure}
		\begin{subfigure}[b]{\textwidth}
			\begin{tcbraster}[
				raster columns=10,
				raster equal height, 
				raster column skip=0pt,
				raster row skip=0pt,
				raster every box/.style={blank}
				]
				\tcbincludegraphics{FigSamplesReprLSUNchurchDeep/lambda1e-3/0}
				\tcbincludegraphics{FigSamplesReprLSUNchurchDeep/lambda1e-3/1}
				\tcbincludegraphics{FigSamplesReprLSUNchurchDeep/lambda1e-3/2}
				\tcbincludegraphics{FigSamplesReprLSUNchurchDeep/lambda1e-3/3}
				\tcbincludegraphics{FigSamplesReprLSUNchurchDeep/lambda1e-3/4}
				\tcbincludegraphics{FigSamplesReprLSUNchurchDeep/lambda1e-3/5}
				\tcbincludegraphics{FigSamplesReprLSUNchurchDeep/lambda1e-3/6}
				\tcbincludegraphics{FigSamplesReprLSUNchurchDeep/lambda1e-3/7}
				\tcbincludegraphics{FigSamplesReprLSUNchurchDeep/lambda1e-3/8}
				\tcbincludegraphics{FigSamplesReprLSUNchurchDeep/lambda1e-3/9}
				\tcbincludegraphics{FigSamplesReprLSUNchurchDeep/lambda1e-3/10}
				\tcbincludegraphics{FigSamplesReprLSUNchurchDeep/lambda1e-3/11}
				\tcbincludegraphics{FigSamplesReprLSUNchurchDeep/lambda1e-3/12}
				\tcbincludegraphics{FigSamplesReprLSUNchurchDeep/lambda1e-3/13}
				\tcbincludegraphics{FigSamplesReprLSUNchurchDeep/lambda1e-3/14}
				\tcbincludegraphics{FigSamplesReprLSUNchurchDeep/lambda1e-3/15}
				\tcbincludegraphics{FigSamplesReprLSUNchurchDeep/lambda1e-3/16}
				\tcbincludegraphics{FigSamplesReprLSUNchurchDeep/lambda1e-3/17}
				\tcbincludegraphics{FigSamplesReprLSUNchurchDeep/lambda1e-3/18}
				\tcbincludegraphics{FigSamplesReprLSUNchurchDeep/lambda1e-3/19}
				\tcbincludegraphics{FigSamplesReprLSUNchurchDeep/lambda1e-3/20}
				\tcbincludegraphics{FigSamplesReprLSUNchurchDeep/lambda1e-3/21}
				\tcbincludegraphics{FigSamplesReprLSUNchurchDeep/lambda1e-3/22}
				\tcbincludegraphics{FigSamplesReprLSUNchurchDeep/lambda1e-3/23}
				\tcbincludegraphics{FigSamplesReprLSUNchurchDeep/lambda1e-3/24}
				\tcbincludegraphics{FigSamplesReprLSUNchurchDeep/lambda1e-3/25}
				\tcbincludegraphics{FigSamplesReprLSUNchurchDeep/lambda1e-3/26}
				\tcbincludegraphics{FigSamplesReprLSUNchurchDeep/lambda1e-3/27}
				\tcbincludegraphics{FigSamplesReprLSUNchurchDeep/lambda1e-3/28}
				\tcbincludegraphics{FigSamplesReprLSUNchurchDeep/lambda1e-3/29}
			\end{tcbraster}
			\subcaption{$\lambda = 10^{-3}$ (FID=8.39)}
		\end{subfigure}
		\begin{subfigure}[b]{\textwidth}
			\begin{tcbraster}[
				raster columns=10,
				raster equal height, 
				raster column skip=0pt,
				raster row skip=0pt,
				raster every box/.style={blank}
				]
				\tcbincludegraphics{FigSamplesReprLSUNchurchDeep/lambda1e-1/0}
				\tcbincludegraphics{FigSamplesReprLSUNchurchDeep/lambda1e-1/1}
				\tcbincludegraphics{FigSamplesReprLSUNchurchDeep/lambda1e-1/2}
				\tcbincludegraphics{FigSamplesReprLSUNchurchDeep/lambda1e-1/3}
				\tcbincludegraphics{FigSamplesReprLSUNchurchDeep/lambda1e-1/4}
				\tcbincludegraphics{FigSamplesReprLSUNchurchDeep/lambda1e-1/5}
				\tcbincludegraphics{FigSamplesReprLSUNchurchDeep/lambda1e-1/6}
				\tcbincludegraphics{FigSamplesReprLSUNchurchDeep/lambda1e-1/7}
				\tcbincludegraphics{FigSamplesReprLSUNchurchDeep/lambda1e-1/8}
				\tcbincludegraphics{FigSamplesReprLSUNchurchDeep/lambda1e-1/9}
				\tcbincludegraphics{FigSamplesReprLSUNchurchDeep/lambda1e-1/10}
				\tcbincludegraphics{FigSamplesReprLSUNchurchDeep/lambda1e-1/11}
				\tcbincludegraphics{FigSamplesReprLSUNchurchDeep/lambda1e-1/12}
				\tcbincludegraphics{FigSamplesReprLSUNchurchDeep/lambda1e-1/13}
				\tcbincludegraphics{FigSamplesReprLSUNchurchDeep/lambda1e-1/14}
				\tcbincludegraphics{FigSamplesReprLSUNchurchDeep/lambda1e-1/15}
				\tcbincludegraphics{FigSamplesReprLSUNchurchDeep/lambda1e-1/16}
				\tcbincludegraphics{FigSamplesReprLSUNchurchDeep/lambda1e-1/17}
				\tcbincludegraphics{FigSamplesReprLSUNchurchDeep/lambda1e-1/18}
				\tcbincludegraphics{FigSamplesReprLSUNchurchDeep/lambda1e-1/19}
				\tcbincludegraphics{FigSamplesReprLSUNchurchDeep/lambda1e-1/20}
				\tcbincludegraphics{FigSamplesReprLSUNchurchDeep/lambda1e-1/21}
				\tcbincludegraphics{FigSamplesReprLSUNchurchDeep/lambda1e-1/22}
				\tcbincludegraphics{FigSamplesReprLSUNchurchDeep/lambda1e-1/23}
				\tcbincludegraphics{FigSamplesReprLSUNchurchDeep/lambda1e-1/24}
				\tcbincludegraphics{FigSamplesReprLSUNchurchDeep/lambda1e-1/25}
				\tcbincludegraphics{FigSamplesReprLSUNchurchDeep/lambda1e-1/26}
				\tcbincludegraphics{FigSamplesReprLSUNchurchDeep/lambda1e-1/27}
				\tcbincludegraphics{FigSamplesReprLSUNchurchDeep/lambda1e-1/28}
				\tcbincludegraphics{FigSamplesReprLSUNchurchDeep/lambda1e-1/29}
			\end{tcbraster}
			\subcaption{$\lambda = 10^{-1}$ (FID=8.00)}
		\end{subfigure}
		
		\caption[LRDM: LSUN-Churches samples for different regularizations]{
			\begin{minipage}[t]{0.8\textwidth}
				Unconditional samples from the LRDM (VQ-AE) for different regularization factors. Increasing $\lambda$ leads to a representation distribution closer to the gaussian prior.
			\end{minipage}
		}
		\label{fig: samples LRDM different lambdas}
	\end{figure}
}

\newcommand{\FigRecReprLSUNchurchDeep}
{
	\newgeometry{top=3cm}
	\begin{figure}[H]
		\centering
		\begin{subfigure}[b]{\textwidth}
			\begin{tcbraster}[
				raster columns=10,
				raster equal height, 
				raster column skip=0pt,
				raster row skip=0pt,
				raster every box/.style={blank}
				]
				\tcbincludegraphics{FigRecReprLSUNchurchDeep/input0-4141}
				\tcbincludegraphics{FigRecReprLSUNchurchDeep/input1-1465}
				\tcbincludegraphics{FigRecReprLSUNchurchDeep/input2-1805}
				\tcbincludegraphics{FigRecReprLSUNchurchDeep/input3-2749}
				\tcbincludegraphics{FigRecReprLSUNchurchDeep/input4-1317}
				\tcbincludegraphics{FigRecReprLSUNchurchDeep/input5-2716}
				\tcbincludegraphics{FigRecReprLSUNchurchDeep/input6-3185}
				\tcbincludegraphics{FigRecReprLSUNchurchDeep/input7-972}
				\tcbincludegraphics{FigRecReprLSUNchurchDeep/input8-2916}
				\tcbincludegraphics{FigRecReprLSUNchurchDeep/input9-1891}
			\end{tcbraster}
		\subcaption{Input images}
		\end{subfigure}
		
		\begin{subfigure}[b]{\textwidth}
			\begin{tcbraster}[
				raster columns=10,
				raster equal height, 
				raster column skip=0pt,
				raster row skip=0pt,
				raster every box/.style={blank}
				]
				\tcbincludegraphics{FigRecReprLSUNchurchDeep/lambda1e-4/rec0-0}
				\tcbincludegraphics{FigRecReprLSUNchurchDeep/lambda1e-4/rec1-0}
				\tcbincludegraphics{FigRecReprLSUNchurchDeep/lambda1e-4/rec2-0}
				\tcbincludegraphics{FigRecReprLSUNchurchDeep/lambda1e-4/rec3-0}
				\tcbincludegraphics{FigRecReprLSUNchurchDeep/lambda1e-4/rec4-0}
				\tcbincludegraphics{FigRecReprLSUNchurchDeep/lambda1e-4/rec5-0}
				\tcbincludegraphics{FigRecReprLSUNchurchDeep/lambda1e-4/rec6-0}
				\tcbincludegraphics{FigRecReprLSUNchurchDeep/lambda1e-4/rec7-0}
				\tcbincludegraphics{FigRecReprLSUNchurchDeep/lambda1e-4/rec8-0}
				\tcbincludegraphics{FigRecReprLSUNchurchDeep/lambda1e-4/rec9-0}
				\tcbincludegraphics{FigRecReprLSUNchurchDeep/lambda1e-4/rec0-1}
				\tcbincludegraphics{FigRecReprLSUNchurchDeep/lambda1e-4/rec1-1}
				\tcbincludegraphics{FigRecReprLSUNchurchDeep/lambda1e-4/rec2-1}
				\tcbincludegraphics{FigRecReprLSUNchurchDeep/lambda1e-4/rec3-1}
				\tcbincludegraphics{FigRecReprLSUNchurchDeep/lambda1e-4/rec4-1}
				\tcbincludegraphics{FigRecReprLSUNchurchDeep/lambda1e-4/rec5-1}
				\tcbincludegraphics{FigRecReprLSUNchurchDeep/lambda1e-4/rec6-1}
				\tcbincludegraphics{FigRecReprLSUNchurchDeep/lambda1e-4/rec7-1}
				\tcbincludegraphics{FigRecReprLSUNchurchDeep/lambda1e-4/rec8-1}
				\tcbincludegraphics{FigRecReprLSUNchurchDeep/lambda1e-4/rec9-1}
				\tcbincludegraphics{FigRecReprLSUNchurchDeep/lambda1e-4/rec0-2}
				\tcbincludegraphics{FigRecReprLSUNchurchDeep/lambda1e-4/rec1-2}
				\tcbincludegraphics{FigRecReprLSUNchurchDeep/lambda1e-4/rec2-2}
				\tcbincludegraphics{FigRecReprLSUNchurchDeep/lambda1e-4/rec3-2}
				\tcbincludegraphics{FigRecReprLSUNchurchDeep/lambda1e-4/rec4-2}
				\tcbincludegraphics{FigRecReprLSUNchurchDeep/lambda1e-4/rec5-2}
				\tcbincludegraphics{FigRecReprLSUNchurchDeep/lambda1e-4/rec6-2}
				\tcbincludegraphics{FigRecReprLSUNchurchDeep/lambda1e-4/rec7-2}
				\tcbincludegraphics{FigRecReprLSUNchurchDeep/lambda1e-4/rec8-2}
				\tcbincludegraphics{FigRecReprLSUNchurchDeep/lambda1e-4/rec9-2}
			\end{tcbraster}
			\subcaption{$\lambda = 10^{-4}$}
		\end{subfigure}
		\begin{subfigure}[b]{\textwidth}
			\begin{tcbraster}[
				raster columns=10,
				raster equal height, 
				raster column skip=0pt,
				raster row skip=0pt,
				raster every box/.style={blank}
				]
				\tcbincludegraphics{FigRecReprLSUNchurchDeep/lambda1e-3/rec0-0}
				\tcbincludegraphics{FigRecReprLSUNchurchDeep/lambda1e-3/rec1-0}
				\tcbincludegraphics{FigRecReprLSUNchurchDeep/lambda1e-3/rec2-0}
				\tcbincludegraphics{FigRecReprLSUNchurchDeep/lambda1e-3/rec3-0}
				\tcbincludegraphics{FigRecReprLSUNchurchDeep/lambda1e-3/rec4-0}
				\tcbincludegraphics{FigRecReprLSUNchurchDeep/lambda1e-3/rec5-0}
				\tcbincludegraphics{FigRecReprLSUNchurchDeep/lambda1e-3/rec6-0}
				\tcbincludegraphics{FigRecReprLSUNchurchDeep/lambda1e-3/rec7-0}
				\tcbincludegraphics{FigRecReprLSUNchurchDeep/lambda1e-3/rec8-0}
				\tcbincludegraphics{FigRecReprLSUNchurchDeep/lambda1e-3/rec9-0}
				\tcbincludegraphics{FigRecReprLSUNchurchDeep/lambda1e-3/rec0-1}
				\tcbincludegraphics{FigRecReprLSUNchurchDeep/lambda1e-3/rec1-1}
				\tcbincludegraphics{FigRecReprLSUNchurchDeep/lambda1e-3/rec2-1}
				\tcbincludegraphics{FigRecReprLSUNchurchDeep/lambda1e-3/rec3-1}
				\tcbincludegraphics{FigRecReprLSUNchurchDeep/lambda1e-3/rec4-1}
				\tcbincludegraphics{FigRecReprLSUNchurchDeep/lambda1e-3/rec5-1}
				\tcbincludegraphics{FigRecReprLSUNchurchDeep/lambda1e-3/rec6-1}
				\tcbincludegraphics{FigRecReprLSUNchurchDeep/lambda1e-3/rec7-1}
				\tcbincludegraphics{FigRecReprLSUNchurchDeep/lambda1e-3/rec8-1}
				\tcbincludegraphics{FigRecReprLSUNchurchDeep/lambda1e-3/rec9-1}
				\tcbincludegraphics{FigRecReprLSUNchurchDeep/lambda1e-3/rec0-2}
				\tcbincludegraphics{FigRecReprLSUNchurchDeep/lambda1e-3/rec1-2}
				\tcbincludegraphics{FigRecReprLSUNchurchDeep/lambda1e-3/rec2-2}
				\tcbincludegraphics{FigRecReprLSUNchurchDeep/lambda1e-3/rec3-2}
				\tcbincludegraphics{FigRecReprLSUNchurchDeep/lambda1e-3/rec4-2}
				\tcbincludegraphics{FigRecReprLSUNchurchDeep/lambda1e-3/rec5-2}
				\tcbincludegraphics{FigRecReprLSUNchurchDeep/lambda1e-3/rec6-2}
				\tcbincludegraphics{FigRecReprLSUNchurchDeep/lambda1e-3/rec7-2}
				\tcbincludegraphics{FigRecReprLSUNchurchDeep/lambda1e-3/rec8-2}
				\tcbincludegraphics{FigRecReprLSUNchurchDeep/lambda1e-3/rec9-2}
			\end{tcbraster}
			\subcaption{$\lambda = 10^{-3}$}
		\end{subfigure}
		\begin{subfigure}[b]{\textwidth}
			\begin{tcbraster}[
				raster columns=10,
				raster equal height, 
				raster column skip=0pt,
				raster row skip=0pt,
				raster every box/.style={blank}
				]
				\tcbincludegraphics{FigRecReprLSUNchurchDeep/lambda1e-1/rec0-0}
				\tcbincludegraphics{FigRecReprLSUNchurchDeep/lambda1e-1/rec1-0}
				\tcbincludegraphics{FigRecReprLSUNchurchDeep/lambda1e-1/rec2-0}
				\tcbincludegraphics{FigRecReprLSUNchurchDeep/lambda1e-1/rec3-0}
				\tcbincludegraphics{FigRecReprLSUNchurchDeep/lambda1e-1/rec4-0}
				\tcbincludegraphics{FigRecReprLSUNchurchDeep/lambda1e-1/rec5-0}
				\tcbincludegraphics{FigRecReprLSUNchurchDeep/lambda1e-1/rec6-0}
				\tcbincludegraphics{FigRecReprLSUNchurchDeep/lambda1e-1/rec7-0}
				\tcbincludegraphics{FigRecReprLSUNchurchDeep/lambda1e-1/rec8-0}
				\tcbincludegraphics{FigRecReprLSUNchurchDeep/lambda1e-1/rec9-0}
				\tcbincludegraphics{FigRecReprLSUNchurchDeep/lambda1e-1/rec0-1}
				\tcbincludegraphics{FigRecReprLSUNchurchDeep/lambda1e-1/rec1-1}
				\tcbincludegraphics{FigRecReprLSUNchurchDeep/lambda1e-1/rec2-1}
				\tcbincludegraphics{FigRecReprLSUNchurchDeep/lambda1e-1/rec3-1}
				\tcbincludegraphics{FigRecReprLSUNchurchDeep/lambda1e-1/rec4-1}
				\tcbincludegraphics{FigRecReprLSUNchurchDeep/lambda1e-1/rec5-1}
				\tcbincludegraphics{FigRecReprLSUNchurchDeep/lambda1e-1/rec6-1}
				\tcbincludegraphics{FigRecReprLSUNchurchDeep/lambda1e-1/rec7-1}
				\tcbincludegraphics{FigRecReprLSUNchurchDeep/lambda1e-1/rec8-1}
				\tcbincludegraphics{FigRecReprLSUNchurchDeep/lambda1e-1/rec9-1}
				\tcbincludegraphics{FigRecReprLSUNchurchDeep/lambda1e-1/rec0-2}
				\tcbincludegraphics{FigRecReprLSUNchurchDeep/lambda1e-1/rec1-2}
				\tcbincludegraphics{FigRecReprLSUNchurchDeep/lambda1e-1/rec2-2}
				\tcbincludegraphics{FigRecReprLSUNchurchDeep/lambda1e-1/rec3-2}
				\tcbincludegraphics{FigRecReprLSUNchurchDeep/lambda1e-1/rec4-2}
				\tcbincludegraphics{FigRecReprLSUNchurchDeep/lambda1e-1/rec5-2}
				\tcbincludegraphics{FigRecReprLSUNchurchDeep/lambda1e-1/rec6-2}
				\tcbincludegraphics{FigRecReprLSUNchurchDeep/lambda1e-1/rec7-2}
				\tcbincludegraphics{FigRecReprLSUNchurchDeep/lambda1e-1/rec8-2}
				\tcbincludegraphics{FigRecReprLSUNchurchDeep/lambda1e-1/rec9-2}
			\end{tcbraster}
			\subcaption{$\lambda = 10^{-1}$}
		\end{subfigure}
		
		\caption[LRDM: LSUN-Churches reconstructions for different $\lambda$]{
			\begin{minipage}[t]{0.8\textwidth}
				Image reconstructions from the LRDM (VQ-AE) for different regularization factors. $\z_T$ is re-sampled for each reconstruction. For low $\lambda$, the representation allows for faithful reconstructions, varying $\z_T$ incudes variations in local image features. For intermediate $\lambda$, only the overall shape and color scheme is encoded. For high $\lambda$, the LRDM approaches the LDM.
			\end{minipage}
		}
		\label{fig: reconstructions LRDM different lambdas}
	\end{figure}
	\restoregeometry
}

\newcommand{\TabZvsEps}
{
	\begin{table}[H]
		\centering
		\begin{tabular}{lcccccc}\toprule
			Model & param. & dropout & FID $\downarrow$ & & IS $\uparrow$ & \\ \midrule
			DDPM$^\ast$ \cite{ho_denoising_2020} & noise & 0.0 & 7.89 & & - & \\
			LDM-8$^\ast$ (KL-AE) \cite{rombach_high-resolution_2021} & noise & - & 4.02 & & - & \\
			StyleGAN2$^\ast$ \cite{karras_analyzing_2020} & - & - & 3.86 & & - & \\
			\midrule \vspace{-1.5em}\\ \midrule
			& & & 100e & 200e & 100e & 200e \\ \midrule
			LDM (KL-AE \cite{rombach_high-resolution_2021}) & noise & 0.2 & 6.08 & 5.23 & 2.56 & 2.60 \\
			& & 0.1 & 6.78 & 5.70 & 2.55 & 2.62 \\
			& image & 0.2 & - & 10.72 & - & 2.41 \\ \midrule
			LDM (VQ-AE \cite{rombach_high-resolution_2021}) & noise & 0.1 & 6.00 & 5.38 & 2.59 & 2.64 \\
			& image & 0.2 & 8.27 & 6.50 & 2.44 & 2.54 \\
			\bottomrule
		\end{tabular}
		\caption[LDM: Unconditional image synthesis]{
			\begin{minipage}[t]{0.8\textwidth}
				Evaluation metrics for unconditional image synthesis. Noise-parameterized LDMs were trained with $L_\mathrm{noise}^\ast$, image-parameterized models with $L_\mathrm{image}^\ast$. Our models were evaluated after 100 and 200 epochs. For models marked with $^\ast$, the scores are presented as given in the respective papers. The image-parameterized LDM (VQ-AE) achieves competitive FID scores after rather short training times.
			\end{minipage}
		}
		\label{tab: FID z0 vs eps}
	\end{table}
}

\newcommand{\TabReprStatsLSUNchurchCombined}
{
	\begin{table}[H]
		\centering
		\begin{tabular}{llcccccc}\toprule
			model & $\lambda$ & FID & recFID & MSE & RMSE & Var & Var* \\ \midrule
			LRDM & $10^{-6}$ & 457.70 & 6.03 & 0.04 & 0.19 & 0.01 & 0.01 \\
			($16\times16\times16$) & $10^{-4}$ & 58.08 & 8.63 & 0.07 & 0.27 & 0.03 & 0.05 \\
			& $5\cdot 10^{-4}$ & 10.54 & 13.30 & 0.17 & 0.41 & 0.08 & 0.11 \\
			& $10^{-3\,\dagger}$ & 10.45 & 14.62 & 0.23 & 0.48 & 0.11 & 0.14 \\
			& $10^{-3}$ & 8.39 & 12.90 & 0.23 & 0.48 & 0.11 & 0.15 \\
			& $10^{-1}$ & 8.00 & 12.53 & 0.53 & 0.73 & 0.28 & 0.29 \\ \midrule 
			VQ-AE & - & - & 3.75 & 0.01 & 0.12 & 0.00 & - \\
			LVAE & $10^{-6}$ & 315.54 & 8.97 & 0.02 & 0.16 & 1e{-}6 & 3e{-}4 \\ \midrule
			LRDM & $10^{-4}$ & 46.84 & 9.31 & 0.08 & 0.29 & 0.03 & 0.05 \\
			($64\times8\times8$) & $5\cdot 10^{-4}$ & 8.89 & 11.75 & 0.18 & 0.42 & 0.08 & 0.11 \\
			& $10^{-3}$ & 8.04 & 12.29 & 0.23 & 0.48 & 0.11 & 0.13 \\
			& $10^{-1}$ & 7.65 & 12.30 & 0.54 & 0.73 & 0.28 & 0.29 \\ \midrule
			t-LRDM & $10^{-4}$ &  106.90 & 8.84 & 0.10 & 0.32 & 0.05 & - \\
			($16\times16\times16$) & $10^{-3}$ & 12.34 & 13.24 & 0.26 & 0.51 & 0.14 & - \\
			\bottomrule
		\end{tabular}
		\caption[LRDM: Sampling and reconstruction quality]{
			\begin{minipage}[t]{0.8\textwidth}
				Sampling and reconstruction quality for various representation regularizations $\lambda$ for LSUN-Churches. All models but the VQ-AE are trained for 100 epochs. $^\dagger$ denotes models trained with relaxed timestep-masking. Var$^\ast$ is calculated using probabilistic sampling. For a visual depiction, see \fig{fig: lambda sweep stats}.
			\end{minipage}
		}
		\label{tab: repr metrics}
	\end{table}
}

\newcommand{\TabHyperparameters}
{
	\newpage
	\thispagestyle{empty}
	\newgeometry{left=-5cm,bottom=-2cm,top=0cm,right=-3.5cm}
	\begin{table}[b]
		\centering
		\begin{tabular}{lcccccc}\toprule
			& LDM & \begin{tabular}[t]{@{}c@{}}LRDM \\ ($16\times16\times16$)\end{tabular} & \begin{tabular}[t]{@{}c@{}}LRDM \\ ($64\times8\times8$)\end{tabular} &
			\begin{tabular}[t]{@{}c@{}}t-LRDM \\ ($16\times16\times16$)\end{tabular} &
			\begin{tabular}[t]{@{}c@{}}LRDM \\ (MNIST)\end{tabular} &
			\begin{tabular}[t]{@{}c@{}}Warp-LRDM \\ ($4\times32\times32$)\end{tabular} \\ \midrule
			\textbf{Diffusion Model Parameters} & & & & & & \\ \midrule
			1st-stage model & VQ-AE-$f4$ & VQ-AE-$f4$ & VQ-AE-$f4$ & VQ-AE-$f4$ & - & VQ-AE-$f4$ \\
			Diffusion steps & 1000 & 1000 & 1000 & 1000 & 1000 & 1000 \\
			Noise schedule & linear & linear & linear & linear & linear & linear \\
			Channels & 128 & 128 & 128 & 128 & 32 & 128 \\
			Channel Multiplier & [1, 2, 4] & [1, 2, 4] & [1, 2, 2, 2] & [1, 2, 4] & [1, 2, 2] & [1, 4] \\
			Residual Blocks per Layer & 2 & 2 & 3 & 2 & 2 & 2 \\
			Attention resolutions & [32, 16] & [32, 16, 8] & [32, 16] & [32, 16] & [14, 7] & [32] \\
			Head Channels & 64 & 64 & 64 & 64 & all & 64 \\
			Dropout & 0.2 & 0.2 & 0.2 & 0.2 & 0.1 & 0.2 \\
			Class-conditional & \xmark & \xmark & \xmark & \xmark & \checkmark & \xmark \\
			& & & & & & \\ \midrule
			\textbf{Training Parameters} & & & & & & \\ \midrule
			Batch Size & 256 & 256 & 256 & 256 & 256 & 256 \\
			Learning Rate & 1e-4 & 1e-4 & 1e-4 & 1e-4 & 1e-4 & 1e-4 \\
			$N_\mathrm{params,\,train}$ & 71.8\,M & 110\,M & 110\,M & 110\,M & 3.7\,M & 91\,M \\
			$N_\mathrm{params,\,total}$ & 144\,M & 182\,M & 182\,M & 182\,M & 3.7\,M & 163\,M \\ \midrule
			\textbf{Representation Model Parameters} & & & & & & \\ \midrule
			Channels & - & 128 & 128 & 128 & 32 & 128 \\
			Channel Multiplier & - & [1, 2, 2] & [1, 2, 2, 2] & [1, 2, 2] & [1, 2, 2] & [1, 2] \\
			Residual Blocks per Layer & - & 4 & 4 & 4 & 2 & 4 \\
			Attention resolutions (1 head) & - & [32, 16] & [32, 16, 8] & [32, 16] & [14, 7] & [32] \\
			Dropout & - & 0.1 & 0.1 & 0.1 & - & 0.1 \\
			Class-conditional & - & \xmark & \xmark & \xmark & \checkmark & \xmark \\
			Timestep-conditional & - & \xmark & \xmark & \checkmark & \xmark & \xmark \\
			& & & & & & \\ \bottomrule
		\end{tabular}
		\captionsetup{justification=centering}
		\caption[Hyperparameters]{Hyperparameters for all models.}
		\label{tab: hyperparameters}
	\end{table}
	\restoregeometry
}

\begin{document}

\pagenumbering{gobble}

\begin{titlepage}
	\begin{center}
		\Large\textbf{Representation Learning with Diffusion Models}
		
		\vspace{8cm}
		
		\linespread{1.5}
		{
			\normalsize
			This Master thesis has been carried out by Jeremias Traub\\
			at the\\
			IWR, Heidelberg University\\
			under the supervision of\\
			Prof.\ Dr.\ Björn Ommer\\
			(Ludwig Maximilian University of Munich \& IWR, Heidelberg University)\\
			and\\
			Robin Rombach\\
			(Ludwig Maximilian University of Munich \& IWR, Heidelberg University)\\
			and\\
			Prof.\ Dr.\ Tilman Plehn\\
			(Institute for Theoretical Physics, Heidelberg University)
		}
		\vfill
	\end{center}
	
\end{titlepage}

\newpage
\textbf{Abstract}\\
\vspace{-0.5em}\\

Diffusion models (DMs) have achieved state-of-the-art results for image synthesis tasks as well as density estimation. Applied in the latent space of a powerful pretrained autoencoder (LDM), their immense computational requirements can be significantly reduced without sacrificing sampling quality.
However, DMs and LDMs lack a semantically meaningful representation space as the diffusion process gradually destroys information in the latent variables.
We introduce a framework for learning such representations with diffusion models (LRDM). To that end, a LDM is conditioned on the representation extracted from the clean image by a separate encoder. In particular, the DM and the representation encoder are trained jointly in order to learn rich representations specific to the generative denoising process.
By introducing a tractable representation prior, we can efficiently sample from the representation distribution for unconditional image synthesis without training of any additional model.
We demonstrate that i) competitive image generation results can be achieved with image-parameterized LDMs, ii) LRDMs are capable of learning semantically meaningful representations, allowing for faithful image reconstructions and semantic interpolations.
Our implementation is available at \href{https://github.com/jeremiastraub/diffusion}{https://github.com/jeremiastraub/diffusion}.

\newpage
\pagenumbering{roman}
\tableofcontents
\pagenumbering{arabic}

\newpage
\chapter{Introduction}\label{ch:introduction}
As humans, we can visually perceive the world around us, recognize objects, and imagine new scenes with great ease.
The actual complexity of the task becomes apparent when trying to make a computer understand images. To computers, images are a big pile of numbers, spatially arranged on a pixel grid with color represented as tuple of 3 numbers in the RGB scheme. Enabling computational models to make sense of, and to solve various tasks on image data is the key challenge in the field of Computer Vision. It comprises many image-related tasks such as image classification, object detection and image segmentation but also that of image synthesis and representation learning. This thesis addresses the topics of visual synthesis (see \sec{sec:visual synthesis}) and representation learning (see \sec{sec:representation learning}). In particular, we tackle the problem of learning meaningful representations with diffusion models \cite{sohl-dickstein_deep_2015, ho_denoising_2020} which have recently shown impressive visual synthesis results \cite{dhariwal_diffusion_2021}.\\

When we visually perceive something, this does not only involve a sensory process but also, and most importantly, the interpretation through the brain, which makes use of previously acquired concepts to make sense of the received signal \cite{zeithamova_brain_2019}. Deep neural networks (DNNs) have proven to be a powerful tool for modeling the latter \cite{lake_building_2017}. Due to their deep architecture, they are capable of capturing more abstract features, i.e., high-level interdependencies and patterns, when trained on a large amount of training images \cite{lecun_deep_2015}.\\

The thesis is structured as follows:
\sec{ch:methods} provides an overview of the relevant machine learning methods, particularly of variational autoencoders (\sec{sec:vae}) and diffusion models (\sec{sec:diffusion models}). \sec{ch:related work} discusses the connections of this work to related literature. The results for our experiments are presented in \sec{ch:experiments}.

\section{Visual Synthesis}\label{sec:visual synthesis}
The task of Visual Synthesis aims at understanding images by learning to create new, unseen images from the data distribution. The corresponding class of models is called deep generative models. More precisely, given a large amount of training images that follow a distribution $p(\x)$, we want to be able to generate new images that could have been samples from the true data distribution. To that end, generative models either explicitly (e.g.\ variational autoencoders \cite{kingma_auto-encoding_2014}) or implicitly (e.g.\ generative adversarial networks \cite{goodfellow_generative_2014}) perform density estimation such that the model distribution $p_\theta(\x)$ approximates the true distribution $p(\x)$.\\

\FigDALLEsamples

In this work, we employ diffusion models (DMs) \cite{sohl-dickstein_deep_2015, ho_denoising_2020} for image synthesis. DMs have achieved state-of-the-art results in image synthesis \cite{dhariwal_diffusion_2021} and density estimation \cite{kingma_variational_2021}, and have shown great flexibility for various conditional synthesis tasks \cite{dhariwal_diffusion_2021, rombach_high-resolution_2021, ramesh_hierarchical_2022}. For example, the DALL-E 2 model is based on DMs for synthesizing images based on a text caption (see \fig{fig: dalle2}).

\section{Representation Learning}\label{sec:representation learning}
An underlying idea of representation learning is that real image data lies on a lower-dimensional manifold of the pixel-space \cite{bengio_representation_2014}, i.e., that the data distribution has a reduced \textit{effective} dimensionality. In other words, we assume that there is a set of underlying generative factors that capture the semantics of an image \cite{liu_learning_2022, locatello_challenging_2019}. There are different approaches to learning such representations. A supervised classification model using a DNN learns to map images to a task-specific representation space in the deeper layers where regression can be employed to classify the image \cite{krizhevsky_imagenet_2017}. Contrastive learning is a self-supervised \cite{chen_simple_2020} or supervised \cite{khosla_supervised_2021} approach where a representation space is learnt through learning similarities and differences between images \cite{radford_learning_2021}. Often, however, representation learning is closely linked to generative modeling (i.e., visual synthesis). That way, the learnt representation is directly evaluated during training in terms of its capabilities for recovering an input image from its representation (through a reconstruction loss, as in VAEs \cite{kingma_auto-encoding_2014}) or for creating new, plausible images (through an adversarial loss, as in GANs \cite{goodfellow_generative_2014}). Generative models coupled with a meaningful representation space allow for controlled synthesis, such that interpolations in the representation space imply semantic interpolations in the corresponding generated images. Diffusion models \cite{sohl-dickstein_deep_2015, ho_denoising_2020} are a type of generative model that, by design, lack a meaningful representation space. On the other hand, DMs are powerful generative models, partly due to their particular inductive bias for image-like data \cite{rombach_high-resolution_2021}. Hence, we seek to extract a just as powerful and rich representation – specific to the generative denoising process – which in turn allows for control over image semantics for synthesis tasks as well as for usage in downstream tasks such as classification. Unfortunately, DMs \cite{sohl-dickstein_deep_2015, ho_denoising_2020}, as such, do not provide a meaningful representation space. In contrast, the generative power lies in the learnt denoising steps in between the thousands of latent variables. In this work, we introduce a framework to equip diffusion models with a semantically structured representation space.\\

\newpage
\noindent In summary, this work makes the following contributions:
\begin{itemize}[noitemsep, topsep=4pt, left=1.5em,]
	\item[(i)] We revisit the image-parameterization for DMs, discuss differences to the commonly used noise-parameterization, and demonstrate that competitive FIDs can be achieved with image-parameterized LDMs.
	\item[(ii)] We introduce a framework for learning semantically meaningful representations with diffusion models (LRDM). In contrast to previous work \cite{preechakul_diffusion_2021}, we can efficiently sample from a tractable representation prior for unconditional image synthesis, without training an extra model.
	\item[(iii)] Further, we extend our framework to timestep-conditional (t-LRDM) and class-conditional representation learning. Additionally, we propose a framework for separating style and shape information.
\end{itemize}

\newpage
\chapter{Methods}\label{ch:methods}

This chapter gives an overview of Variational Autoencoders (\sec{sec:vae}) and Diffusion Models (\sec{sec:diffusion models}), which lay the foundation of this thesis.

\section{Variational Autoencoders}\label{sec:vae}
The goal of generative machine learning approaches is to model the data distribution $p(\x)$. The Variational Autoencoder (VAE) \cite{kingma_auto-encoding_2014} does so by learning to reconstruct input images from a compressed latent code. An underlying idea of the model is that real world images can be represented by a relatively small set of higher-lever features, i.e., that the data lies on a lower-dimensional manifold in the image space. Expressed in the framework of probabilistic graphical models, it is assumed that the observed (i.i.d.) data distribution $p(\x)$ is generated by some random process from an unobserved random variable $\z$ (see \fig{fig: overview vae}):

\begin{equation}\label{eq:vae data distr}
	p(\x) = \int\! p(\x \mid \z) p(\z) \mathrm{d}\!\z
\end{equation}

This expression, and thus also the posterior density

\begin{equation}
	p(\z \mid \x) = \frac{p(\x\mid \z)p(\z)}{p(\x)}
\end{equation}

are usually intractable without making strong assumptions about the likelihood function $p(\x \mid \z)$, which we do not want to do here in order to train an expressive model. Instead of resorting to sampling-based inference methods (e.g.\ MCMC), we merely assume the prior $p(\z) = p_\theta(\z)$ and the likelihood $p(\x \mid \z) = p_\theta(\x \mid \z)$ to come from a parameterized families of distributions – allowing us to employ NNs as function approximators – and then apply the concept of variational inference.\\

We then perform maximum likelihood estimation to find the optimal parameters $\theta$. This means that we seek those parameters $\theta$ for which the training data is most probable given the statistical model $p_\theta(\x)$, i.e., which maximize the \mbox{(log-)likelihood} of the data. For that, we first introduce a parameterized encoder $q_\phi(\z \mid \x)$ – which, again, can be approximated by a NN – to have a tractable inference model at hand. We will see in the following that we can then avoid computing the intractable integral in \eq{eq:vae data distr} by optimizing a lower bound of the log-likelihood.\\

\FigOverviewVAE

We rewrite the log-likelihood of the generative model as follows:
\begin{align}
	\log p_\theta(\x) &= \log \int\!  p_\theta(\x, \z) \mathrm{d}\!\z\\
	&= \log \int\! q_\phi(\z \mid \x) \frac{p_\theta(\x, \z)}{q_\phi(\z \mid \x)} \mathrm{d}\!\z\\
	&\overset{(\ast)}{\geq} \int\! q_\phi(\z \mid \x) \log\frac{p_\theta(\x, \z)}{q_\phi(\z \mid \x)} \mathrm{d}\!\z \label{eq:Jensen}\\
	&= \int\! q_\phi(\z \mid \x) \log\frac{p_\theta(\z \mid \x)p_\theta(\x)}{q_\phi(\z \mid \x)} \mathrm{d}\!\z\\
	&=  \int\! q_\phi(\z \mid \x) \log p_\theta(\x) \mathrm{d}\!\z - \int\! q_\phi(\z \mid \x) \log\frac{q_\phi(\z \mid \x)}{p_\theta(\z \mid \x)} \mathrm{d}\!\z\\
	&\overset{(\ast\ast)}{=} \log p_\theta(\x) - \kldiv{q_\phi(\z \mid \x)}{p_\theta(\z \mid \x)} \label{eq:ELBO1}\\
\end{align}

At $(\ast)$, Jensen's inequality is used to get a lower bound estimate. The expression in \eq{eq:Jensen} is called Evidence Lower Bound (ELBO). At $(\ast\ast)$, we identify the log-likelihood and the Kullback-Leibler divergence $D_\mathrm{KL}$ between the approximated and the true posterior. The KL-divergence is a measure of distance between two probability distributions, it is non-negative and equals zero if and only if the two distributions match almost everywhere. From \eq{eq:ELBO1} we see that the ELBO becomes exact for the right choice of $q_\phi(\z \mid \x)$.\\

Since we want to employ NNs as function approximators with the parameters $\{\phi, \theta\}$, we reformulate the maximum likelihood estimation as an optimization problem that can be solved with gradient methods. Rearranging the terms in \eq{eq:Jensen} differently yields
\vspace{4pt}
\begin{equation}\label{eq:ELBO2}
	\mathrm{ELBO} = \mathbb{E}_{q_\phi(\z \mid \x)}\big[\log p_\theta(\x \mid \z)\big] - \kldiv{q_\phi(\z \mid \x)}{p_\theta(\z)}
\end{equation}

Maximizing the ELBO thereby comprises two concurrent tasks: (1) Maximizing the reconstruction quality and (2) keeping the approximate posterior close to the latent prior. For the latter, the latent prior is typically assumed to be a standard Gaussian distribution $p(\z) = \mathcal{N}(\z; \bm 0, \bm I)$. For the former, the MSE between input and reconstruction is typically employed as reconstruction loss (assuming a Gaussian distribution again).\\

In order to be able to compute a low-variance estimate of the gradient $\nabla_{\phi, \theta}\mathrm{ELBO}$, we further need to reparameterize the random variable $\z$ as\footnote{While the concept of reparameterization is a general one, we specifically choose to estimate the mean and variance of a gaussian distribution in order to match the functional form of the latent prior.}

\begin{equation}
	q_\phi(\z \mid \x) = \mathcal{N}(\z; \mu_\phi(\x), \sigma_\phi(\x)^2 \bm I)
\end{equation}

That way, we can compute $\z$ given an input $\x$ as

\begin{equation}
	\z = \mu_\phi(\x) + \sigma_\phi(\x) \cdot \bm \epsilon \quad\quad \mathrm{with} \quad \bm\epsilon\sim\mathcal{N}(\bm \epsilon; \bm 0, \bm I)
\end{equation}

Separating the random nature of $\z$ from the deterministic encoder allows for propagating the gradient backwards through the whole model to jointly optimize $\phi$ and $\theta$.\\

VAEs are generative models that readily provide a meaningful latent space. The structure of the data representation is enforced by the second term in \eq{eq:ELBO2} which regularizes the information encoded by $q_\phi(\z \mid \x)$ while the (first) reconstruction term encourages unique encodings. It is this trade-off that gives the learnt representation a semantic structure. In terms of an autoencoder-pipeline, the regularization term restricts the size of bottleneck from which the input image is reconstructed. In order to explicitly control the width of the bottleneck, \cite{higgins_-vae_2017} propose to introduce a regularization weighting factor $\lambda$, such that the minimization objective becomes

\begin{equation}\label{eq:betaVAE}
	L_\mathrm{VAE} = - \mathbb{E}_{q_\phi(\z \mid \x)}\big[\log p_\theta(\x \mid \z)\big] + \lambda \cdot \kldiv{q_\phi(\z \mid \x)}{p_\theta(\z)}
\end{equation}

For $\lambda\rightarrow 0$ the VAE approaches a deterministic autoencoder, whereas a larger $\lambda$ narrows down the bottleneck and therefore increases the disentanglement of the representation, i.e., it forces the encoder to restrict itself to encoding only the main latent factors in $\z$.\\

While VAEs proved to be useful for representation learning in the image domain \cite{liu_learning_2022, locatello_challenging_2019}, they tend to generate blurry samples \cite{kingma_introduction_2019} and thus, in the task of image synthesis, other generative model types (e.g. GANs, Diffusion Models) often outperform VAEs.

\newpage
\section{Diffusion Models}\label{sec:diffusion models}
Diffusion Models (DMs) synthesize data by reversing a gradual noising process. A forward diffusion process successively adds small amounts of gaussian noise to the training data while the model learns to recover the destroyed information. Samples can then be generated by passing random noise through the learnt reverse process.\\

There are various design choices involved when setting up a DM – not all of which are addressed in the following sections. In particular, we only consider discrete diffusion processes. The derivations and notation roughly follow that of \cite{ho_denoising_2020}. \sec{sec:LDM} extends the classic DM by employing it in the latent space of an autoencoder model. Further, \sec{sec:LRDM} introduces a designated representation encoder.

\subsection{Diffusion Process}\label{sec:diffusion process}
\FigDDPM

A DM can be described as latent variable model where the latent variables form a Markov chain, i.e., each variable $\x_t$ only depends on the previous one $\x_{t-1}$ (see \fig{fig: ddpm}). Under the Markov assumption, we can write the approximate posterior as follows:

\begin{equation}\label{eq: forward process}
	q(\x_{1:T} \mid \x_0) \coloneqq \prod_{t=1}^T q(\x_t \mid \x_{t-1})
\end{equation}\\

The forward process transition distributions are assumed to be gaussian

\begin{equation}\label{eq: forward process transition}
	q(\x_t \mid \x_{t-1}) \coloneqq \mathcal{N}(\x_t; \sqrt{1-\beta_t}\x_{t-1}, \beta_t I)
\end{equation}

with a variance schedule $\{\beta_t\}_{t=1,...,T}$. In this work, the schedule is fixed to a linear noise schedule $\beta_t = \frac{T-t}{T-1}\beta_1+\frac{t-1}{T-1}\beta_T$, with $t=1,...,T$ and $T=1000$. However, other schedules such as the cosine-schedule can be chosen \cite{nichol_improved_2021}, or $\beta_t$ may also be learned by the model \cite{kingma_variational_2021}. The scaling factor $\sqrt{1-\beta_t}$ of the mean ensures that the variance is bounded, such that for large enough $T$, $\x_T$ follows a gaussian distribution with unit variance.\\

The pre-defined forward diffusion process $q(\x_{1:T} \mid \x_0)$ gradually transforms the data distribution $q(\x_0)$ to a standard gaussian distribution. To synthesize new images, we are interested in learning the reverse process

\begin{equation}\label{eq: reverse process}
	p_\theta(\x_{0:T}) \coloneqq p(\x_T) \prod_{t=1}^T p_\theta(\x_{t-1} \mid \x_t)
\end{equation}

The reverse process transition distributions are usually assumed to be gaussian

\begin{equation}\label{eq: reverse process transition}
	p_\theta(\x_{t-1} \mid \x_t) \coloneqq \mathcal{N}(\x_{t-1}; \bm \mu_\theta(\x_t, t), \bm \Sigma_\theta(\x_t, t))
\end{equation}

This is justified by the fact that the true reverse process transitions $q(\x_{t-1} \mid \x_t)$ are of the same functional form for small enough perturbations $\beta_t$ \cite{ho_denoising_2020, xiao_tackling_2021}. We will see in \sec{sec:training objective} that the training objective can then be reduced to KL-Divergences between gaussian distributions, which have a closed form. For that, let us first make some additional observations about the diffusion process that help making the training objective tractable.\\

With $\alpha_t \coloneqq 1-\beta_t$ and $\bar{\alpha}_t \coloneqq \prod_{s=1}^{t}\alpha_s$, it follows from \eq{eq: forward process transition}:

\begin{equation}\label{eq: forward process jump}
	q(\x_t \mid \x_0) = \mathcal{N}(\x_t; \sqrt{\bar{\alpha}_t}\x_0, (1-\bar{\alpha}_t) I)
\end{equation}

By reparameterizing the gaussian distribution we can directly sample $\x_t$ at any timestep $t$ via

\begin{equation}\label{eq: forward process epsilon}
	\x_t(\x_0, \bm \epsilon) = \sqrt{\bar{\alpha}_t}\x_0 + \sqrt{1-\bar{\alpha}_t}\bm\epsilon \quad\quad\mathrm{with}\enspace \bm\epsilon\sim\mathcal{N}(\bm\epsilon;\bm 0, \bm I)
\end{equation}

(The linear noise schedule $\beta_t$ and the relevant parameters for the diffusion process are visualized in \sec{sec:apx linear noise schedule}.) This means that the model can be trained by looking at each transition separately and not having to sample along the forward or reverse Markov chain during training.\\

Moreover, the forward process posteriors are tractable when conditioned on $\x_0$:

\begin{equation}\label{eq: forward process posterior}
	q(\x_{t-1}\mid \x_t, \x_0) = \mathcal{N}(\x_{t-1}; \tilde{\bm \mu}_t(\x_t, \x_0), \tilde{\beta}_t I)
\end{equation}

\begin{equation}\label{eq: mu beta tilde}
	\begin{split}
		\mathrm{where}\quad \tilde{\bm \mu}_t(\x_t, \x_0) & \coloneqq \frac{\sqrt{\bar{\alpha}_{t-1}}\beta_t}{1-\bar{\alpha}_t}\x_0 + \frac{\sqrt{\alpha_t}(1-\bar{\alpha}_{t-1})}{1-\bar{\alpha}_t}\x_t \\
		\mathrm{and}\quad \tilde{\beta}_t & \coloneqq \frac{1-\bar{\alpha}_{t-1}}{1-\bar{\alpha}_t}\beta_t
	\end{split}
\end{equation}

\subsection{Training Objective}\label{sec:training objective}
DMs are trained to minimize the negative log-likelihood by maximizing the ELBO (cf. \sec{sec:vae}).

\begin{equation}\label{eq: variational bound ansatz}
	\mathbb{E}[-\log p_\theta(\x_0)] \leq \mathbb{E}_q \bigg[-\log \frac{p_\theta(\x_{0:T})}{q(\x_{1:T}\mid \x_0)}\bigg] \eqqcolon L_{vlb}
\end{equation}

The upper bound $L$ can be rewritten in terms of tractable KL-Divergences \cite{ho_denoising_2020, sohl-dickstein_deep_2015} (see \sec{sec:apx dm objective} for a derivation):

\begin{equation}\label{eq: variational bound}
	\begin{split}
		L_{vlb} =  \mathbb{E}_q \bigg[ &
		\underbrace{\kldiv{(q(\x_T\mid \x_0)}{p(\x_T)}}_\text{$L_T$} \underbrace{- \log p_\theta(\x_0\mid \x_1)}_\text{$L_0$} \\
		& + \sum_{t>1} \underbrace{\kldiv{q(\x_{t-1}\mid \x_t, \x_0)}{p_\theta(\x_{t-1}\mid \x_t)}}_\text{$L_{t-1}$}\bigg]
	\end{split}
\end{equation}

With the forward process being fixed, $L_T$ is a constant w.r.t.\ the optimization of the model parameters $\theta$. Moreover, for large enough $T$, the true distribution $q(\x_T\mid \x_0)$ is very close to the assumed prior distribution $p(\x_T) = \mathcal{N}(\x_T; \bm 0, \bm I)$ and thus $L_T$ very close to zero. For discrete image data $\x_0$, discrete log-likelihoods are required. \cite{ho_denoising_2020} derive a discrete decoder from $\mathcal{N}(\x_0; \bm \mu_\theta(\x_1, 1), \sigma_1^2\bm I)$ in order to compute $L_0$. Minimizing the $L_{t-1}$ terms ensures that in each step the model recovers the corresponding lost information. Note that since we can sample $\x_t$ directly for any timestep $t$ we can sample $t$ uniformly during training.\\

From an alternative derivation (see \sec{sec:apx dm objective}), it follows that minimizing \eq{eq: variational bound} is equivalent to approximating the unconditional true reverse transitions $q(\x_{t-1}|\x_t)$. This is intuitive when considering that $p_\theta(\x_{t-1}\mid \x_t)$ does not have access to the information that is destroyed by the diffusion process up to the timestep $t$.

\subsection{Parameterizing the Reverse Process}\label{sec:parameterization reverse process}
This section discusses different possible parameterizations of the reverse process $p_\theta(\x_{t-1}\mid \x_t)$. First, we fix the reverse process variance to be the same as the forward process variance $\bm\Sigma_\theta(\x_t,t) = \sigma_t^2\bm I \coloneqq \beta_t\bm I$ \cite{ho_denoising_2020}. $\Sigma_\theta(\x_t,t)$ could also be trained, as explored in \cite{dhariwal_diffusion_2021}. The most straightforward parameterization is to let the model predict $\bm\mu_\theta(\x_t,t)$. The loss term $L_{t-1}$ in \eq{eq: variational bound} then becomes

\begin{equation}\label{eq: Lt-1 mu}
	L_{t-1} = \expectation{q}{\frac{1}{2\sigma_t^2} \norm{\tilde{\bm\mu}_t(\x_t,\x_0) - \bm\mu_\theta(\x_t,t)}^2} + C 
\end{equation}

with $C$ being a constant w.r.t.\ $\theta$. Another parameterization option is to let the model directly predict the denoised image ${\x_0}_\theta(\x_t,t)$. Rewriting \eq{eq: Lt-1 mu} with \eq{eq: mu beta tilde}, we get the following expression:

\begin{equation}\label{eq: Lt-1 x0}
	L_{t-1}-C = \expectation{q}{ \frac{\bar{\alpha}_t\beta_t^2}{2\sigma_t^2\alpha_t(1-\bar{\alpha}_t)^2}\norm{\x_0 - {\x_0}_\theta(\x_t, t)}^2}
\end{equation}

Further, using the reparameterization in \eq{eq: forward process epsilon}

\begin{equation}\label{eq: epsilon parameterization}
\begin{split}
	\tilde{\bm\mu}_t(\x_t,\x_0) &= \tilde{\bm\mu}_t(\x_t, \frac{1}{\sqrt{\bar{\alpha}_t}}(\x_t-\sqrt{1-\bar{\alpha}_t}\bm\epsilon)) \\
	&= \frac{1}{\sqrt{\alpha_t}} \bigg(\x_t - \frac{\beta_t}{\sqrt{1-\bar{\alpha}_t}} \bm\epsilon\bigg)
\end{split}
\end{equation}

we can also rewrite $L_{t-1}$ in terms of a noise predictor $\bm\epsilon_\theta(\x_t,t)$:

\begin{equation}\label{eq: Lt-1 eps}
	L_{t-1}-C = \expectation{\x_0,\bm\epsilon}{\frac{\beta_t^2}{2\sigma_t^2\alpha_t(1-\bar{\alpha}_t)}  \norm{\bm\epsilon - \bm\epsilon_\theta(\sqrt{\bar{\alpha}_t}\x_0 + \sqrt{1-\bar{\alpha}_t}\bm\epsilon, t)}^2}
\end{equation}

In this work, we compare the \textit{image}-parameterization ${\x_0}_\theta(\x_t,t)$ and the \textit{noise}-parameterization $\bm\epsilon_\theta(\x_t,t)$ (see \sec{sec:LDM reparameterization}) and their roles for representation learning (see \sec{sec:LRDM}). Predicting the mean of the reverse transitions lead to unstable training and much worse results in all learning configurations.\\

The estimator function is realized through a neural network. A convolution-based U-Net architecture \cite{ronneberger_u-net_2015} is very suitable for this task, since the predicted quantity has the same dimensionality as the input. Moreover, there is no need for a bottleneck (cf. \sec{sec:vae}) and the spatial inductive bias of the U-Net can be exploited. The timestep information is integrated via a sinusoidal positional embedding, as used in transformer models \cite{vaswani_attention_2017}.\\

\cite{ho_denoising_2020} find that the sample quality improves by a large margin when dropping the prefactors in \eq{eq: Lt-1 eps}. They reweight the loss terms in $L_{vlb}$ and propose an alternative minimization objective

\begin{equation}\label{eq: simple objective eps}
	L_\mathrm{noise}^\ast \coloneqq \expectation{t,\x_0,\bm\epsilon}{\norm{\bm\epsilon - \bm\epsilon_\theta(\sqrt{\bar{\alpha}_t}\x_0 + \sqrt{1-\bar{\alpha}_t}\bm\epsilon, t)}^2}
\end{equation}

(The subscript denotes the predicted quantity and the superscript asterisk denotes the dropped prefactors compared to the variational lower bound formulation.) We want to consider yet another reweighted objective for the image-parameterization which we obtain by dropping the prefactors in \eq{eq: Lt-1 x0}:

\begin{equation}\label{eq: simple objective x0}
	L_\mathrm{image}^\ast \coloneqq \expectation{t,\x_0,\bm\epsilon}{\norm{\x_0 - {\x_0}_\theta(\sqrt{\bar{\alpha}_t}\x_0 + \sqrt{1-\bar{\alpha}_t}\bm\epsilon, t)}^2}
\end{equation}

Note that minimizing $L_\mathrm{noise}^\ast$ or $L_\mathrm{image}^\ast$ no longer maximizes the ELBO. Instead, the objectives emphasize different parts of the reverse process. The reweighting of the loss terms is shown in \fig{fig: loss-weighting}. $L_\mathrm{noise}^\ast$ emphasizes intermediate timesteps whereas $L_\mathrm{image}^\ast$ puts a strong weight on higher $t$.

\FigLossWeightings

After training we can generate samples by sequentially evaluating the learnt reverse process $p_\theta(\x_{t-1}\mid \x_t)$, starting from $\x_T\sim\mathcal{N}(\x_T; \bm 0, \bm I)$. In order to sample from the gaussian transition probability distributions, in each step we sample a noise vector $\bm z_T\sim\mathcal{N}(\bm z; \bm 0, \bm I)$. The sampling procedure can then be written for the noise-parameterization as

\begin{equation}\label{eq: sampling eps}
	\x_{t-1} = \frac{1}{\sqrt{\alpha_t}}\bigg(\x_t - \frac{\beta_t}{\sqrt{1-\bar{\alpha}_t}}\bm\epsilon_\theta(\x_t,t)\bigg) + \sigma_t \bm z
\end{equation}

and for the image-parameterization as

\begin{equation}\label{eq: sampling x0}
	\x_{t-1} = \frac{\sqrt{\bar{\alpha}_t}\beta_t}{\sqrt{\alpha_t}(1-\bar{\alpha}_t)} {\x_0}_\theta(\x_t,t) + \frac{\sqrt{\alpha_t}(1-\bar{\alpha}_{t-1})}{1-\bar{\alpha}_t} \x_t + \sigma_t \bm z
\end{equation}

Alternative to the probabilistic sampling, \cite{song_denoising_2020} derive a sampling scheme (further denoted as DDIM-sampling) that deterministically maps $\x_T$ onto $\x_0$. While it technically corresponds to a different, non-Markovian forward process, it can be readily employed for any pretrained diffusion model.

\begin{equation}\label{eq: sampling x0 ddim}
	\x_{t-1} = \sqrt{\bar{\alpha}_{t-1}}\bigg( \frac{\x_t - \sqrt{1-\bar{\alpha}_t}\bm\epsilon_\theta(\x_t,t)}{\sqrt{\bar{\alpha}_t}}\bigg)
	+ \sqrt{1-\bar{\alpha}_{t-1}}\bm\epsilon_\theta(\x_t,t)
\end{equation}

An analogous scheme (obtained by making use of \eq{eq: forward process epsilon}) can be used for image-parameterized models.

\subsection{Latent Diffusion Models}\label{sec:LDM}
DMs have shown impressive results in terms of sample quality as well as mode coverage, i.e., the samples exhibit a great diversity \cite{ho_denoising_2020, dhariwal_diffusion_2021}. However, both training and sampling with DMs is computationally very demanding. They typically have $\sim\!1000$ latent variables that are of the same dimensionality as the input and sampling requires sequential evaluation of the Markov chain.\footnote{Sampling methods such as DDIM-sampling allow for evaluating fewer steps than the model was trained on. While this reduces sampling time, it usually comes at the cost of lower sample quality.}\\

In order to leverage the expressiveness of DMs while reducing the model size, they can be trained in the latent space of an autoencoder \cite{rombach_high-resolution_2021, vahdat_score-based_2021}. The main idea of the autoencoder is to reduce the dimensionality of the diffusion model while retaining as much semantic information as possible. This was motivated by the observation that DMs (and likelihood-based models in general) use much capacity on modeling imperceptible details of the data \cite{dieleman2020typicality, rombach_high-resolution_2021, ho_denoising_2020}. While reweighting the diffusion loss term (see \fig{fig: loss-weighting}) addresses this to some extent, employing the DM on the compressed data drastically reduces its computational demands and explicitly masks out low-level image details for the likelihood-based DM optimization.\\

The diffusion model learns the distribution of continuous latents $\z\sim q(\z)$ which are obtained by encoding the images $\z = \mathcal{E}(\x)$. The decoder recovers the image from the latent $\tilde{\x} = \mathcal{D}(\z)=\mathcal{D}(\mathcal{E}(\x))$. We experiment with the autoencoders $\{\mathcal{E},\mathcal{D}\}$ provided by \cite{rombach_high-resolution_2021}, which are trained with a perceptual and an adversarial loss. The autoencoder is then fixed for the DM training. In the following, the DM trained in the latent space, as well as the whole model setup, will be referred to as LDM (Latent Diffusion Model).

\FigOverviewLatentDiffusion

The training objective for the LDM is the same as that for the DM, i.e., all derivations in the previous sections\footnote{There is no need for the discrete decoder in the ELBO derivation since the latents are continuous. The $L_0$ term thus becomes an additional $L_{t-1}$-summand.} still hold with $\x$ replaced by $\z$. During training, $\z_t$ is obtained by a single pass of $\x$ through the encoder $\mathcal{E}$ and then applying \eq{eq: forward process epsilon}. For image synthesis, $\z$ is sampled from the latent DM and subsequently passed through the decoder $\mathcal{D}$.\\

The core of the LDM is a time-conditional U-Net \cite{dhariwal_diffusion_2021} consisting of convolutional residual blocks and attention layers at certain depth-levels. Early experiments show that choosing an autoencoder with spatial compression by a factor of $f=4$ and unchanged channel number yields the best results in terms of sampling and reconstruction quality. This can be explained by the strong spatial inductive bias of both the autoencoder and the LDM architecture which can then be leveraged the most. An input image of size $H\times W \times c$ is thus encoded to a latent of size $H/f\times W/f \times c$.

\subsection{Representation-conditional Latent Diffusion}\label{sec:LRDM}
A striking difference between DMs and VAEs is that in the former the latent variables are designed s.t.\ information is sequentially destroyed, whereas in the latter the latent variable contains a dense representation of the data, allowing for faithful image reconstruction and interpolation (depending on the regularization towards the latent prior). In order to learn a meaningful data representation in the framework of diffusion models, we extend the LDM by a jointly trained representation encoder $\mathcal{E}_r \colon \z_0 \mapsto \rep(\z_0)$. The extracted code $\rep(\z_0)$ is then passed to the denoising U-Net as conditioning information.

\FigOverviewRepresentationLearning

\fig{fig: overview representation learning} shows the graphical model representation for the LRDM (Latent Representation Diffusion Model). The joint distribution becomes

\begin{equation}\label{eq: repr joint probability}
	p_\theta(\z_{0:T}, \rep) = p_\theta(\x_{0:T}\mid\rep)p(\rep) = p(\rep)p(\z_T) \prod_{t=1}^T p_\theta(\z_{t-1} \mid \z_t, \rep)
\end{equation}

and the extended forward process is now
\begin{equation}
	q_\phi(\z_{1:T},\rep\mid\z_0) = q(\z_{1:T}\mid\z_0)q_\phi(\rep\mid\z_0)
\end{equation}

with $\phi$ being the parameters of the representation encoder $\mathcal{E}_r$. We set the latent prior to a gaussian distribution $p(\rep)=\mathcal{N}(\rep;\bm 0, \bm I)$ and parameterize $\mathcal{E}_r$ accordingly through $q_\phi(\rep\mid\z_0) = \mathcal{N}(\rep;\bm\mu_\phi(\z_0),\bm\sigma_\phi(\z_0)^2\bm I)$. As training objective, we propose the (representation-conditional) LDM loss combined with a regularization term controlled by a parameter $\lambda$.

\begin{equation}\label{eq: LRDM loss}
\begin{split}
	L_{LRDM} & \coloneqq L_\mathrm{image}^\ast + \lambda L_\mathrm{prior} \\
	& = \expectation{\mathcal{E}(\x_0),\bm\epsilon,t}{\norm{\z_0 - {\z_0}_\theta(\z_t, t, \rep(\z_0))}^2} + \lambda\kldiv{q_\phi(\rep\mid\z_0)}{p(\rep)}
\end{split}
\end{equation}

The first term is the reweighted diffusion loss (for the image-parameterization) and the second term pushes the representation distribution towards the gaussian prior. $L_{LRDM}$ corresponds to a reweighted ELBO-derived objective (see \sec{sec:apx lrdm objective} for a derivation).\\

Note that maximizing the ELBO is no longer equivalent to approximating the unconditional true reverse transitions $q(\z_{t-1}\mid\z_t)$ because $p_\theta(\z_{t-1}\mid\z_t,\rep(\z_0))$ can now receive information through $\rep(\z_0)$ that is destroyed in $\z_t$. That, in turn, also provides the incentive for learning a meaningful representation. The information encoded by $\mathcal{E}_r$ can bridge the information gap between $\z_t$ and $\z_0$, which is to be predicted.\\

In order to increase the expressivity of the representation, on can consider a separate latent variable $\rep_t$ for each timestep (see \fig{fig: overview t-cond representation learning}). This can be realized by making the representation encoder $\mathcal{E}_r$ timestep-conditional. We call this model \mbox{t-LRDM} in the following. The corresponding loss is then

\begin{equation}\label{eq: t-LRDM loss}
	L_{t-LRDM} \coloneqq \expectation{\mathcal{E}(\x_0),\bm\epsilon,t}{\norm{\z_0 - {\z_0}_\theta(\z_t, t, \rep_t(\z_0,t))}^2 + \lambda\kldiv{q_\phi(\rep_t\mid\z_0)}{p(\rep_t)}}
\end{equation}

\FigOverviewTimeConditionalRepresentationLearning

\newpage
\chapter{Related Work}\label{ch:related work}
With Diffusion Models having received a lot of attention in recent research, several concurrent projects were published over the course of this thesis. In particular, these include (in chronological order) \cite{vahdat_score-based_2021, abstreiter_diffusion-based_2021, preechakul_diffusion_2021, rombach_high-resolution_2021, pandey_diffusevae_2022, ramesh_hierarchical_2022}. Those and other related work are addressed in this chapter.

\subsection*{Variational Autoencoders}
Our work brings VAEs \cite{kingma_auto-encoding_2014, kingma_introduction_2019} and diffusion models \cite{sohl-dickstein_deep_2015, ho_denoising_2020} closely together. VAEs are suitable for representation learning as the VAE objective enforces a meaningful structured latent space \cite{liu_learning_2022}. However, they can exhibit "posterior collapse" \cite{oord_neural_2018} which DMs do not \cite{preechakul_diffusion_2021}. Our approach to representation learning with diffusion models (LRDM) combines an adapted VAE objective with the DM objective. Typically, VAEs assume gaussian posteriors and gaussian priors over continuous latents. The VQ-VAE \cite{oord_neural_2018} uses discrete latents with vector-quantization as regularization. Hierarchical approaches can strongly increase the sampling quality \cite{vahdat_nvae_2021}.

\subsection*{Diffusion Models}
Diffusion generative models \cite{sohl-dickstein_deep_2015, ho_denoising_2020} have shown impressive results in multiple domains (especially on images \cite{ho_denoising_2020, dhariwal_diffusion_2021} and audio \cite{mittal_symbolic_2021}) on both generative tasks \cite{dhariwal_diffusion_2021} and density estimation \cite{kingma_variational_2021}. Competitive sampling quality could be reached by introducing the reweighted training objective in \cite{ho_denoising_2020}, which also establishes a close connection to score-based models \cite{song_generative_nodate, song_score-based_2020, vincent_connection_nodate}. For the latter, a continuous formulation based on SDEs is presented in \cite{song_score-based_2020}. DMs exhibit stable training and allow for likelihood evaluation, as opposed to GANs \cite{goodfellow_generative_2014}. However, the computational requirements for DMs are high, and sampling is slow compared to GANs due to the sequential evaluation of the reverse process. Thus, multiple works proposed imrovements to the sampling process \cite{song_denoising_2020, jolicoeur-martineau_adversarial_2020, song_score-based_2020}. Important architecture improvements were presented in \cite{dhariwal_diffusion_2021}, further pushing the limits of image synthesis.
\enlargethispage{\baselineskip}
Other variants employ different noise schedules \cite{nichol_improved_2021} or a reweighted learning objective \cite{choi_perception_nodate} to improve sampling quality.

\subsection*{Latent Diffusion Models}
Training and sampling with DMs requires a lot of computational resources, because all latent variables are of the same dimensionality as the input and hence model evaluations and gradient computations are costly for high-resolution images. By employing them in a compressed latent space of an autoencoder, they can be trained more efficiently, also on high-resolution images \cite{rombach_high-resolution_2021}. In \cite{vahdat_score-based_2021}, a diffusion model is employed in the latent space of a jointly trained NVAE. In \cite{rombach_high-resolution_2021}, the difficulty of weighing reconstruction quality against learning the prior over the latent space is avoided by separating the training of the first-stage autoencoder and the latent DM. As first-stage model, a VQGAN \cite{esser_taming_2020} is pretrained.

\subsection*{Representation Learning with Diffusion Models}
In DMs, information contained in the latent variables is successively destroyed by the diffusion process. The semantic information is then rather contained in the learnt reverse process as a whole, distributed over the different timesteps. This makes representation learning difficult for DMs, as opposed to other generative models, such as VAEs, which inherently provide a low-dimensional latent representation space.\\

\cite{abstreiter_diffusion-based_2021} propose to condition the denoising U-Net on the output of a jointly trained representation encoder which has access to the denoised input image. Their work differs from ours in various respects: i) They employ a DM directly in image space, whereas the LRDM uses the representation-conditional DM to model the distribution of compressed latents. This allows us provide results for images of higher resolution. ii) In \cite{abstreiter_diffusion-based_2021}, the noise-parameterization is used. We find that the image-parameterization is favorable in the LRDM setup and show that it provides a closer connection to the unconditional VAE for representation learning. iii) They use a non-spatial representations that are passed to the denoising U-Net together with the positional timestep-conditioning. We use spatial representations and concatenation, respectively.
\cite{abstreiter_diffusion-based_2021} also introduces the timestep-conditional representation. \cite{mittal_points_2022} find that it contains information of different kind at different timesteps, i.e., training a timestep-conditional encoder provides a richer representation.\\

\newpage
\textit{Diffusion autoencoders} \cite{preechakul_diffusion_2021} use a deterministic semantic encoder on which a jointly trained DDIM is conditioned. They are able to learn a meaningful and decodable non-spatial representation. However, in order to sample from the representation space, a separate DM has to be trained on the encoded latents. Our work focuses on regularizing the representation through a KL-penalty. For a strong enough regularization, we can directly sample from the latent gaussian prior without training an additional model. For \textit{DiffuseVAE}s \cite{pandey_diffusevae_2022}, a VAE and DM is trained separately. The DM thereby refines the VAE samples. \textit{DALL-E 2} \cite{ramesh_hierarchical_2022} uses CLIP \cite{radford_learning_2021} embeddings on which a diffusion model is conditioned.

\newpage
\chapter{Experiments}\label{ch:experiments}
In this chapter, the main results of this thesis are presented. In \sec{sec:experiment latent diffusion}, the LDM and the role of the reverse process parameterization are analyzed. \sec{sec:experiment repr learning} shows results for the representation learning setup (LRDM), \sec{sec:from LVAE to LDM} further investigates the role of the regularization term, and \sec{sec:experiment t-cond repr} evaluates the t-LRDM. Finally, \sec{sec:conditional repr learning} presents results for experiments with additional conditioning information.\\

The implementation of the framework used in this thesis is based on \cite{dhariwal_diffusion_2021} and \cite{song_score-based_2020}, the architecture of the LDM is based on \cite{dhariwal_diffusion_2021} (with the pretrained autoencoder from \cite{rombach_high-resolution_2021}). The code is available at \href{https://github.com/jeremiastraub/diffusion}{GitHub}.

\section{Latent Diffusion Model}\label{sec:experiment latent diffusion}
We begin by giving additional details on the model architecture and training, before discussing quantitative results in \sec{sec:LDM reparameterization}.\\

Because of the mild spatial compression rate ($f$=4) of the first-stage model, the latents $\z$ are still image-like, thereby allowing us to exploit the spatial bias of the well-developed denoising U-Net architecture. Experiments confirm that an image-like dimensionality of the latents (with 3 channels) yields the best sampling quality for the used architecture.\\

We compare the two autoencoders provided by \cite{rombach_high-resolution_2021} in terms of sampling quality. Both are trained with a perceptual and an adversarial loss for high reconstruction quality, but one is (slightly) regularized by a KL-divergence towards a standard normal (referred to as KL-AE in the following) and the other is regularized by a vector quantization layer \cite{oord_neural_2018} within the decoder (referred to as VQ-AE in the following). The latter learns a discrete codebook of fixed size to represent the images as spatial collection of codebook entries \cite{esser_taming_2020}. In this case, the DM is trained on the pre-quantizations, i.e., on the continuous latents just before the quantization layer. For the KL-AE, the latents are sampled from the reparameterized posterior $\mathcal{E}(\x)=\mathcal{E}_\mu(\x)+\mathcal{E}_\sigma(\x)\epsilon$, with $\epsilon\sim\mathcal{N}(0,1)$.\\

For both autoencoders, we rescale the latents $\z$ to have unit variance. This ensures that the gaussian assumption of the latent prior $p(\z_T)$ is valid by decreasing the signal-to-noise ratio. The rescaling has significant effects for the KL-AE, while it has no noticable effect for the VQ-AE \cite{rombach_high-resolution_2021}. In order to rescale the latents $\z \leftarrow \z/\hat{\sigma}$, the standard deviation $\sigma$ is estimated from the first 100 batches via Welford's online algorithm.\\

In the denoising U-Net, Adaptive Group Normalization \cite{dhariwal_diffusion_2021} is used to inject the timestep-embedding into the residual blocks. Upsampling and downsampling is done by the residual blocks. Similar to \cite{vahdat_score-based_2021}, we find that a dropout of around $0.2$ for the LDM yields better results than without dropout. An overview of all hyperparameters for the different experiments can be found in \sec{sec:apx hyperparameters}.

\subsection{Reparameterizing the Reverse Process}\label{sec:LDM reparameterization}
This section quantitatively evaluates the LDM, in particular comparing $L_\mathrm{noise}^\ast$ and $L_\mathrm{image}^\ast$. We investigate differences between the parameterizations and demonstrate that the image-parameterized LDM can achieve competitive FID \cite{heusel_gans_2018} scores.\\

As described in \sec{sec:parameterization reverse process}, the reverse process can be parameterized in different ways. Most frequently, the noise-parameterization is chosen – driven by the success of the $L_\mathrm{noise}^\ast$ objective \cite{ho_denoising_2020}, where the $\epsilon$-MSE for each noise scale is weighted equally. Relative to the ELBO-derived objective, the loss terms for low $t$ are given very little weight and intermediate noise scales are weighted the most. Previous experiments have shown that the bits allocated at low $t$ correspond to imperceptible distortions \cite{ho_denoising_2020, rombach_high-resolution_2021, choi_perception_nodate}. Reweighting the loss terms allows for introducing an inductive bias towards high perceptual sample quality. However, it is not clear which reweighting factors maximize the perceptual quality.
In \cite{choi_perception_nodate}, this question is approached by looking at the signal-to-noise ratio which reduces during the diffusion process, depending on the noise schedule. They identify a so-called \textit{content}-range of intermediate SNR-values ($10^{-2}-10^0$) and show that putting a strong weight on the corresponding noise scales consistently improves the sample quality in terms of FID \cite{heusel_gans_2018}.
We want to note that the empirically estimated \textit{content}-range is not universal and might not be the same when training DMs not on the image data directly but on a latent representation. Moreover, our results indicate that the optimal weighting scheme also depends on the underlying learning task. In particular, we obtain the best sample quality for the image-parameterization with a weighting scheme that strongly differs from $L_\mathrm{noise}^\ast$ and the one in \cite{choi_perception_nodate}.\\

\noindent\textbf{Structural differences between the image and noise parameterization}\\

There is little research on the different reparameterizations. \cite{ho_denoising_2020} show results for the $\bm\mu_\theta(\x_t,t)$-parameterization (with $L_{vlb}$) and report that training is instable when dropping the loss-prefactors. We observe instable training with this parameterization for both weighting schemes. Furthermore, \cite{ho_denoising_2020} discard the image-parameterization mentioning worse sampling quality. Other than that, the image-parameterization is used in \cite{xiao_tackling_2021}, although they formulate a very different, adversarial learning objective.\\

In the following, we want to discuss fundamental differences in the learning task for the different parameterizations. First, we compare the pixel-wise distribution of the actual NN output. For $\bm\mu_\theta(\x_t,t)$, the NN directly performs the small denoising step, which means that its output is close to a standard gaussian for $t\rightarrow T$ and ideally matches the input image for $t\rightarrow0$. $\bm\epsilon_\theta(\x_t,t)$ and ${\z_0}_\theta(\x_t,t)$, on the other hand, have a fixed target during a sampling process – with the former always predicting gaussian noise and the latter always predicting the input image.\footnote{For another perspective, let us briefly consider the whole generative process as a single neural network with $\x_{1:T-1}$ being hidden layers and shared parameters across layers. Then, $\bm\epsilon_\theta(\x_t,t)$ and ${\z_0}_\theta(\x_t,t)$ can be viewed as ResNet-like NNs (compare \eq{eq: sampling eps} and \eq{eq: sampling x0}), as opposed to $\bm\mu_\theta(\x_t,t)$. This might give an intuition for why training the latter is more difficult. Note that this comparison somewhat falls short since the model is not really trained "end-to-end".} For the noise-parameterization, \ref{eq: forward process epsilon} shows that the learnt mapping for $t\rightarrow T$ is close to the identity transform. For $t\rightarrow0$, this is less and less so, but also the information gap between $\z_t$ and $\z_0$ decreases. For the image-parameterization, this is the other way round in the sense that the mapping approaches the identity transform for $t\rightarrow0$, and the prediction task for $t\rightarrow T$ is particularly difficult.\\

\noindent\textbf{Image-parameterized LDMs achieve competitive sampling quality}\\

Can the same sample quality be achieved with the image-parameterization as with the widely used noise-parameterization? \tab{tab: FID z0 vs eps} shows results on the sampling quality for various models trained on the LSUN-Churches \cite{yu_lsun_2016} ($256\times256$) dataset. The reported FID and IS values were evaluated with $50k$ samples (using all 1000 sampling steps) against the training set, using the Python package provided by \cite{obukhov2020torchfidelity}. The LDM achieves competitive results after a relatively short training time for both $L_\mathrm{noise}^\ast$ and $L_\mathrm{image}^\ast$. The FID scores for the image-parameterization are consistently slightly higher than for the noise-parameterization but they don't fall behind by far. Unconditional samples are displayed in \fig{fig: samples latent diffusion eps} and \fig{fig: samples latent diffusion z0}.\\

\noindent\textbf{Efficient training in latent space}\\

LDMs can be trained efficiently on high-dimensional image data. After only 100 epochs, which corresponds to $47k$ gradient updates with a batch-size of $256$, an FID better than in \cite{ho_denoising_2020} is reached. In terms of V100 training days, this corresponds to 4 vs 50 in \cite{ho_denoising_2020}.\footnote{Our models were trained on a single NVIDIA A100 GPU and we assume a speedup of $\times2.2$ for A100 vs V100 \cite{rombach_high-resolution_2021}. \cite{ho_denoising_2020} assume a speedup of $\times8$ for their TPU v3-8 vs V100.} For comparison, LDM-8$^\ast$ was trained for around 400 epochs. Note that the FID scores were evaluated after a fixed number of training epochs (instead of reporting the lowest overall FID, as done frequently). The training time was constricted to avoid excessive usage of computational resources while being able to perform ablation studies. However, the FID scores consistently decreased over training time, and we expect further improvement for longer training (The image-parameterized LDM (VQ-AE) reaches an FID of 5.64 and an IS of 2.62 after 400 training epochs). Evaluating different dropout probabilities for the latent DM shows that using dropout is beneficial when training in latent space. We fix the dropout to 0.2 for further LDM experiments.\\

\newpage
\noindent\textbf{Distortion analysis: Structural differences between DMs and LDMs}\\

Next, we take a closer look at what the different estimators learn. To that end, we compute the distortion (RMSE) curves both in the image space ($\x_0$-distortion) and in the latent space of the first-stage model ($\z_0$-distortion), i.e., $\sqrt{\langle\norm{\x_0-\hat{\x}_0}^2\rangle}$ and $\sqrt{\langle\norm{\z_0-\hat{\z}_0}^2\rangle}$, respectively. For the image-parameteri\-za\-tion, $\hat{\z}_0$ is directly given by the NN output. For the noise-parameterization, it can be obtained from the predicted noise via \eq{eq: forward process epsilon}. The distortion curves for the different LDMs are displayed in \fig{fig: eps vs x0 RMSE}. When comparing the the parameterizations, we see that for a large timestep range, the $\z_0$-distortions are very similar (see \fig{fig: distortion b}). Visualizing the ratio (see \fig{fig: distortion c} and \fig{fig: distortion c}) shows that the $\z_0$-predictions are more accurate for the image-parameterized model at high $t$ but significantly less accurate at very low $t$. We attribute the former to the strong weighting that $L_\mathrm{image}^\ast$ puts on large $t$ and the latter to the difficulty of predicting crisp images. However, we also see that this accuracy gap at low $t$ vanishes in the image space, i.e., after passing the latent through the decoder of the first-stage model. Hence, the first-stage model (VQ-AE and KL-AE) can compensate for the extra noise in $\hat{\z}_0$.\\

\noindent\textbf{Image-parameterized LDMs harmonize well with VQ-AEs}\\

Comparing the two different first-stage models, we observe, as in \cite{rombach_high-resolution_2021}, that the sample quality is better for the VQ-AE than for the KL-AE (see \tab{tab: FID z0 vs eps}) even though the reconstruction quality (without latent DM) is slightly better for the latter \cite{rombach_high-resolution_2021}. While that difference is small for the noise-parameterized model, it is significant for the image-parameterized model, improving the FID by more than 4 points (after 200 epochs). In \fig{fig: distortion b}, we observe a striking difference between the $\z_0$-distortion curves for the VQ-AE and the KL-AE, indicating structural differences between the latent spaces of the two first-stage models. Recall that in the decoder of the VQ-AE, the latents are first quantized via a discrete codebook, i.e., each latent vector (one for each pixel in the latent image representation) is mapped to its closest codebook entry. This might induce a certain invariance to $\z_0$-distortions that harmonizes well with the image-parameterization, as the $\x_0$-distortion is again similar for the VQ-AE and KL-AE, and even slightly lower for the VQ-AE at low $t$ (see \fig{fig: distortion a}). Overall, the distortion curves differ both between image and latent space and between the VQ-AE and KL-AE as first-stage model. This indicates that DM training configurations optimized for DMs that are employed in image space (e.g. noise schedule, loss-reweighting) need not be the most favorable for LDMs. For example, using the loss-weighting corresponding to $L_\mathrm{noise}^\ast$ together with the image-parameterized model leads to a significant decrease in sample quality (around 6 FID points after 100 epochs).\\

\TabZvsEps
\FigRMSEoverTnoRepr
\FigSamplesLatentDiffusionEps
\FigSamplesLatentDiffusionZ

\newpage
\section{Representation Learning}\label{sec:experiment repr learning}
We introduce LRDMs by extending LDMs by an additional encoding path which is regularized by a KL-penalty towards a gaussian prior, similar to a VAE. Experiments show that the LRDM is able to learn a meaningful structured representation that allows for faithful reconstructions and semantic interpolations. Before presentation quantitative results in \sec{sec:from LVAE to LDM} and further extensions in \sec{sec:experiment t-cond repr} and \sec{sec:conditional repr learning}, we give an overview of architectural choices and investigate the properties of the representation learning qualitatively.

\FigArchitectureRepresentationLearning

\fig{fig: architecture representation learning} depicts the architecture of the LRDM. The representation encoder $\mathcal{E}_r$ shares the same base architecture as the contracting path of the diffusion U-Net. The encoded representation $\rep$ is then injected into the contracting path of the diffusion U-Net by concatenation. For better guidance, a conditioning signal is provided on all depth-levels. For that, an intermediate module $\mathcal{D}_r$ with the inverse encoder base architecture, computes a conditioning signal for all spatial resolutions of the diffusion U-Net. The concatenation is done only before the first residual block of each depth-level. We use representations with spatial dimensions to introduce some spatial bias. In particular, we compare representations of shape $64\times8\times8$ ($C\times W\times W$) and $16\times16\times16$.\\

\noindent\textbf{A connection between VAEs and LRDMs}\\

The LRDM setup can be interpreted in two different ways:
i) As conditional LDM: The diffusion U-Net is now conditioned on the representation signal $\rep(\z_0)$ which is directly informed by the denoised image.
ii) As stack of conditional VAEs: With $\mathcal{E}_r$ as encoder and ${\z_0}_\theta$ as decoder (absorbing $\mathcal{D}_r$ in the decoder), we have a stack of $T$ VAEs with shared parameters that are conditioned on different signals $\z_t$. This becomes clear from \eq{eq: LRDM loss}, which matches a conditional VAE objective with a MSE reconstruction loss. At different timesteps, the conditioning $\z_t$ contains a different amount of information about $\z_0$. For $t=T$ (and large enough $T$), $\z_T$ is independent of $\z_0$ and thus the objective becomes that of an unconditional VAE. Therefore, the learning of a comprehensive representation is particularly enforced at $t\rightarrow T$.\\

In \cite{abstreiter_diffusion-based_2021}, the same argument is given, but for the noise-parameterized DM.
However, we argue that the last statement only holds for image-parameterized diffusion models. For the noise-parameterized model at $t=T$, the conditioning $\z_t$ is independent of $\z_0$, but it also matches the prediction target, since $\z_T\approx\bm\epsilon$. Thus, the unconditional VAE objective is not truly reached.
We hypothesize that this may affect representation learning in the way that capturing the global semantics (for high $t$) is not so much enforced, compared to the image-parameterization. In early experiments, we observe better results in terms of image reconstruction and interpolation for the latter. In the following, only image-parameterized LRDMs are used.\\

\newpage
\noindent\textbf{Faithful reconstructions and semantic interpolations}\\

Next, we want to qualitatively assess the properties of representation learning with LRDMs. We train LRDMs with a representation of size $64\times8\times8$ on LSUN-Churches \cite{yu_lsun_2016} and CelebA-HQ \cite{karras_progressive_2018} data. We then reconstruct input images from the encoded representation. We choose to take the mode of the approximate posterior $\rep = \bm\mu_\phi(\z_0)$ and use the deterministic DDIM sampling scheme.\footnote{In \fig{fig: reconstructions LSUNchurch deep} and \fig{fig: reconstructions CelebA-HQ}, we show reconstructions where $\rep$ is re-sampled from the approximate posterior and the probabilistic sampling scheme is used.} \fig{fig: reconstructions LSUNchurch deep DDIM} and \fig{fig: reconstructions CelebA-HQ DDIM} show reconstructions for re-sampled $\z_T$. The representation $\rep$ captures high-level semantics and global features, changing the latent code $\z_T$ affects local details like windows, mouth, eyes, or local texture and color.\\

The aim of regularizing the learnt $q_\phi(\rep\mid\z_0)$ towards a gaussian prior is to obtain a semantically structured representation space. We evaluate this property qualitatively by interpolating in representation space. For that we encode two input images into $\rep$ and $\z_T$ via $\rep = \bm\mu_\phi(\z_0)$ and \eq{eq: forward process epsilon}, respectively. Both are then interpolated via Slerp (Spherical linear interpolation) \cite{song_denoising_2020}. Finally, we decode the interpolated $(\rep, \z_T)$ pairs using DDIM sampling.\footnote{See \fig{fig: interpolations LSUNchurch deep} and \fig{fig: interpolations CelebA-HQ} for interpolations with probabilistic sampling and random $\z_T$ for each sample.} Interpolation results are shown in \fig{fig: interpolations LSUNchurch deep repr vs VQGAN DDIM} and \fig{fig: interpolations CelebA-HQ DDIM}. The LRDM produces smooth interpolations, gradually transforming semantic features from start to end point. In \fig{fig: interpolations LSUNchurch deep repr vs VQGAN DDIM}, a comparison to interpolations directly in the latent space of the first-stage model is shown. There, the latents are not semantically structured and the interpolations are merely a blending of the two input images. Note that the first-stage model is explicitly designed to not perform semantic compression, we include this comparison rather as additional motivation for why a learnt semantic representation is desirable. We further compare to DDIM interpolations with a LDM (VQ-AE) trained for 400 epochs. To that end, we evaluate the (deterministic) forward DDIM process (inverse process to \eq{eq: sampling x0 ddim}) to obtain $\z_T$s. We then evaluate the DDIM sampling process for the interpolated $\z_T$s. The DDIM produces less smooth transitions than the LRDM. Comparing the CelebA-HQ interpolations (\fig{fig: interpolations CelebA-HQ DDIM}) to those in \cite{preechakul_diffusion_2021}, the LRDM also achieves smooth changes of head pose and facial attributes.\\

\noindent\textbf{Sampling progression of LDMs and LRDMs}\\

How is the representation used for image synthesis? To address this question, we display exemplary $\hat{\x}_t$ and $\hat{\x}_0$ progressions for the sampling process in \fig{fig: progression plots repr vs no repr}. Note that the latter can be interpreted as lossy decoding for $t>0$. For the LDM, the $\hat{\x}_0$-estimate gradually changes from close to the dataset mean towards the final sample. For the LRDM, the $\hat{\x}_0$-estimate is close to the final sample in terms of the overall shape and course features, from the first sampling step on. $\hat{\x}_0$ then stays almost the same for a certain timestep range, before local features and details emerge eventually. The results suggest that for a certain timestep range the $\hat{\x}_0$-prediction is only based on the representation, and for lower $t$ the sampling process is more similar to that of the LDM.\\

The above observations indicate that the noise or sampling schedule could be adapted and the number of steps reduced while preserving sample quality. In \cite{preechakul_diffusion_2021}, the authors can confirm this for their model setup. With the additional information from the denoised image, $p(\x_{t-1}\mid\x_t, \rep)$ may be gaussian when $p(\x_{t-1}\mid\x_t)$ was not (e.g. due to an increased step size). However, we leave investigations in this direction for future work.

\FigReconstructionsDDIMLSUNchurchDeep

\FigReconstructionsDDIMCelebAHQ

\FigInterpolationsDDIMReprVsVQGAN

\FigInterpolationsDDIMCelebAHQ

\FigProgressionRepr

\subsection{Combining Latent VAE and Latent Diffusion}\label{sec:from LVAE to LDM}
This section quantitatively analyzes the properties of the learnt representation space and the performance in the task of image synthesis, dependent on the regularization strength $\lambda$. Increasing $\lambda$ in \eq{eq: LRDM loss} reduces the information throughput in the representation bottleneck as the approximate posterior is pushed towards the gaussian prior. Thus, for high $\lambda$ the model should approach a LDM, leaving the representation encoder unused. Decreasing $\lambda$ imposes a stronger relative weight on $L_\mathrm{image}^\ast$ encouraging the model to learn a more powerful representation of the input data.\\

First, we look at the distortion of $\hat{\z_0}$ over time, see \fig{fig: rmse repr vs no repr}. For weaker regularization, the distortion curve develops a plateau above a certain timestep, where the distortion is reduced to an almost constant value compared to the LDM. The plateau matches the observations made in \fig{fig: progression plots repr vs no repr}. For $\lambda=0.1$, the distortion curve closely matches that of the LDM.\footnote{Note that the LDM $\z_0$-distortion curve in \fig{fig: rmse repr vs no repr} does not match that in \fig{fig: distortion b} because the LDM was trained twice as long for the latter.}\\

\FigRMSEreprVSnorepr

\newpage
\noindent\textbf{Quantitative evaluation of LRDMs}\\

For a quantitative evaluation, LRDMs (VQ-AE) are trained for different regularization strengths $\lambda$. \tab{tab: repr metrics} shows results for the sampling and reconstruction quality after 100 epochs. The statistics in \tab{tab: repr metrics} are visualized in \fig{fig: lambda sweep stats}. For the unconditional sampling FID ($50k$), we sample $\rep$ and $\z_T$ from the gaussian priors and then evaluate the 1000 reverse diffusion steps via probabilistic sampling (\eq{eq: sampling x0}). All reconstruction metrics were evaluated on the validation set. For the reconstruction FID, we obtain the representation by encoding the input image and sampling from the posterior distribution $q_\phi(\rep\mid\z_0)$. Here, too, we use the probabilistic sampling scheme. For the MSE $\langle\norm{\x_0-\hat{\x}_0}^2\rangle$, RMSE $\sqrt{\langle\norm{\x_0-\hat{\x}_0}^2\rangle}$, and the variance, $\rep$ is obtained by taking the mode of $q_\phi(\rep\mid\z_0)$, and we use DDIM sampling for less stochasticity in the generative process. The metrics are then evaluated in image space, i.e., after passing the sampled latent through the VQ-AE decoder.\\

As baselines for the reconstruction quality, the scores for the pretrained VQ-AE as well as for a latent VAE (LVAE) are reported. For the latter, we trained a model with the LRDM architecture only on the last timestep $t=T$ such that the loss reduces to that of an unconditional VAE. Sampling is then done by a single pass through the model. In the image of DMs as stacked VAEs (see \sec{sec:experiment repr learning}), this corresponds to training and evaluating only the top-most VAE of the stack.\\

The FID for unconditional sampling indeed reaches that of the LDM\footnote{For a fair comparison, we compare to the LDM (VQ-AE) evaluation after 100 epochs, see \tab{tab: FID z0 vs eps}.} for stronger regularizations. Below a $\lambda$ of around 5e-4, the approximate posterior is no longer close the gaussian prior and hence the FID grows rapidly. In order to sample from the representation space in this regime, one could train a separate model (e.g. diffusion model) there – as done in \cite{preechakul_diffusion_2021}.
The recFID is improves that of the LVAE for lower $\lambda$. There, we see the benefits of both VAE and DM coming together, with the VAE-like reconstruction objective allowing for faithful reconstructions and the (reverse) diffusion process improving the perceptual image quality.
\enlargethispage{\baselineskip}
Regarding the reconstruction RMSE and variance, we see that changing the regularization factor "interpolates between" the LDM and LVAE.\\

Futhermore, we compare different dimensionalities of the representation space. The results indicate a small improvement for the spatially more compressed representation, i.e., when ther is less inductive bias towards spatial features. However, for the spatially more compressed representation, the diffusion U-Net also has an additional depth-level which might also be the relevant factor responsible for the improvement. Further architectural ablations could help investigating this question, see \sec{sec:limitations discussion}.\\

VAEs inherently entail a trade-off between reconstruction and sampling quality. We observe the same for the LRDM. For $\lambda>\num{5e-4}$, where the sampling quality approaches that of the LDM, the reconstruction fidelity already starts to decrease rapidly. Samples and reconstructions for different regularization factors are shown in the Appendix in \fig{fig: samples LRDM different lambdas} and \fig{fig: reconstructions LRDM different lambdas}, respectively. In order to learn a comprehensive representation more efficiently, we experiment with a relaxed timestep-masking during training, i.e., we sample $t$ uniformly from a timestep-range that successively increases from $[T-\epsilon,T]$ to $[0,T]$ over a certain initial training phase. However, we observe no improvements in terms of sampling and reconstruction quality.

\TabReprStatsLSUNchurchCombined

\FigFromVAEtoDDPM

\subsection{Timestep-conditional Representation}\label{sec:experiment t-cond repr}
This section evaluates the t-LRDM, which extends the LRDM by introducing a latent variable $\rep_t$ for each timestep. This increases the dimensionality of the representation space by a factor of $T$ and the model should thus be able to learn a richer representation \cite{abstreiter_diffusion-based_2021}. In particular, the encoded $\rep_t$ may contain the specific information that is needed to predict $\z_0$ from $\z_t$. E.g., for $t\rightarrow0$, almost no extra is needed, and useful, to predict the denoised image. For the LRDM, this causes information to be pushed out of the representation space due to the KL-divergence loss, while for the t-LRDM other $\rep_t$ may be unaffected. The timestep-conditioning of the representation model is realized in the same way as in the denoising U-Net, i.e., the positional timestep-embedding is integrated into the residual blocks.\\

We train t-LRDMs with representation shape $16\times16\times16$ for two different regularization factors for 100 epochs. Note that while the training time is the same as for the LRDM, the reconstruction time increases since at each reverse process step, $\rep_t$ has to be computed anew. Results for the sampling and reconstruction quality can be found in \tab{tab: repr metrics}. \fig{fig: rmse t-cond repr} shows the distortion curve (evaluated in latent space) and the KL-Divergence between the encoded distribution of $\rep_t(\z_t,t)$ and the gaussian prior. Interestingly, the latter exhibits a sharp transition at a certain timestep below which the KL-Divergence is close to zero, and thus the corresponding latent variables $\rep_t$ basically unused, and above which the KL-Divergence is constant. The weaker the regularization the more $\rep_t$ encode information about $\z_0$. The distortion curve has a similar form as that of the LRDM, but with the same sharp transition. Due to these observations, we argue that this transition separates the model in two stages: i) For high $t$, the denoised image is reconstructed from the encoded representation $\rep_t$, in a (conditional) VAE-like manner. ii) For lower $t$, the task becomes an actual image denoising task (from $\z_t$) as for the LDM.\\

\newpage
Compared to the LRDM, the FID scores and reconstruction metrics are slightly worse, though the timestep-conditional representation should be – in theory – at least as good as the non-$t$-conditional representation \cite{abstreiter_diffusion-based_2021}. Possibly, this might be due to the relatively short training times of 100 epochs. Future investigations may also include ablations of the representation dimensionality. Moreover, the representation learning depends on the parameterization, the reweighting of the diffusion loss terms and the noise schedule – which are all partly interdependent, and worth exploring.\\

\FigTcondReprRMSE

\newpage
\subsection{Conditional Representation Learning}\label{sec:conditional repr learning}
We perform additional experiments for explicit conditioning information. First, we present results for class-conditional representation learning on MNIST \cite{ciresan_high-performance_2011} data. Then, preliminary results for an approach to shape-style disentanglement are shown.\\

\noindent\textbf{Class-conditioning}\\

For the class-conditional model, we additionally condition the denoising U-Net and the representation encoder on the class-label \mbox{$p_\theta(\z_{t-1}\mid\z_t, c, \rep(\z_0, c))$}. The forward process is unchanged. We realize the class-conditioning by adding the class-label-embedding to the positional timestep-embedding \cite{dhariwal_diffusion_2021}. Due to the small size of the MNIST images, the diffusion model is employed directly in the image space. Through the explicit conditioning we ideally remove the class-information from the learnt representation space, such that only the appearance of the digits is encoded. We visualize the learnt representation space by using PCA to identify the two dimensions of highest variance and then sampling from linearly spaced grid-points on a plane that is spanned by the first two principal components. I.e., first the grid-points in representation space are determined, then we sample from the model for each of them (with fixed class-label). As displayed in \fig{fig: mnist}, appearance features such as linewidth and tilt gradually change along the grid axes. In such a representation space we can also smoothly interpolate between different appearances. Having separated the class and appearance information, we can change one while keeping the other fixed. In \fig{fig: mnist class switch}, we obtain the $\rep$ from the input images and then reconstruct from $\rep$ conditioned on different class-labels. The representation used here has a significant spatial extent ($24\times7\times7$) and thus provides strong spatial guidance during sampling. This could be modified by including less spatial bias, e.g.\ by using no spatial extent for the representation.\\

\FigMNIST

\newpage
\noindent\textbf{Style-Shape-Disentanglement}\\

In another experiment, we seek to disentangle the style and the shape of images. For that, the representation encoder $\mathcal{E}_r$ is replaced by two encoders $\mathcal{E}_r^\mathrm{shape}$ and $\mathcal{E}_r^\mathrm{style}$ (compare \fig{fig: architecture representation learning}). The former is analogous to $\mathcal{E}_r$ and directly receives the latent $\z_0$, the latter receives $\z_0$ with the shape information removed. That way, $\mathcal{E}_r^\mathrm{style}$ can only encode style information and $\mathcal{E}_r^\mathrm{shape}$ encodes the remaining information, i.e., the shape. How to remove the shape information from an image while preserving the style? Here, we use image warping via Thin Plate Spine (TPS) transformations where the image is deformed given some (random) control points – $\mathcal{E}_r^\mathrm{style}$ then receives $\mathcal{E}(\mathrm{warp}(\x_0))$ (with $\mathcal{E}$ being the pretrained first-stage model). We decide to impose a KL-penalty towards a gaussian prior only for the shape representation to ease the learning task and add skip-connections between $\mathcal{E}_r^\mathrm{style}$ and $\mathcal{D}_r$ (the shape representation $\rep$ is concatenated at the middle block between the two modules). Note that unconditional sampling from the model is no longer possible in this setup.\\

Qualitative results for LSUN-Churches with a representation of shape $4\times32\times32$ are shown in \fig{fig: style shape swap}. For the style-shape-switching, two input images are encoded – one by $\mathcal{E}_r^\mathrm{style}$ and the other by $\mathcal{E}_r^\mathrm{shape}$. The warp transformation is not applied for sampling, but only during training. If both images are the same, the samples are faithful reconstructions of the input. For the mixed samples, we observe that the shape is recovered very accurately, and the style is transferred mainly in terms of the color scheme. Local style attributes like wall texture is not transferred but encoded in the shape representation. We find that increasing the regularization and compressing the shape representation more spatially rapidly leads to a worse shape reconstruction. On the other hand, since we use a symmetric base architecture for $\mathcal{E}_r^\mathrm{style}$ and $\mathcal{E}_r^\mathrm{shape}$, $\mathcal{E}_r^\mathrm{style}$ also only performs spatial compression up to $32\times32$ – whereas less spatial bias might be favorable there. We see room for improvement in choosing the right inductive biases and fine-tuning the architecture as well as the regularization.

\FigStyleShapeSwap

\newpage
\chapter{Conclusion \& Limitations}\label{sec:limitations discussion}
Diffusion models have been able to push the limits of image synthesis through a structured generative process composed of sequential transitions between latent random variables. However, they lack a semantically meaningful and decodable representation as the encoding diffusion process is designed to gradually destroy information in the latent variables.\\

This work explores the topic of representation learning with diffusion models.
To this end, we revisited the image-parameterization for diffusion models. We investigated differences to the commonly used noise-parameteri\-za\-tion and demonstrated that competitive FID scores can be achieved when employing the diffusion model in the latent space of a pretrained autoencoder with VQ-regularization (LDM). While the sample quality slightly falls short of that of the corresponding noise-parameterized models, we think that this gap may be closed by further fine-tuning of the training setup – which has already been done specifically for the noise-parameterization.
By employing the diffusion models in a compressed latent space of a pretrained autoencoder, the computational load compared to employing them directly in image space can be reduced. However, sampling from the model still remains rather slow compared to other generative models like GANs, since the reverse process has to be evaluated sequentially.\\

We then presented the LRDM for representation learning in the framework of diffusion models, finding that the image-parameterization is favorable for this setup and recovering the unconditional VAE objective as edge case ($t=T$) of the LRDM objective. By changing the regularization parameter, we can, to some extent, interpolate between a latent VAE and the LDM.
We evaluated the timestep-conditional t-LRDM as extension of the LRDM. We believe that further fine-tuning of the representation learning can help in extracting a comprehensive and timestep-specific representation, in particular encoding image features of different size at the corresponding timesteps. Future investigations might include a timestep-specific regularization $\lambda(t)$ or finding a stable setup for the $\bm\mu_\theta$-parameterization which is biased in that respect.
Future work may include evaluating the learnt representations in terms of their capabilities in downstream tasks such as classification.
We presented additional results for class-appearance and style-shape separation through conditional representation learning. While for the latter the spatial nature of the representation proved to be crucial, this bias might be limiting for learning global semantics. Thus, exploring other architectural choices and using non-spatial representations would be interesting to investigate in future work.\\

Overall, we introduced a framework for learning semantically meaningful representations with diffusion models and demonstrated that LRDMs are capable of efficiently learning such representations, which allow for faithful reconstructions and smooth interpolations.\\

\bibliographystyle{unsrt}
\bibliography{ms}

\newpage
\appendix
\chapter[Appendix]{}\label{ch:appendix}
\vspace{-2em}

\section{Hyperparameters}\label{sec:apx hyperparameters}
For all diffusion model parameters, we use an EMA with a decay rate of 0.9999. No EMA is used for the representation encoder. Optimization is done using the Adam optimizer. An overview of the various model configurations used for the experiments is shown in \tab{tab: hyperparameters}.\\

\section{Linear Noise-Schedule}\label{sec:apx linear noise schedule}

\FigLinearNoiseSchedule

\begin{landscape}
	\TabHyperparameters
\end{landscape}

\section{DM Objective}\label{sec:apx dm objective}
For completeness, we include the derivation of the DM objective in \eq{eq: variational bound} as done in \cite{sohl-dickstein_deep_2015}.

\begin{equation}\label{apx:eq: dm objective}
	\begin{split}
		L & = \expectation{q}{-\log \frac{p_\theta(\x_{0:T})}{q(\x_{1:T}\mid \x_0)}} \\
		& = \expectation{q}{-\log p(\x_T) - \sum_{t=1}^{T} \log\frac{p_\theta(\x_{t-1}\mid \x_t)}{q(\x_t\mid \x_{t-1})}} \\
		& = \expectation{q}{-\log p(\x_T) - \sum_{t>1}^{T} \log\frac{p_\theta(\x_{t-1}\mid \x_t)}{q(\x_t\mid \x_{t-1})} - \log\frac{p_\theta(\x_0\mid \x_1)}{q(\x_1\mid \x_0)}} \\
		& = \expectation{q}{-\log p(\x_T) - \sum_{t>1}^{T} \log\frac{p_\theta(\x_{t-1}\mid \x_t)}{q(\x_t\mid \x_{t-1},\x_0)}\frac{q(\x_{t-1}\mid\x_0)}{q(\x_t\mid\x_0)} - \log\frac{p_\theta(\x_0\mid \x_1)}{q(\x_1\mid \x_0)}} \\
		& = \mathbb{E}_q\bigg[-\log p(\x_T) - \sum_{t>1}^{T} \log\frac{p_\theta(\x_{t-1}\mid \x_t)}{q(\x_t\mid \x_{t-1},\x_0)} \\ & \hspace{3em} - \log\frac{q(\x_1\mid\x_0)\cancel{q(\x_2\mid\x_0)}\ldots\cancel{q(\x_{T-1}\mid\x_0)}}{\cancel{q(\x_2\mid\x_0)}\ldots\cancel{q(\x_{T-1}\mid\x_0)}q(\x_T\mid\x_0)} - \log\frac{p_\theta(\x_0\mid \x_1)}{q(\x_1\mid \x_0)}\bigg] \\
		& = \mathbb{E}_q\bigg[\kldiv{q(\x_T\mid \x_0)}{p(\x_T)} + \log p_\theta(\x_0\mid\x_1) \\ & \hspace{3em} + \sum_{t>1}^{T}\kldiv{q(\x_{t-1}\mid \x_t,\x_0)}{p_\theta(\x_{t-1}\mid \x_t)}\bigg]
	\end{split}
\end{equation}

An alternative derivation of the DM objective is presented in \cite{ho_denoising_2020}, which is not tractable to estimate but demonstrates progressive coding properties of classical DMs.

\begin{equation}\label{apx:eq: alternative dm objective}
\begin{split}
	L & = \expectation{q}{-\log \frac{p_\theta(\x_{0:T})}{q(\x_{1:T}\mid \x_0)}} \\
	& = \mathbb{E}_q\bigg[-\log p(\x_T) - \sum_{t=1}^{T} \log\frac{p_\theta(\x_{t-1}\mid \x_t)}{q(\x_t\mid \x_{t-1})}\bigg]\\
	& = \mathbb{E}_q\bigg[-\log p(\x_T) - \sum_{t=1}^{T} \log\frac{p_\theta(\x_{t-1}\mid \x_t)}{q(\x_{t-1}\mid \x_t)}\frac{q(\x_{t-1})}{q(\x_t)}\bigg]\\
	& = \mathbb{E}_q\bigg[-\log p(\x_T) - \sum_{t=1}^{T} \log\frac{p_\theta(\x_{t-1}\mid \x_t)}{q(\x_{t-1}\mid \x_t)} -\log\frac{q(\x_0)\cancel{q(\x_1)}\ldots \cancel{q(\x_{T-1})}}{\cancel{q(\x_1)}\ldots \cancel{q(\x_{T-1})}q(\x_T))}\bigg]\\
	& = \kldiv{q(\x_T)}{p(\x_T)} + \mathbb{E}_q\bigg[\sum_{t=1}^{T}\kldiv{q(\x_{t-1}\mid \x_t)}{p_\theta(\x_{t-1}\mid \x_t)}\bigg] + H(\x_0)
\end{split}
\end{equation}

\section{Representation-conditional Objective}\label{sec:apx lrdm objective}
Below, we the derive the ELBO for the representation-conditional DM (\eq{eq: LRDM loss}) as used in the LRDM.\\

For the forward process
\begin{equation}
	q_\phi(\x_{1:T},\rep\mid\x_0) = q(\x_{1:T}\mid\x_0)q_\phi(\rep\mid\x_0)
\end{equation}

and the joint probability
\begin{equation}
	p_\theta(\x_{0:T}, \rep) = p_\theta(\x_{0:T}\mid\rep)p(\rep) = p(\rep)p(\x_T) \prod_{t=1}^T p_\theta(\x_{t-1} \mid \x_t, \rep)
\end{equation}

the (negative) ELBO becomes

\begin{equation}
\begin{split}
	\mathbb{E}[-\log p_\theta(\x_0)] & \leq \expectation{q}{-\log \frac{p_\theta(\x_{0:T},\rep)}{q(\x_{1:T}\mid\x_0)q(\rep\mid\x_0)}} \\
	& = \expectation{q}{-\log p(\x_T) - \log\frac{p(\rep)}{q(\rep\mid\x_0)} - \sum_{t=1}^{T} \log\frac{p_\theta(\x_{t-1}\mid \x_t,\rep)}{q(\x_t\mid \x_{t-1})}} \\
	& = \expectation{q}{-\log p(\x_T) - \log\frac{p(\rep)}{q(\rep\mid\x_0)} - \sum_{t=1}^{T} \log\frac{p_\theta(\x_{t-1}\mid \x_t,\rep)}{q(\x_{t-1}\mid \x_t,\x_0)}\frac{q(\x_{t-1}\mid\x_0)}{q(\x_t\mid\x_0)}} \\
	& = \expectation{q}{\sum_{t=1}^{T} \kldiv{q(\x_{t-1}\mid \x_t,\x_0)}{p_\theta(\x_{t-1}\mid \x_t,\rep)}} \\
	& \hspace{1.2em} + \kldiv{q(\x_T\mid\x_0)}{p(\x_T)} \\
	& \hspace{1.2em}+ \kldiv{q(\rep\mid\x_0)}{p(\rep)}
\end{split}
\end{equation}

where the first two terms can be identified as the DM loss and the third term regularizes the representation encoder towards the latent prior.\\

For the t-LRDM, we have a set of representations $\{\rep_t\}_{t=1...T}$ with the following model setup:
\begin{equation}
\begin{gathered}
	q_\phi(\x_{1:T},\rep_{1:T}\mid\x_0) = q(\x_{1:T}\mid\x_0)q_\phi(\rep_{1:T}\mid\x_0) = \prod_{t=1}^Tq(\x_t\mid\x_{t-1})q_\phi(\rep_t\mid\x_0) \\
	p_\theta(\x_{0:T}, \rep_{1:T}) = p(\x_T) \prod_{t=1}^T p_\theta(\x_{t-1} \mid \x_t, \rep_t)p(\rep_t)
\end{gathered}
\end{equation}

The negative ELBO can be derived analogously to the LRDM.

\begin{equation}
	\begin{split}
		\mathbb{E}[-\log p_\theta(\x_0)] & \leq \expectation{q}{-\log \frac{p_\theta(\x_{0:T},\rep_{1:T})}{q_\phi(\x_{1:T},\rep_{1:T}\mid\x_0)}} \\
		& = \expectation{q}{\sum_{t=1}^{T}\kldiv{q(\x_{t-1}\mid \x_t,\x_0)}{p_\theta(\x_{t-1}\mid \x_t,\rep_t)} \\
		& \hspace{3em}+ \sum_{t=1}^{T}\kldiv{q(\rep_t\mid\x_0)}{p(\rep_t)} }\\
		& \hspace{1.2em} + \kldiv{q(\x_T\mid\x_0)}{p(\x_T)}
	\end{split}
\end{equation}

\section{Additional Samples}\label{sec:apx additional results}

\FigReconstructionsCelebAHQ
\FigReconstructionsLSUNchurchDeep
\FigInterpolationsLSUNchurchDeep
\FigInterpolationsCelebAHQ
\FigSamplesReprLSUNchurchDeep
\FigRecReprLSUNchurchDeep

\end{document}